\documentclass{article} 
\usepackage{iclr2027_conference,times}

\usepackage{amsmath,amsfonts,bm}

\def\eqref#1{equation~\ref{#1}}

\def\1{\bm{1}}

\DeclareMathAlphabet{\mathsfit}{\encodingdefault}{\sfdefault}{m}{sl}
\SetMathAlphabet{\mathsfit}{bold}{\encodingdefault}{\sfdefault}{bx}{n}

\usepackage{hyperref}
\usepackage{url}
\usepackage{enumitem} 
\setlist{topsep=2pt, partopsep=0pt, itemsep=1pt, parsep=0pt, leftmargin=*}
\usepackage{graphicx}
\usepackage{algorithm}
\usepackage{algpseudocode}
\usepackage{amssymb}
\usepackage{booktabs}   
\usepackage{amsmath}
\usepackage{tabularx}
\usepackage{multirow}
\usepackage{etoc}   
\usepackage[most]{tcolorbox}
\usepackage{fvextra}
\newtcolorbox{promptbox}[1][]{%
     colback=gray!4, colframe=black!60, boxrule=0.4pt, arc=2pt,
     left=5pt, right=5pt, top=3pt, bottom=3pt, breakable,
     fonttitle=\bfseries\small, #1}
\usepackage{caption}
\usepackage{placeins}
\usepackage{etoolbox}
\title{Neighborhood Elites: Quality-Diversity Search for Uncertainty Quantification in Question Answering}

\author{Mesut Ceylan\thanks{ Corresponding author.} \;\textsuperscript{1}, Arda Baris Basaran\textsuperscript{1}, Elgar Fleisch\textsuperscript{1,2}, Filipe Barata\textsuperscript{1}\\
\textsuperscript{1}ETH Zurich, Zurich, Switzerland\\
\textsuperscript{2}University of St.Gallen, St.Gallen, Switzerland\\
\texttt{\{ceylanm,abasaran,efleisch,fbarata\}@ethz.ch} \\
}

\iclrfinalcopy 
\begin{document}

\maketitle

\begin{abstract}
Large language models (LLMs) are increasingly capable, yet they produce wrong answers indistinguishable from correct ones. In question answering (QA), such an error propagates unnoticed into downstream tasks. Perturbation-based uncertainty quantification (UQ) addresses this by probing a model over a set of input rewrites and reading uncertainty from how much its predictions vary. Yet this set is usually built generically, by free paraphrasing, token-level perturbation, or adversarial search. Such rewrites leave two things uncontrolled that matter in QA: what kind of change is made to the context, and how explicitly the context supports the answer. When neither is controlled, the composition of the set is determined by the generator rather than by design, and the estimate inherits whatever that generator happens to produce. Quality-diversity (QD) search is built to cover a predefined exploration space, so we recast input perturbation as a QD problem and propose \emph{neighborhood elites (NE)}, a neighborhood constructor that spreads answer-preserving rewrites across two axes, the kind of edit and the explicitness of support. We compare NE against three other neighborhood constructors, holding the editor, QA model, uncertainty measure and search objective fixed, and against three scores read from the original context. Across five extractive-QA datasets and five instruction-tuned LLMs, NE improves failure detection against every other constructor in at least 24 of the 25 dataset-model pairs, reaching 0.726 AUROC and 0.211 AUPRC lift over the base rate. Additionally, NE detects failure better than semantic entropy computed over decoder samples, by 0.074 AUROC and 0.062 AUPRC lift, and this advantage widens as the neighborhood grows.

\end{abstract}
\vspace{-5pt}
\section{Introduction}
\vspace{-5pt}
Large language models (LLMs) have demonstrated strong capabilities in natural language understanding and generation \citep{openai2023gpt4}, question answering and multi-step reasoning \citep{wei2022chain}, and program synthesis \citep{chen2021codex,li2023starcoder}. Beyond single prompts, they increasingly act as the reasoning engine of larger systems, from retrieval-augmented generation (RAG) that grounds answers in external corpora \citep{lewis2020rag,asai2024selfrag} to agentic workflows that plan and call external tools to perform tasks \citep{yao2023react}. In such pipelines, a single wrong answer can propagate through later steps \citep{zhang2024snowball}, so the usefulness of an LLM depends not only on producing an answer, but also on signaling whether that answer can be trusted.

LLMs hallucinate, presenting fabricated facts with the same assurance they provide to correct answers \citep{maynez2020faithfulness,ji2023survey}. They tend to agree with the user rather than correct a mistaken assumption \citep{sharma2024sycophancy}, and rarely express doubt or admit when they do not know \citep{yin2023know}. Asking the model how confident it is, is not helpful, as its stated confidence is poorly aligned with its actual accuracy \citep{xiong2024can}. This motivates uncertainty quantification (UQ): estimating how much to trust an
answer without relying on the model's own account of it \citep{malinin2021uncertainty,kuhn2023semantic,lin2024generating}.

UQ methods can be categorized by the signal used to quantify uncertainty. One family relies on the predictive probabilities assigned by the model \citep{malinin2021uncertainty}, while another measures the variability between multiple predictions. The latter requires only model outputs and is therefore applicable in black-box settings. These approaches differ mainly in the way the predictions are produced. Sampling-based methods sample the model several times on a fixed input
\citep{manakul2023selfcheckgpt,farquhar2024semanticentropy}. Perturbation-based methods instead vary the input, rewriting the prompt in ways that should not change the answer \citep{gao2024spuq,song2025invent}. In either way, the predictions are compared by meaning rather than surface form, which is what semantic entropy measures \citep{kuhn2023semantic}.

In the text setting, a perturbation-based estimate is only as informative as the input variants it compares. QD algorithms \citep{mouret2015illuminating,pugh2016quality}, coupled with LLMs, can orchestrate the search and assembly of such a set of input variants. The LLM carries out the variations, while quality-diversity steers the search, keeping the best candidate found in each cell it explores, so the set spans many strong candidates rather than a single best one. This pairing has proven useful across domains, generating diverse code \citep{lehman2022elm}, creative writing \citep{bradley2024qdaif}, and adversarial prompts \citep{samvelyan2024rainbow}. It is therefore a natural engine for building a set of input variants that vary systematically across a chosen feature space.

However, existing perturbation-based UQ builds this set generically, by model-generated paraphrases of the question and inserted tokens \citep{gao2024spuq}, or by searching over word substitutions \citep{song2025invent}. A key limitation of these methods is that their rules apply to each rewrite on its own, rather than to the set as a whole. A rewrite is described by the operation applied and by how far it moved from the original, and not by its relation to the answer, though a rephrased question can ask for the same answer more or less directly. The composition of the set is therefore determined largely by the generator, rather than by design.

In extractive QA, a perturbation set has to control two dimensions: how the context is modified and how explicitly the rewrite still supports the answer. The answer is inside of the rewritten input, so a rewrite changes its support whether or not anything records the change. Two contexts can preserve the same answer while differing substantially in how directly they support it: a rewrite may keep the answer in place but reach its support through a referring expression or spread that support across several sentences. The answer has not changed, but recovering it now places greater demands on the model.

To this end, we propose a new perturbation-based UQ method built on a QD search that systematically constructs a variation set across a predefined descriptor space. We refer to this set as the \emph{neighborhood} of the input. We hypothesize that uncertainty estimated over such structured neighborhoods contains a richer uncertainty signal for detecting model failures than uncertainty derived from less structured variations or repeated sampling of the original input.
Since the context passage already contains the answer, the neighborhood consists of whole-passage rewrites leaving that answer unchanged. The search steers the generator toward specified combinations of edit type and answer support, targeting one cell after another rather than accepting whatever it produces. We show that neighborhoods built this way detect failure better than those from free paraphrasing and two alternative constructors. This advantage also holds over repeated decoder sampling on the original context when using the same uncertainty measure. Our contributions are as follows.

\begin{itemize}[left=0pt]
\item \textbf{Problem formulation.} We propose the notion of an input \emph{neighborhood} for extractive QA and formulate its construction as a coverage problem over a space with two axes: three kinds of edit and four levels of answer support. A rewrite's coordinates are fixed by deterministic rules rather than by a model
judging its own output. To our knowledge, this is the first perturbation set for UQ built by QD search to cover a space declared in advance.
\item \textbf{Method.} We introduce \emph{neighborhood elites (NE)}, an LLM-guided QD search that steers an editor toward the empty cells of a descriptor space spanned by edit type and answer support, and admits a rewrite only when a deterministic check confirms the requested edit was realized. Its rewrites compound along a lineage, each inheriting an already-edited context and applying one further answer-preserving edit, so the neighborhood reaches contexts that no single edit of the original could produce.
\item \textbf{Empirical evaluation.} We show across five extractive-QA datasets and five instruction-tuned LLMs that NE improves failure detection over every other constructor in at least 24 of the 25 dataset--model pairs, reaching 0.726 AUROC and 0.211 AUPRC lift over the base rate. It exceeds semantic entropy over decoder samples by 0.074 AUROC and 0.062 AUPRC lift, at the same measure, predictions and instances. Its neighborhoods span roughly twice as many kinds of rewrite as the nearest constructor. NE gains 0.100 AUROC as its neighborhood grows from three rewrites to seven while decoder sampling gains 0.030 from the same number of samples.

\end{itemize}
\vspace{-5pt}
\section{Method}\label{sec:method}
\vspace{-5pt}
\subsection{Probing a QA model by perturbing its context}
\begin{figure}[t]
	\centering
	\includegraphics[width=\textwidth]{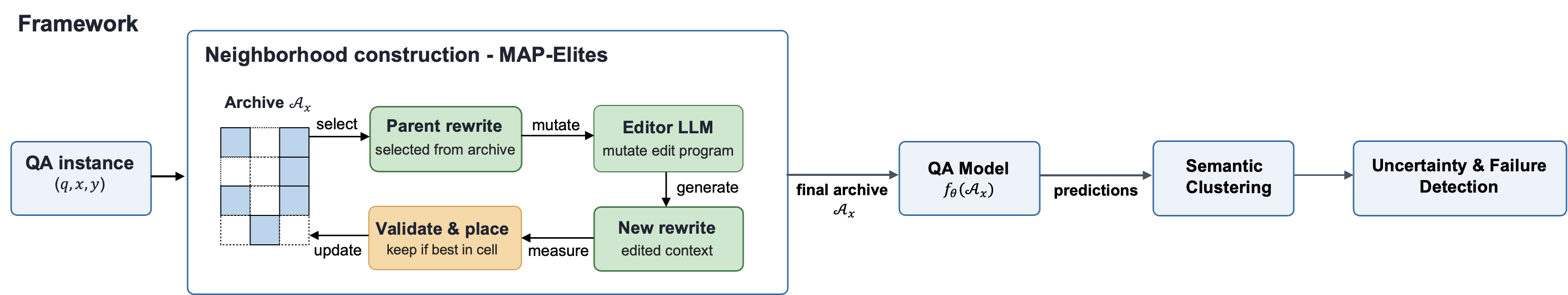}
    \caption{A MAP-Elites search builds a per-instance archive of answer-preserving context rewrites spread over the descriptor space $\mathcal{D}$ (Section~\ref{subsec:archive}). The frozen QA model $f_\theta$ is queried on $K$ of them, and the semantic entropy of its answers is the uncertainty signal (Section~\ref{subsec:readout}).}
  \label{fig:overview}
\end{figure}

\vspace{-5pt}
Given an instance $(q, x, y)$, question, context, and gold answer along with a fixed QA model $f_\theta$, we build a set of alternative contexts that are rewritten yet still contain the answer, query $f_\theta$ on each, and treat disagreement among its answers as an uncertainty signal (Figure~\ref{fig:overview}). Because every rewrite preserves the answer, disagreement among the predictions reflects the model rather than a change in what is being asked.
\vspace{-5pt}
\paragraph{A neighborhood of answer-preserving context.}
We are inspired by the local-distribution view of vicinal risk minimization (VRM) \citep{chapelle-2000}, which replaces each training point with a distribution over nearby examples and applies it at inference on a frozen QA model. We associate each context $x$ with a neighborhood $\nu_x$, a distribution over answer-preserving rewrites $\tilde{x}$ of $x$. We never train on $\nu_x$ and never sample it in closed form, but construct it from a finite archive produced by a QD search.


\vspace{-5pt}
\paragraph{Membership, locality, and coverage.}
Three considerations shape $\nu_x$. First, membership is decided by answer preservation alone: $\tilde{x}$ belongs to $\nu_x$ only if $y$ remains recoverable from it. Among qualifying rewrites, $\nu_x$ places more mass on those semantically close to $x$, so the neighborhood stays local rather than an arbitrary collection of answer-containing contexts. Second, spanning qualitatively different ways of presenting the same answer is not a membership condition but the goal of the search, and we make it concrete through a $3\times 4$ \emph{descriptor space} $\mathcal{D}$. Its rewrite-type axis $\mathcal{T}=\{\text{distractor},\ \text{reorder},\ \text{style rewrite}\}$ names the kind of edit applied, and its support-explicitness axis $\mathcal{S}=\{\text{direct},\ \text{paraphrastic},\ \text{discourse-mediated},\ \text{multi-sentence}\}$ records how explicitly the rewritten context states the answer, from a single verbatim supporting sentence to support dispersed across several sentences. Third, covering $\mathcal{D}$ is how the search explores the neighborhood broadly rather than resampling one cell of it. The deterministic rule assigning a rewrite to its support-explicitness cell is in Appendix~\ref{app:support_explicitness}.

\begin{figure}[t]
	\centering
    \includegraphics[width=\textwidth, trim=0 1 0 1, clip]{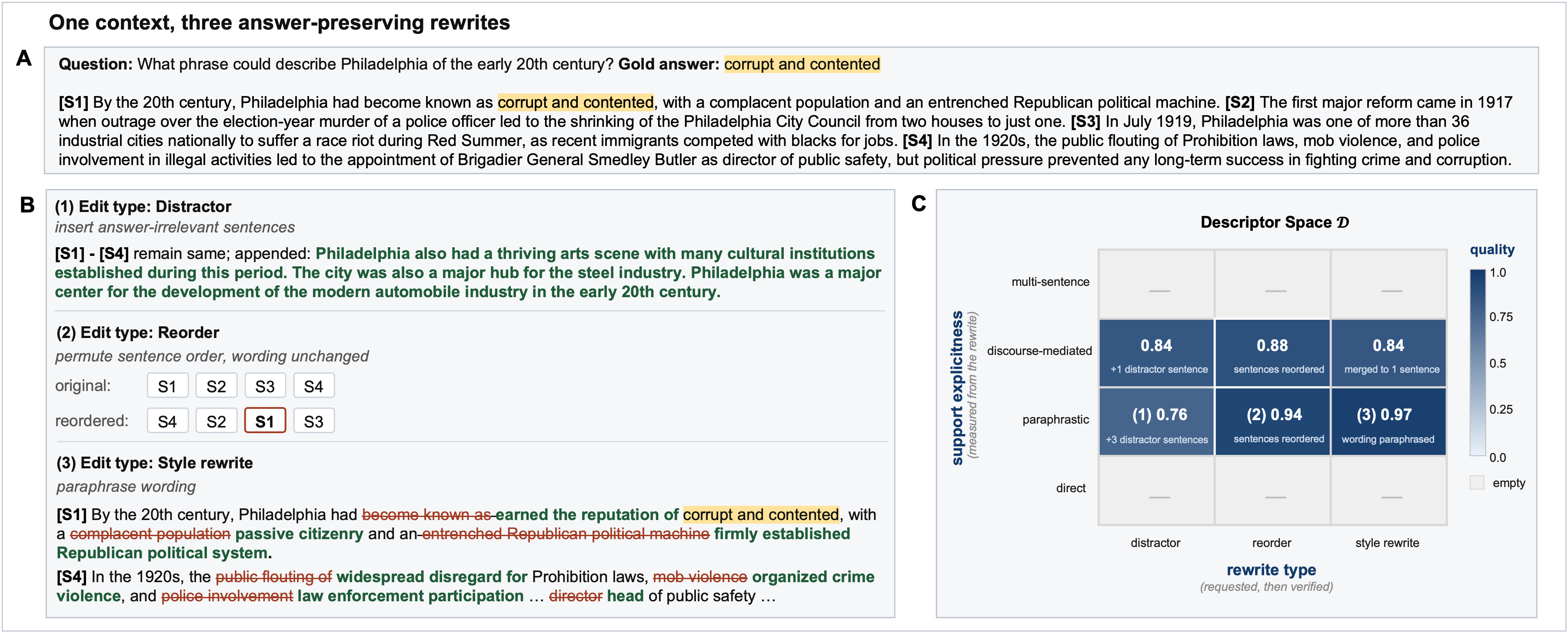}
    \caption{\textbf{A}~SQuAD v1.1 instance: question, gold answer (highlighted), context \texttt{[S1]}--\texttt{[S4]}. \textbf{B}~three rewrites, distractor (1), reorder (2) and style rewrite (3), each preserving the gold answer (red deleted, green inserted). \textbf{C} ~the instance archive over $\mathcal{D}$, one elite per cell, shaded by search-objective value (darker higher, grey empty). Edits (1)--(3) are the marked paraphrastic-support cells; $6/12$ cells are filled.}
  \label{fig:qualitative_example}
\end{figure}

\vspace{-10pt}
\paragraph{The archive as a finite coverage of the neighborhood.}
Because $\nu_x$ is only defined implicitly, we represent it per instance by a finite archive $\mathcal{A}_x = \{\tilde{x}_1, \dots, \tilde{x}_N\}$ of validated rewrites that spreads across the cells of $\mathcal{D}$. The archive is a \emph{coverage} of the neighborhood rather than a sample from a known distribution. Its representation comes from occupying many distinct cells of  $\mathcal{D}$, not from reproducing any generative frequency. Figure~\ref{fig:qualitative_example} shows one archive, three of its rewrites and the cells they occupy. 
\vspace{-10pt}

\subsection{Constructing the neighborhood with targeted MAP-Elites}
\label{subsec:archive}
\vspace{-5pt}
We construct the archive $\mathcal{A}_x$ with \emph{neighborhood elites} (NE), an LLM-guided QD search that separates linguistic variation from archive organization: an editor LLM produces candidate answer-preserving rewrites of $x$, and MAP-Elites structures their exploration and decides which are kept. The rest of this section specifies NE.
\vspace{-5pt}
\paragraph{MAP-Elites and the archive.}
By design, MAP-Elites maintains many high-performing rewrites rather than converging to a single best one \citep{mouret2015illuminating}. We partition the archive into the cells of $\mathcal{D}$, each a distinct realized mode of context variation, and every occupied cell keeps its highest-scoring valid rewrite. Coverage is a property of the archive rather than a term in the search objective.
\vspace{-5pt}
\paragraph{Editor LLM and edit programs.}
The editor LLM is the core component that turns a structured edit request into a natural-language rewrite. Given the question $q$, the context $x$, the fixed answer specification, and an edit program (a request for a particular kind of change), editor LLM returns a rewrite that makes that change while keeping the answer to $q$ recoverable. Our edit programs follow the tradition of behavioral and robustness testing for NLP, in which meaning-preserving transformations probe a model's stability
\citep{ribeiro2020checklist,textflint2021}. Several of these edits mirror standard operations from this line, such as syntax changes, discourse organization, the placement of the answer's support, and the surrounding context. We provide the prompt templates that instantiate each edit type, together with their strategy variants in Appendix~\ref{app:editor_prompt_templates}.

\vspace{-5pt}
\paragraph{Evolutionary search with hard inheritance.}The QD search begins by applying edit programs directly to $x$ and filling the archive with one rewrite at a time, which seeds the initial cells. Each later step either selects an archived rewrite as a parent or targets an empty cell to widen coverage. It then mutates the edit program and asks the editor to apply it. When a parent rewrite is used, the editor edits that parent rather than $x$, so edits accumulate over a lineage, and the search reaches rewrites that no single edit of $x$ could produce. Because empty cells are targeted before occupied ones are refined, the archive first spreads across $\mathcal{D}$ and only then improves the rewrites it already holds.

Throughout a lineage, we keep the question $q$, gold answer set, and the original context $x$ fixed for answer preservation, and allow only the current parent to be edited, yielding an offspring. An offspring must therefore be a genuine change to its parent yet still keep the original answer recoverable. This limits semantic drift without preventing multi-step change. Appendix~\ref{alg:hard_inheritance_mapelites} gives the algorithm and Appendix~\ref{app:search_scheduler} the scheduling and budget. The constructors we compare against are defined in Section~\ref{sec:experimental_setup}.

\vspace{-5pt}

\vspace{-5pt}
\paragraph{Archive admission and update.}
Admission to the archive separates the \emph{feasibility} of a rewrite from its value under the \emph{search objective}. Every candidate must clear three deterministic gates before it is scored or placed in archive. The rewrite must first pass the deterministic answer-preservation gate (see Appendix~\ref{app:ans_preserving_val}). Failed rewrites are discarded from the search. A second gate checks that the rewrite realizes the requested edit type. This gate is a deterministic, rule-based check, and it is independent of the editor. The requested type is taken from the edit program, and the guard compares the output rewrite against the context from which it is edited (the original $x$ for seeds, the parent for offspring). The third check rejects a candidate that is identical to a rewrite already in the archive, so each retained rewrite is distinct.

Concretely, each edit type has a structural signature the guard verifies: a \emph{distractor} must keep every reference sentence intact as an ordered subsequence while adding at least one sentence; a \emph{reorder} must keep the same sentences and change only their order; and a \emph{style rewrite} must change the wording without reducing to a pure reordering or a pure insertion (exact conditions in Appendix~\ref{app:realization}). A rewrite that fails the rule for its requested edit type is rejected.

Two coordinates then place a surviving rewrite in  $\mathcal{D}$: the requested edit type gives its rewrite-type cell, and the produced text is measured to give its support-explicitness cell. Because a rewrite failing its type check is rejected, the rewrite-type axis records the edit actually applied rather than the one requested.

\vspace{-5pt}
\paragraph{Cell competition and the search objective.}
Placement of a rewrite in the descriptor space does not guarantee survival. Each cell keeps only a single best rewrite, so the rewrites that land in a cell \emph{compete} for it. The competition is decided by the \emph{search objective}, a normalized weighted mean of three terms: semantic closeness to $x$, surface edit-magnitude from $x$, and fluency (Appendix~\ref{app:quality_score}). A new rewrite takes an empty cell, or replaces the current occupant only if its search objective value is strictly higher. In MAP-Elites terms, this is the fitness function, which ranks the solutions competing for the same cell. We call this term the search objective throughout. Answer preservation is not one of these terms; it is the hard gate that every scored rewrite has already cleared, so it governs admission but not ranking.

\vspace{-5pt}
\subsection{From neighborhood disagreement to an uncertainty score}
\label{subsec:readout}
After search ends, the archive $\mathcal{A}_x$ turns into an uncertainty signal in one step. We query $f_\theta$ once on each rewrite $\tilde{x}$ kept in $\mathcal{A}_x$, and save its prediction. The uncertainty score reads only the neighborhood. We group the predictions into semantic-equivalence classes and take the dispersion of that class distribution as the per-instance score $u(x)$, so that predictions concentrated in one class indicate confidence and predictions spread across several indicate uncertainty. 

We use the discrete form of semantic entropy \citep{kuhn2023semantic,farquhar2024semanticentropy}, which takes the class distribution from the proportion of predictions in each class rather than from their likelihoods, so the score separates surface variation from disagreement in meaning without access to model probabilities. Because $u$ depends only on the multiset of predictions, it sees neither the gold answer nor which constructor built the neighborhood, so any difference in failure detection is attributable to the neighborhood rather than to the measure read from it.

\vspace{-10pt}
\section{Experimental Setup}\label{sec:experimental_setup}
\vspace{-5pt}
\paragraph{Editor and QA LLMs.}
We use a fixed Qwen3-8B model as the editor for all neighborhood constructors. We evaluate the constructed neighborhoods using five frozen instruction-following LLMs from distinct model families within a comparable parameter regime: Gemma3-12B, OLMo2-13B-Instruct, Phi4-14B, Mistral-Nemo-12B, and Falcon3-10B-Instruct. We use all models for inference only, and inference settings are held fixed across experiments. 
\vspace{-10pt}
\paragraph{Datasets.}
We evaluate five extractive QA benchmarks: 
SQuAD v1.1 \citep{rajpurkar-etal-2016-squad}, Quoref \citep{dasigi-etal-2019-quoref}, NewsQA \citep{trischler-etal-2017-newsqa}, the Natural Questions
partition of MRQA \citep{fisch-etal-2019-mrqa,kwiatkowski-etal-2019-natural},
and AdversarialQA \citep{bartolo-etal-2020-beat}. We randomly sample $1000$ examples per dataset, shared by all constructors, and normalize each dataset's reference answers, which may be multiple, into a uniform answer specification used for answer-preserving validation. Dataset statistics are in Appendix~\ref{app:datasets}.
\vspace{-10pt}
\paragraph{Neighborhood constructors.}
The three comparators share NE's editor and answer-preservation gate and differ in what they leave out: free paraphrasing has no archive, bucket sampling verifies its edits without evolving them, and the untargeted search evolves them without verifying. Neither of the last two is a single-factor ablation, but between them
they bracket the mechanisms NE combines.


\vspace{-5pt}
\begin{enumerate}[left=0pt]
\item \textbf{Free-form paraphrasing (FP).}
\textsc{FP} generates independent answer-preserving rewrites directly from the original context $x$. We denote the resulting paraphrases by $\tilde{x}_1, \tilde{x}_2,\ldots, \tilde{x}_N$, giving the generation pattern $x \rightarrow \tilde{x}_1, x \rightarrow \tilde{x}_2,\ldots, x \rightarrow \tilde{x}_N.$
Each $\tilde{x}_i$ is produced independently from $x$ with the same editor LLM, without a descriptor space, parent--offspring lineage, or evolutionary search. This constructor tests whether conventional free-form reformulation alone provides sufficient predictive variation for uncertainty estimation.

\item \textbf{Untargeted MAP-Elites (U-ME).}
\textsc{U-ME} adds parent--offspring generation and descriptor cell-based selection, following $x \rightarrow P \rightarrow O$, where $x$ is the original context, $P$ is an archived rewrite selected as the parent, and $O$ is the generated offspring. The parent $P$ guides the generation of $O$, but inheritance is not enforced, so the offspring may only weakly reflect the parent transformation or may ignore $P$ entirely. Candidate rewrites are validated, assigned to cells, and retained through competition on the search objective. 
\item \textbf{Static bucket sampling (SBS).}
\textsc{SBS} keeps the archive but removes the search. It issues a fixed budget of independent proposals from the original context $x$, cycling the three rewrite types in turn under a single prompt strategy, so the generation pattern is $x \rightarrow \tilde{x}_1,\ x \rightarrow \tilde{x}_2,\ \ldots$ with no parent and no lineage. Each candidate clears the same gates as NE, including the realization check, and is then placed in the cell given by its rewrite type and its measured support, where it competes on the search objective exactly as in the two MAP-Elites constructors. Placement is therefore decided after generation rather than aimed at. This constructor isolates the evolutionary component of the proposed constructor: it asks whether the archive and its gates alone recover the coverage that lineage and cell targeting produce.

\end{enumerate}
\vspace{-10pt}
\paragraph{Original-context metrics.}
We additionally report three scores that measures uncertainty from the original context, and never the correctness of its prediction. \textbf{NLL} is the negative log-likelihood of the QA model's own greedy answer on $x$, normalised by answer-token count \citep{malinin2021uncertainty}. It is one score per instance. \textbf{Predictive entropy (PE)} is the mean of the length-normalised negative log-likelihoods of answers sampled on $x$, the length-normalised predictive entropy of \citet{malinin2021uncertainty} as used by \citet{kuhn2023semantic}. \textbf{Decoder semantic entropy (DSE)} applies the same discrete semantic entropy we read from a neighborhood, over the same clustering, to decoder samples of $x$ rather than to rewrites of it \citep{farquhar2024semanticentropy}. For each instance, it draws as many samples as NE's archive holds rewrites, so against NE the measure, the number of predictions per instance and the instances are all held fixed, and the only difference is whether the variation comes from resampling the decoder or from rewriting the context. Samples are drawn at temperature $0.7$ and top-$p$ $0.95$, since greedy decoding on a fixed context returns the same answer every time and an entropy of zero.

\vspace{-5pt}
\paragraph{Grouping answers by meaning.} 
The uncertainty measure requires grouping the predictions by meaning rather than by surface string. We use a frozen natural-language-inference (NLI) model, \texttt{DeBERTa-v3-base-mnli-fever-anli}, and place two answers in the same semantic-equivalence class when each entails the other (bidirectional entailment), following the standard construction for semantic entropy \citep{kuhn2023semantic}. Semantic entropy measures disagreement over answer meanings rather than surface forms. With $p_c$ the fraction of $K$ predictions in semantic class $c$, the discrete semantic entropy is $H_{\mathrm{sem}}=-\sum_c p_c\ln p_c$.

\vspace{-6pt}
\paragraph{Evaluation metrics.}
We evaluate each uncertainty score by failure detection, measured by AUROC and AUPRC. A reliable score is higher on failures than on correct predictions, so we take an incorrect original-context prediction as the positive class, paired with that instance's uncertainty score, following
established LLM uncertainty and hallucination-detection evaluations
\citep{kuhn2023semantic,farquhar2024semanticentropy,xiong2024can,manakul2023selfcheckgpt}. AUROC measures how well the score ranks failures above correct predictions. AUPRC additionally depends on the failure rate, which differs across methods with their scoring pools, so we report it as \emph{lift}, the AUPRC minus the base rate (the share of instances in the pools whose original-context prediction is incorrect).

\vspace{-5pt}
\section{Results}

\paragraph{How the neighborhood is built moves detection more than which score is
read from it.} \label{sec:results_span}
NE reaches $0.726$ AUROC and $0.211$ AUPRC lift on the $25$-pair macro, ahead of every other method (Table~\ref{tab:main}). The untargeted search (U-ME) is the nearest constructor, by $0.062$ AUROC and $0.060$ AUPRC lift. Across the four constructors, which share the same uncertainty measure, detection runs from $0.621$ to $0.726$. Across the three scores read from the original context it runs from $0.652$ to $0.676$, less than a quarter as wide, and on AUPRC lift the two spans are $0.113$ to $0.211$ against $0.130$ to $0.149$. Section~\ref{sec:results_pairs} reports where the lead does not hold pair by pair.

 
\begin{table}[t]
\centering
\small
\setlength{\tabcolsep}{5pt}
\caption{Failure detection, macro-averaged over $25$ dataset--model pairs. Upper block: construction varies, measure fixed. Lower block: scores read from the original context. A method is scored where it has at least two predictions, the smallest count at which semantic entropy is defined, so $n$ differs by method, and the NLL needs only the greedy answer. AUPRC is lift over the base rate ($0.315$--$0.361$). $\Delta$ is the difference from NE on shared instances, with $95\%$ intervals from $2000$ bootstrap replicates resampled within each pair.}
\label{tab:main}
\begin{tabular}{l r cc cc}
\toprule
& & \multicolumn{2}{c}{AUROC} & \multicolumn{2}{c}{AUPRC lift}\\
\cmidrule(lr){3-4}\cmidrule(lr){5-6}
Method & $n$ & Value & $\Delta$ & Value & $\Delta$\\
\midrule
\multicolumn{6}{l}{\textit{Neighborhood constructed}}\\
\textbf{NE} & $21{,}916$ & $\mathbf{0.726}$ & --- & $\mathbf{0.211}$ & ---\\
U-ME & $23{,}911$ & $0.661$ & $+0.062$ {\tiny$[+0.053,+0.070]$} & $0.148$ & $+0.060$ {\tiny$[+0.048,+0.071]$}\\
SBS       & $16{,}288$ & $0.642$ & $+0.096$ {\tiny$[+0.085,+0.106]$} & $0.122$ & $+0.101$ {\tiny$[+0.090,+0.116]$}\\
FP     & $17{,}862$ & $0.621$ & $+0.114$ {\tiny$[+0.105,+0.123]$} & $0.113$ & $+0.110$ {\tiny$[+0.097,+0.123]$}\\
\midrule
\multicolumn{6}{l}{\textit{Original context}}\\
PE & $21{,}916$ & $0.676$ & $+0.050$ {\tiny$[+0.041,+0.059]$} & $0.131$ & $+0.080$ {\tiny$[+0.068,+0.091]$}\\
NLL                & $24{,}975$ & $0.674$ & $+0.053$ {\tiny$[+0.045,+0.063]$} & $0.130$ & $+0.084$ {\tiny$[+0.071,+0.095]$}\\
DSE   & $21{,}916$ & $0.652$ & $+0.074$ {\tiny$[+0.066,+0.082]$} & $0.149$ & $+0.062$ {\tiny$[+0.052,+0.074]$}\\
\bottomrule
\end{tabular}
\end{table}

\begin{figure}[t]
\centering
\includegraphics[width=\textwidth]{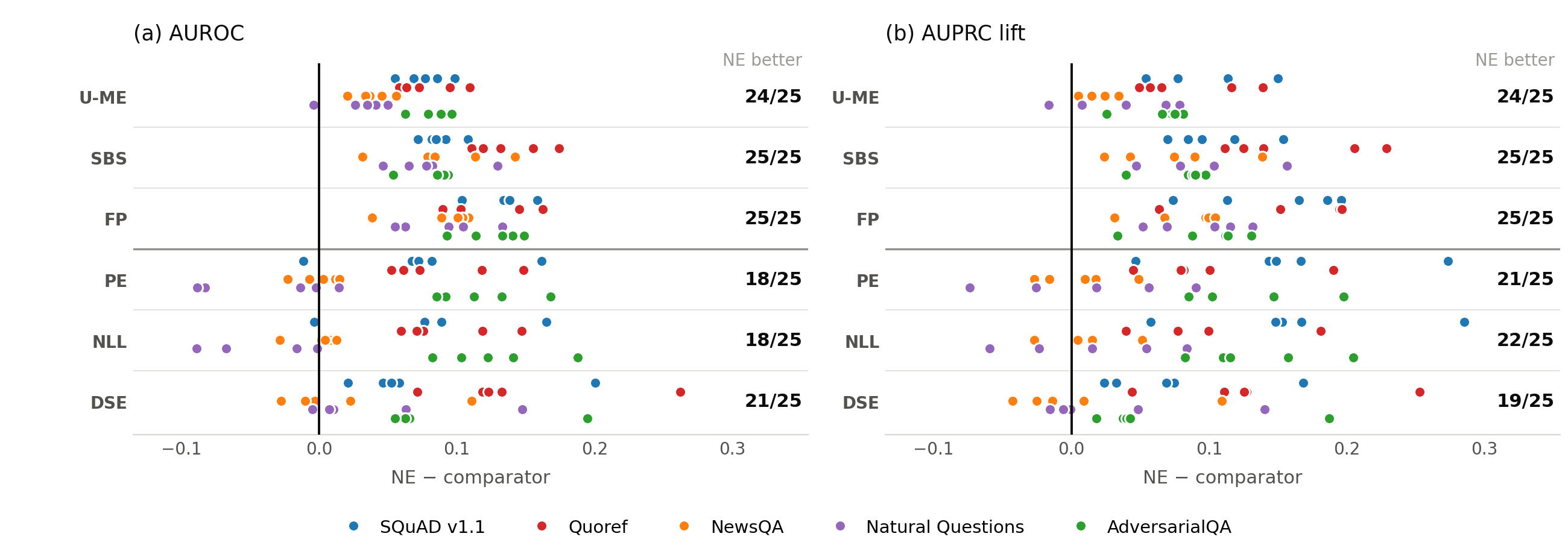}
\caption{Per-pair differences, NE minus each comparator on the instances they share, in AUROC (a) and AUPRC lift (b). Each point is one of the $25$ dataset--model pairs, with colour and vertical offset both encoding the dataset. The upper three rows are neighborhood constructors, the lower three are scores read from the original context. Points right of zero favour NE, and the count beside each row gives how many of the $25$ do. Table~\ref{tab:app_perpair_auroc} and Table~\ref{tab:app_perpair_auprc} in Appendix \ref{app:per_dataset_model_pair} list all $150$ values.}
\label{fig:per_pair_consistency}
\end{figure}

\paragraph{NE leads 74 of 75 constructor and 57 of 75 original-context
comparisons.}\label{sec:results_pairs}
Figure~\ref{fig:per_pair_consistency} plots each of the six differences for all $25$ pairs. NE leads SBS and FP in $25$ of $25$ pairs on both measures, and and U-ME in $24$ of $25$, trailing by $0.004$ AUROC and $0.016$ AUPRC lift on Natural Questions with OLMo2-13B. Median differences across these three rows run from $+0.063$ to $+0.109$ AUROC. Against the original-context scores, the AUROC counts are $18$ of $25$ for PE and the NLL, and $21$ for Decoder SE, and on AUPRC lift $21$, $22$, and $19$. Of the $150$ comparisons, $19$ are negative on AUROC, eleven on Natural Questions, six on NewsQA and two on SQuAD~v1.1, and all $14$ negative AUPRC differences fall on Natural Questions and NewsQA (Table~\ref{tab:app_perpair_auroc} and Table~\ref{tab:app_perpair_auprc}). NE trails furthest against the NLL and predictive entropy, by $0.089$ and $0.088$, both on Natural Questions with OLMo2-13B.
 
\paragraph{NE occupies 4.43 of the twelve cells per neighborhood, roughly twice the nearest comparator, while keeping a comparable number of rewrites.} Table~\ref{tab:cells} reports how much of $\mathcal{D}$ a single neighborhood occupies. On the macro over five datasets NE reaches $4.43$ cells, against $2.33$ for U-ME, $2.00$ for SBS and $0.98$ for FP, while keeping $4.43$ rewrites per instance against U-ME's $5.90$. The two figures coincide because NE, like SBS, places exactly $1.00$ rewrite in each cell it occupies, while U-ME places $2.53$ and FP places $3.77$. NE reaches four or more cells on $0.658$ of instances, against $0.143$ for U-ME, $0.110$ for SBS and $0.001$ for FP, with a standard deviation of $2.15$ cells, about twice that of any comparator. Every constructor occupies fewer cells on Natural Questions and NewsQA than on the other three: NE falls from $5.69$ on AdversarialQA and $5.40$ on SQuAD~v1.1 to $3.33$ and $3.07$, and SBS from $2.87$ on both to $1.17$ and $1.27$ (Appendix \ref{app:archive_diag}, Table \ref{tab:cells_per_dataset}).

\begin{table}[t]
\centering
\small
\setlength{\tabcolsep}{4pt}
\renewcommand{\arraystretch}{1.12}
\caption{What a single neighborhood covers of the descriptor space $\mathcal{D}$, the
grid of twelve cells spanned by edit type and answer support. The first row is the
mean number of distinct cells a neighborhood occupies, with the standard deviation in
brackets. The lower block pools all 5000 instances. Per-dataset values are in Appendix \ref{app:archive_diag},
Table~\ref{tab:cells_per_dataset}.}
\label{tab:cells}
\begin{tabular}{@{}l cccc@{}}
\toprule
& FP & SBS & U-ME & \textbf{NE}\\
\midrule
Cells occupied, of twelve        & $0.98$ \tiny{$(0.66)$} & $2.00$ \tiny{$(1.01)$} & $2.33$ \tiny{$(1.07)$} & $\mathbf{4.43}$ \tiny{$(2.15)$}\\
\addlinespace
Rewrites kept                    & $18439$ & $10024$ & $29506$ & $\mathbf{22163}$\\
Rewrites per instance            &  $3.69$ &  $2.01$ &  $5.90$ & $\mathbf{4.43}$\\
Rewrites per occupied cell       &  $3.77$ &  $1.00$ &  $2.53$ & $\mathbf{1.00}$\\
Instances reaching $\ge 4$ cells & $0.001$ & $0.110$ & $0.143$ & $\mathbf{0.658}$\\
\bottomrule
\end{tabular}
\end{table}

\begin{figure}[t]
\centering
\includegraphics[width=\linewidth]{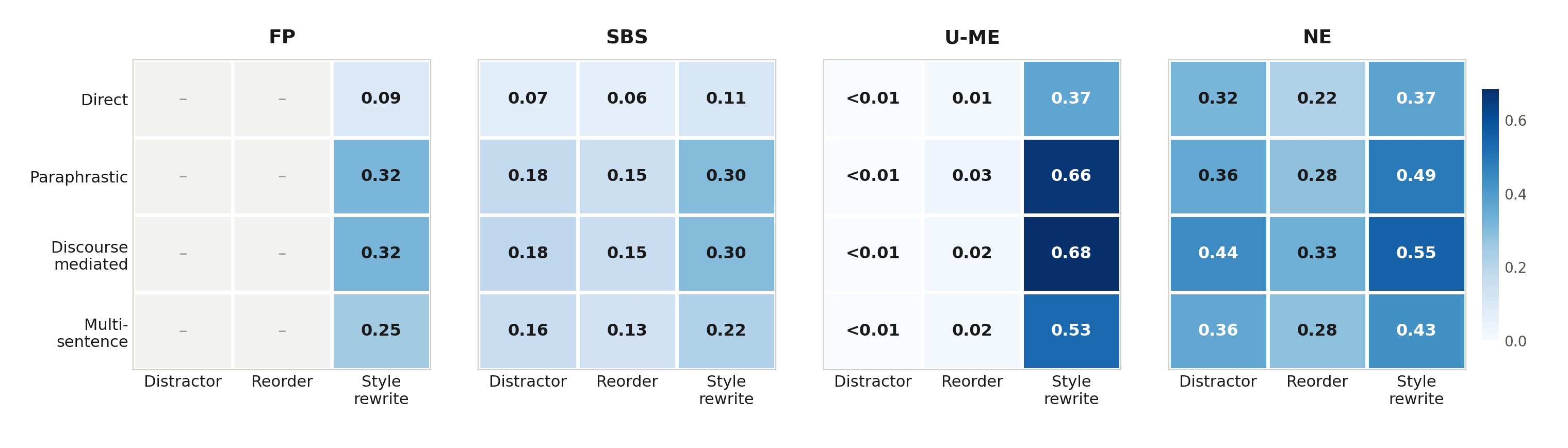}
\caption{For each cell, the fraction of the $5000$ instances whose neighborhood places
at least one rewrite in that cell. The twelve cells of a panel therefore sum to that
constructor's entry in Table~\ref{tab:cells}. A grey cell marked with a dash was never
reached, and $<\!0.01$ marks a cell reached by fewer than one instance in a hundred.
Rewrite totals by column are in Appendix \ref{app:archive_diag}, Table~\ref{tab:diagnostics_details}.}
\label{fig:archives_main}
\end{figure}

\paragraph{NE and SBS reach every cell of $\mathcal{D}$, while U-ME and FP fill the style-rewrite column almost exclusively.} Figure~\ref{fig:archives_main} reports, for each cell, the share of the 5000 instances whose neighborhood places at least one rewrite there. NE's largest cell is style rewriting with discourse-mediated support, at 2749 instances (0.55), and its smallest is reordering with direct support, at 1086 (0.22). SBS reaches every cell at lower levels, from 1484 (0.30) down to 300 (0.06). U-ME exceeds NE in the style-rewrite column, with 3411 instances (0.68) in its largest cell, but reaches only 8 $(<\!0.01)$ in the corresponding distractor cell, having realized the requested edit in 0.27 of its retained rewrites (Appendix \ref{app:archive_diag}, Table~\ref{tab:diagnostics_details}), 24 distractors and 430 reorderings among the 23707 whose type resolved. FP reaches the style-rewrite column only, from 470 (0.09) to 1605 (0.32), with no distractor and no reordering among 18439 rewrites, which is what its editor prompt permits. Style rewriting is the most reached edit type for all four constructors, reordering the least for NE and SBS, and distractors the least for U-ME. Direct support is the least reached level in every panel. Per-dataset values are in Appendix~\ref{app:further_results}, Figure~\ref{fig:archive_per_dataset}.

\paragraph{NE leads at every neighborhood size, and by more as $K$ grows.}
Table \ref{tab:k_sweep_results} demonstrates that at $K{=}5$, NE reaches 0.765 AUROC and 0.282 AUPRC lift, ahead of U-ME by 0.082 and FP by 0.139, where at $K{=}3$ those gaps are 0.068 and 0.098. It leads both other constructors in all 25 dataset--model pairs on AUROC up to $K{=}5$, and in at least 23 at every $K$ (Appendix~\ref{app:per_dataset_model_pair}, Table \ref{tab:app_perpair_auroc}). The original-context scores do not move with $K$: predictive entropy runs from 0.677 to 0.686 across the sweep and the NLL from 0.672 to 0.684.

\begin{table}[t]
\centering
\small
\setlength{\tabcolsep}{1pt}
\caption{Failure detection as neighborhood size $K$ increases, on the matched instance pool for each $K$ (evaluation method described in Appendix~\ref{app:matched_eval}). SBS cannot supply $K$ rewrites and is omitted.}
\label{tab:k_sweep_results}
\begin{tabular}{lcccccccccc}
\toprule
& \multicolumn{2}{c}{$K{=}3$} & \multicolumn{2}{c}{$K{=}4$} & \multicolumn{2}{c}{$K{=}5$}
& \multicolumn{2}{c}{$K{=}6$} & \multicolumn{2}{c}{$K{=}7$} \\
\cmidrule(lr){2-3}\cmidrule(lr){4-5}\cmidrule(lr){6-7}\cmidrule(lr){8-9}\cmidrule(lr){10-11}
Method & AUROC & AUPRC & AUROC & AUPRC & AUROC & AUPRC & AUROC & AUPRC & AUROC & AUPRC \\
\midrule
\textbf{NE}
  & \textbf{0.705} & \textbf{0.215} & \textbf{0.746} & \textbf{0.262} & \textbf{0.765} & \textbf{0.282}
  & \textbf{0.780} & \textbf{0.301} & \textbf{0.805} & \textbf{0.329} \\
U-ME
  & 0.637 & 0.150 & 0.667 & 0.184 & 0.683 & 0.206 & 0.685 & 0.212 & 0.713 & 0.252 \\
FP
  & 0.607 & 0.130 & 0.622 & 0.150 & 0.626 & 0.151 & 0.633 & 0.155 & 0.654 & 0.202 \\
\midrule
PE
  & 0.677 & 0.133 & 0.677 & 0.139 & 0.674 & 0.137 & 0.684 & 0.156 & 0.686 & 0.166 \\
NLL
  & 0.672 & 0.127 & 0.670 & 0.129 & 0.666 & 0.124 & 0.678 & 0.144 & 0.684 & 0.170 \\
DSE
  & 0.635 & 0.156 & 0.647 & 0.171 & 0.650 & 0.178 & 0.655 & 0.183 & 0.665 & 0.202 \\
\bottomrule
\end{tabular}
\end{table}

\section{Discussion and Conclusion}\label{sec:discussion}
\vspace{-5pt}
\paragraph{Principal findings.}
Our results confirm the hypothesis in both of its parts. Uncertainty measured over a structured neighborhood detects failure better than uncertainty from less structured rewrites, and better than uncertainty from decoder sampling of the original context. Neighborhoods that occupy many cells probe the model along several different kinds of rewrite, and each one is a further chance for its prediction to move, so disagreement accumulates. Holding every constructor on the same instances and matching predictions, and raising the number of neighborhood size $K$ from three to seven, widens the gaps: NE gains 0.100 AUROC across the sweep against 0.047 for FP and 0.030 for decoder sampling (Table \ref{tab:k_sweep_results}). A neighborhood confined to one kind of rewrite asks the same question repeatedly, which is why FP, at 0.98 cells, detects failure below NLL. Volume does not compensate for narrow coverage, since U-ME keeps more rewrites per instance than NE (5.90 vs.\ 4.43), covers half as many cells, and scores lower. Against decoder sampling, the comparison is tighter. Resampling needs the decoder made stochastic before it produces any signal, at temperature 0.7, while NE's disagreement appears under greedy decoding. NE leads it by 0.074 AUROC and 0.062 AUPRC lift, so a model's own output distribution on a fixed context carries less signal about failure than the neighborhood of NE around that context. The lead narrows on Natural Questions and NewsQA, the two datasets where every constructor covers the least (Appendix~\ref{app:archive_diag}, Figure~\ref{fig:archive_per_dataset}).

\vspace{-7pt}
\paragraph{Practical implications.}Resampling the predictions from the QA model reads only what the model does with a single context, which limits what it can reveal. Varying the context draws on a different source of variation, and in our comparison it goes further. Under the same uncertainty measure and sample count, NE detects failure better than over decoder samples. Extracting a richer uncertainty signal from a QD search is mostly a matter of supervising the editor. An editor asked to paraphrase produces one kind of change, so the edit to perform and the cell it should reach have to be stated in the prompt. The rewrite that it returns does not guaranty that the requested edit is applied because the editor repeats itself and does not reliably perform the edit it was asked for. Without a check, 0.17 of retained rewrites are duplicates, and only 0.27 realize the requested type. Neither failure is visible to the search unless a deterministic rule catches it, so these are gates rather than preferences. Without them, the descriptor axis records a request rather than a behavior. Empty cells in the defined descriptor space must also be targeted, since an editor asked for more of the same returns to cells the archive already holds. Detection improves from the 8B editor to the 14B and does not progress beyond that, even as the archives continue to grow, so a larger editor does not result in further gains. Detection keeps improving as the neighborhood grows, so the useful size is determined by the supply of rewrites. 

\vspace{-5pt}
\paragraph{Comparison with prior work.} 
What distinguishes this work from prior perturbation-based UQ is not that we vary the input, but that we decide in advance what the variation should span. Sampling-based methods vary the decoder and hold the input fixed \citep{kuhn2023semantic,wang2022selfconsistency}. Rewrite-based methods vary the question or prompt through free paraphrasing \citep{elazar2021consistency}, token-level edits with temperature sampling \citep{gao2024spuq}, or search over token-level mutations \citep{song2025invent}, but none of them lays out a descriptor space that neighborhoods are required to cover edit types and answer support axes. Therefore, the neighborhood is whatever the generator happened to produce, and FP shows the consequence. Among any three rewrites of the same passage, its edit magnitude is narrow, spanning 0.06 against 0.31 for NE (Appendix~\ref{app:within_neighborhood}, Figure \ref{fig:within_instance}), so its probes of a single instance are close to repetitions of one another. QD search has been used to find inputs that break a model \citep{samvelyan2024rainbow}. We use it to find answer-invariant rewrites instead.
\vspace{-5pt}
\paragraph{Limitations.}
A few choices set the scope of what these results show. The axes of the descriptor space took shape through earlier experiments, and it is a design choice rather than a global optimality. Twelve cells may not provide enough resolution for exploring the neighborhood space. Both axes are assigned by rule rather than by a person: the edit type by a deterministic structural check and the support level by a heuristic. Although neither rule is validated against human judgment, both are reproducible, keeping the models out of the decision and applied to the text itself, so \emph{realized} records what a rewrite does rather than what the editor says it did. The four constructors differ in more than one respect each, so we do not attribute the improvement to any single mechanism. NE also runs ten prompt strategies against two for the untargeted search and one each for the other two constructors, a registry that cell targeting entails rather than an advantage beside it. We provide further limitations in Appendix~\ref{app:guarantees_limitations}.


\subsection*{Reproducibility statement}

Everything needed to reproduce the results is specified in the appendix. The four
neighborhood constructors are given as pseudocode over a shared set of operators in
Appendix~\ref{app:constructor_details}, with the deterministic admission gates in
Appendix~\ref{app:deterministic_gates}, the rule that assigns a rewrite to a descriptor cell
in Appendix~\ref{app:descriptor_space}, the search objective that decides cell competition in
Appendix~\ref{app:quality_score}, and the per-constructor schedulers in
Appendix~\ref{app:search_scheduler}. The editor prompts are reproduced verbatim in
Appendix~\ref{app:editor_prompt_templates}, including the system prompt, the per-type
operation strings and the ten prompt strategies. Every fixed value is collected in the
configuration table of Appendix~\ref{app:configuration} and held constant across all datasets
and QA models: the random seed, the editor and its decoding parameters, the objective weights,
the semantic-closeness encoder, the per-example budgets and the evaluation settings. Dataset
preparation and the answer specification format are described in
Appendix~\ref{app:datasets}, prediction scoring and semantic clustering in
Appendix~\ref{app:prediction_scoring}, and the correctness cascade, failure labels and
bootstrap procedure in Appendix~\ref{app:evaluation_protocol}. One limit is stated in
Appendix~\ref{app:guarantees_limitations}: the editor samples without a fixed per-call seed,
so generations are not bit-for-bit reproducible. We therefore treat each run's archive as a
fixed artifact and compute every reported number on it.

\subsection*{AI use statement}
Generative AI tools were used to assist with language/manuscript editing, research-code implementation and verification. The authors independently reviewed and verified all AI-assisted content, executed the reported experiments, and take full responsibility for the final manuscript and its claims.


\bibliography{iclr2027_conference}
\bibliographystyle{iclr2027_conference}

\appendix
\pretocmd{\section}{\FloatBarrier}{}{}
\pretocmd{\subsection}{\FloatBarrier}{}{}
\makeatletter
\setlength{\@fptop}{0pt}                    
\setlength{\@fpsep}{10pt}                   
\setlength{\@fpbot}{0pt plus 1fil}          
\makeatother
\setlength{\floatsep}{8pt plus 2pt minus 2pt}
\setlength{\textfloatsep}{10pt plus 2pt minus 4pt}
\setlength{\intextsep}{8pt plus 2pt minus 2pt}
\renewcommand{\topfraction}{0.9}
\renewcommand{\bottomfraction}{0.8}
\renewcommand{\textfraction}{0.07}
\renewcommand{\floatpagefraction}{0.75}
\setcounter{topnumber}{4}
\setcounter{totalnumber}{6}
\section*{Appendix Contents}
\begin{itemize}[left=0pt]\setlength\itemsep{2pt} 
  \item \hyperref[app:related_work]{\ref*{app:related_work}\quad\nameref*{app:related_work}}  
 \item \hyperref[app:further_results]{\ref*{app:further_results}\quad\nameref*{app:further_results}}
  \begin{itemize}\setlength\itemsep{1pt} 
    \item \hyperref[app:matched_eval]{\ref*{app:matched_eval}\quad\nameref*{app:matched_eval}}
    \item \hyperref[app:per_dataset_model_pair]{\ref*{app:per_dataset_model_pair}\quad\nameref*{app:per_dataset_model_pair}}
    \item \hyperref[app:archive_diag]{\ref*{app:archive_diag}\quad\nameref*{app:archive_diag}}
    \item \hyperref[app:as_deployed]{\ref*{app:as_deployed}\quad\nameref*{app:as_deployed}}
    \item \hyperref[app:editor_size]{\ref*{app:editor_size}\quad\nameref*{app:editor_size}}
    \item \hyperref[app:within_neighborhood]{\ref*{app:within_neighborhood}\quad\nameref*{app:within_neighborhood}}
    \item \hyperref[app:model_acc_results]{\ref*{app:model_acc_results}\quad\nameref*{app:model_acc_results}}
  \end{itemize}
  \item \hyperref[app:neighborhood_construction]{\ref*{app:neighborhood_construction}\quad\nameref*{app:neighborhood_construction}}
  \begin{itemize}\setlength\itemsep{1pt}
    \item \hyperref[app:constructor_details]{\ref*{app:constructor_details}\quad\nameref*{app:constructor_details}}
    \begin{itemize}[label=\textperiodcentered, itemsep=0pt]
      \item \hyperref[app:algo_details_hard_map_elites]{\ref*{app:algo_details_hard_map_elites}\quad\nameref*{app:algo_details_hard_map_elites}}
      \item \hyperref[app:algo_details_soft_map_elites]{\ref*{app:algo_details_soft_map_elites}\quad\nameref*{app:algo_details_soft_map_elites}}
      \item \hyperref[app:algo_details_llm_paraphrase]{\ref*{app:algo_details_llm_paraphrase}\quad\nameref*{app:algo_details_llm_paraphrase}}
    \item \hyperref[app:algo_details_static_bucket_sampling]{\ref*{app:algo_details_static_bucket_sampling}\quad\nameref*{app:algo_details_static_bucket_sampling}}
    \end{itemize}
    \item \hyperref[app:descriptor_space]{\ref*{app:descriptor_space}\quad\nameref*{app:descriptor_space}}
    \begin{itemize}[label=\textperiodcentered, itemsep=0pt]
      \item \hyperref[app:rewrite_type]{\ref*{app:rewrite_type}\quad\nameref*{app:rewrite_type}}
      \item \hyperref[app:support_explicitness]{\ref*{app:support_explicitness}\quad\nameref*{app:support_explicitness}}
      \item \hyperref[app:cell_assignment]{\ref*{app:cell_assignment}\quad\nameref*{app:cell_assignment}}
    \end{itemize}
    \item \hyperref[app:editor_edit_programs]{\ref*{app:editor_edit_programs}\quad\nameref*{app:editor_edit_programs}}
    \item \hyperref[app:deterministic_gates]{\ref*{app:deterministic_gates}\quad\nameref*{app:deterministic_gates}}
    \begin{itemize}[label=\textperiodcentered, itemsep=0pt]
      \item \hyperref[app:ans_preserving_val]{\ref*{app:ans_preserving_val}\quad\nameref*{app:ans_preserving_val}}
      \item \hyperref[app:realization]{\ref*{app:realization}\quad\nameref*{app:realization}}
    \end{itemize}
    \item \hyperref[app:quality_score]{\ref*{app:quality_score}\quad\nameref*{app:quality_score}}
    \item \hyperref[app:search_scheduler]{\ref*{app:search_scheduler}\quad\nameref*{app:search_scheduler}}
    \begin{itemize}[label=\textperiodcentered, itemsep=0pt]
      \item \hyperref[app:search_step]{\ref*{app:search_step}\quad\nameref*{app:search_step}}
      \item \hyperref[app:search_staged]{\ref*{app:search_staged}\quad\nameref*{app:search_staged}}
      \item \hyperref[app:search_soft]{\ref*{app:search_soft}\quad\nameref*{app:search_soft}}
      \item \hyperref[app:search_paraphrase]{\ref*{app:search_paraphrase}\quad\nameref*{app:search_paraphrase}}
    \end{itemize}
  \end{itemize}
  \item \hyperref[app:prediction_scoring]{\ref*{app:prediction_scoring}\quad\nameref*{app:prediction_scoring}}
  \item \hyperref[app:evaluation_protocol]{\ref*{app:evaluation_protocol}\quad\nameref*{app:evaluation_protocol}}
  \item \hyperref[app:configuration]{\ref*{app:configuration}\quad\nameref*{app:configuration}}
  \item \hyperref[app:editor_prompt_templates]{\ref*{app:editor_prompt_templates}\quad\nameref*{app:editor_prompt_templates}}
  \item \hyperref[app:datasets]{\ref*{app:datasets}\quad\nameref*{app:datasets}}
  \item \hyperref[app:guarantees_limitations]{\ref*{app:guarantees_limitations}\quad\nameref*{app:guarantees_limitations}}
\end{itemize}

\section{Related Work}\label{app:related_work}
\paragraph{Uncertainty quantification for LLM QA.}
A prerequisite for reliable QA models is knowing when their answers should be trusted, and post-hoc uncertainty estimation for LLMs comprises several families of methods that differ in what they inspect to estimate it: the probabilities a model assigns to its answer, the confidence it expresses in words, its own judgment of whether output is correct, and the agreement among its repeated outputs. The most direct method assesses confidence from the model's own output probabilities in the form of token-level likelihood, length-normalized sequence likelihood, or sequence predictive entropy \citep{ott2018analyzing, jiang2021how, malinin2021uncertainty}. However, free-form answers expose a difficult problem. As the same answer can be expressed in many surface forms, token-level probability conflates linguistic variation with genuine uncertainty about the answer. A second studies confidence in words, prompting the model to \emph{state} how sure it is, as an explicit numerical probability or a qualitative expression of certainty \citep{lin2022teaching, tian2023just, xiong2024can}. A third has the model judge its own answers, estimating the probability that a proposed answer is correct \citet{kadavath2022language}. A fourth derives uncertainty from \emph{consistency across multiple samples}. Self-consistency aggregates agreement over sampled model outputs \citep{wang2022selfconsistency}, SelfCheckGPT identifies hallucinations by checking whether resampled responses corroborate one another \citep{manakul2023selfcheckgpt}, and semantic entropy clusters sampled answers into semantic equivalence classes under bidirectional entailment and taking the entropy over those classes \citep{kuhn2023semantic}. 

\paragraph{Perturbing the input to probe uncertainty.}
A separate line of work quantifies uncertainty by perturbing the \emph{input}
rather than the output, taking a prediction's stability under perturbation as a confidence signal. SPUQ perturbs the prompt and aggregates the resulting answers into a confidence score \citep{gao2024spuq}, while input clarification ensembling generates disambiguated variants of an ambiguous input and ensembles predictions across them \citep{hou2024decomposing}. More recent methods enrich this family: Inv-Entropy studies uncertainty probabilistically, deriving it from the diversity of inputs consistent with a given output and using a genetic algorithm to diversify the perturbations \citep{song2025invent}, while paraphrase-based decomposition separates prompt-driven from sampling uncertainty by aggregating predictions over semantically equivalent rewordings \citep{cox2025mapping}. These methods share the intuition behind the proposed approach. Self-disagreement across variants of a query signals uncertainty. This premise is well founded, since pretrained models are known to answer inconsistently across meaning-preserving paraphrases \citep{elazar2021consistency}. Nevertheless, they leave two components uncontrolled. First, the rewrites form an unstructured, ad-hoc set with no guarantee of covering distinct kinds of change. Second, they need not preserve the answer, so disagreement can reflect a changed answer rather than genuine uncertainty. Without an explicit construction-time guarantee, answer preservation can be verified only post hoc (e.g., by an LLM judge \citep{zheng2023judging}) rather than enforced by design.

\paragraph{Behavioral testing and answer-preserving perturbations.}
Beyond uncertainty estimation, input perturbation is the basis of behavioral and robustness testing, which probes a model by perturbing its inputs and checking whether predictions stay consistent \citep{dong2023revisit}. CheckList casts this as \emph{invariance} testing: label-preserving perturbations are applied to an input, and the prediction is expected to remain unchanged \citep{ribeiro2020checklist}. Relatedly, semantically equivalent adversarial rules identify meaning-preserving rewrites that flip a model's prediction \citep{ribeiro2018sears}. For reading comprehension specifically, TextFlint provides task-level transforms such as inserting distractor sentences and shifting the answer's position \citep{textflint2021}, and adversarially inserted distractors are a classic stress test of QA models \citep{jia2017adversarial}, and syntactically controlled paraphrasing rewrites an input while varying its syntax \citep{iyyer2018scpn}. These suites are built to measure robustness: they combine answer-preserving and answer-breaking edits such as minimal contrastive edits that flip the gold answer \citep{gardner2020contrast} and report how far accuracy falls. Our context edits are drawn from this vocabulary, and we claim none of the edit types as new. What differs is their use: we retain only the answer-preserving edits and repurpose them not as a robustness benchmark but as a probe for post-hoc uncertainty, where disagreement among predictions, rather than a drop in accuracy, is the quantity of interest.

\paragraph{Quality-diversity search with large language models.}
QD algorithms search for a set of solutions that are individually high-performing yet collectively diverse \citep{pugh2016quality}. One type of QD algorithm is MAP-Elites, which partitions a space of behavioral-descriptor cells over the search space and keeps the single best solution found in each. The search result is an archive of the best solutions that spans the space rather than a single optimum \citep{mouret2015illuminating}. Driving this search requires a variation operator that turns existing solutions into new candidates, and LLMs have proven effective in that role as mutation operators that rewrite a candidate \citep{lehman2022elm}, and as crossover operators that recombine candidates \citep{meyerson2024lmx}. Paired this way, QD has produced broad, systematically diverse collections of model outputs: diverse adversarial prompts for red-teaming \citep{samvelyan2024rainbow}, open-ended creative text guided by model-based feedback \citep{bradley2024qdaif}, and coverage of the prompt space itself \citep{diverseprompts2024}.
\section{Further Results}\label{app:further_results}

\subsection{Detection on Matched Instances and Equal Neighborhood Size}\label{app:matched_eval}
\paragraph{Matched evaluation design.}
Constructors differ in how many valid rewrites they yield, and runs can differ in prediction availability. We therefore build a strict matched evaluation set separately for each dataset, QA model, and neighborhood size $K$. An example enters the set only if it satisfies all of:
\vspace{-5pt}
\begin{enumerate}[left=0pt,itemsep=1pt]
  \item \textbf{Shared instance.} The same test instance has a neighborhood built by every constructor and answered by the QA model.
  \item \textbf{Identical original prediction.} The original-context prediction agrees across runs after standard answer normalization.
  \item \textbf{Identical failure label.} The failure label (the original-context prediction is incorrect) is present and identical for all constructors.
  \item \textbf{Sufficient supply.} Every constructor provides at least $K$ valid answer-preserving rewrites for the example.
\end{enumerate}
\vspace{-5pt}
For every eligible test instance, we then draw exactly $K$ predictions from each constructor, so all constructors are compared on the identical set of examples with an equal number of inputs per example. We sweep $K\in\{3,4,5,6,7\}$, building a separate matched set per $K$; eligibility tightens as $K$ grows, so the set shrinks with $K$, but within a given ($K$,\text{dataset}--\text{model}) it is identical across constructors. 
\vspace{-5pt}
\paragraph{Estimation and reporting.}
A single choice of which $K$ predictions enter a neighborhood is arbitrary, so we average over that choice rather than fixing it. For each $K$ we run $R{=}2000$ repeats; each repeat independently redraws $K$ predictions per example and constructor from the saved archive predictions, then recomputes every uncertainty measure and the resulting AUROC/AUPRC over the fixed matched set (seeding in
Appendix~\ref{app:evaluation_protocol}). We report the mean over repeats and the empirical $[2.5,97.5]$ percentile interval, which quantifies sensitivity to \emph{which} $K$ variants are selected. Constructors are compared \emph{within} a repeat: a paired difference $\Delta$ is formed in each repeat and summarized by its own mean and interval. We call $\Delta$ \emph{reliably positive} when $P(\Delta>0)\geq 0.975$, the fraction of repeats in which it is positive, which is a resampling statement rather than a null-hypothesis test. We report $K=3$ by default because it is the largest matched set and the most conservative point for our method, and we sweep $K$ to show how the comparison behaves as the measure gains resolution. Results are macro-averaged over the $25$ dataset--model pairs, with incorrect original-context predictions as the positive class. Archive-level quantities, marked where they occur, are computed once over the retained rewrites, with no subselection and hence no interval.

\paragraph{Matched set sizes and base rates.}
Table~\ref{tab:matched_sets} gives the size of each matched set and its failure rate.
Counts are dataset--model--example observations, so the ceiling is 25000, five
datasets of 1000 examples answered by five QA models. At $K{=}3$ the matched set holds
15730 of those, and at $K{=}7$ it holds 2273. Shrinkage is uneven across datasets:
NewsQA falls by a factor of eleven over the sweep and SQuAD~v1.1 by under six, so the
composition of the set shifts as $K$ grows and not only its size. The base rate is the
mean of the 25 per-pair failure rates, which is the convention Table~\ref{tab:main}
uses. It does not fall monotonically, because the instances surviving at $K{=}7$ are
drawn disproportionately from the harder datasets.

\begin{table}[ht]
\centering
\small
\setlength{\tabcolsep}{5pt}
\caption{Matched evaluation sets by $K$. Counts are dataset--model--example
observations out of a possible 25000. ``Per pair'' is the mean over the 25
dataset--model pairs. ``Base rate'' is the mean of the per-pair failure rates.\\}
\label{tab:matched_sets}
\begin{tabular}{@{}lcccccrrc@{}}
\toprule
$K$ & SQuAD v1.1 & Quoref & AdversarialQA & Nat.\ Questions & NewsQA & Total & Per pair & Base rate\\
\midrule
3 & 3980 & 3587 & 3673 & 2566 & 1924 & 15730 & 629 & 0.300\\
4 & 2866 & 2637 & 2751 & 1822 & 1281 & 11357 & 454 & 0.284\\
5 & 2476 & 1924 & 2475 & 1257 &  830 &  8962 & 358 & 0.276\\
6 & 1786 & 1209 & 1819 &  675 &  441 &  5930 & 237 & 0.276\\
7 &  705 &  435 &  747 &  210 &  176 &  2273 &  91 & 0.300\\
\bottomrule
\end{tabular}
\end{table}


\begin{table}[ht]
\centering
\small
\setlength{\tabcolsep}{10pt}
\renewcommand{\arraystretch}{1.1}
\caption{Per dataset--model paired $\Delta$AUROC (semantic entropy, $K=3$) for the three
pairwise contrasts among the constructors FP, U-ME, and NE. $\Delta$ is the mean
$\Delta$AUROC and $P=P(\Delta>0)$ over $2000$ repeats; a positive $\Delta$ favors the
first-named constructor. A difference is reliably positive when $P\geq 0.975$.}
\label{tab:per_pair_auroc}

\begin{tabular}{@{}lcccccc@{}}
\\
\toprule
& \multicolumn{2}{c}{\textbf{NE $-$ FP}} & \multicolumn{2}{c}{\textbf{NE $-$ U-ME}} & \multicolumn{2}{c}{\textbf{U-ME $-$ FP}} \\
\cmidrule(lr){2-3}\cmidrule(lr){4-5}\cmidrule(lr){6-7}
\textbf{QA model} & $\Delta$ & $P$ & $\Delta$ & $P$ & $\Delta$ & $P$ \\
\midrule
\multicolumn{7}{@{}l}{\textbf{SQuAD}}\\
Falcon3-10B      & $0.090$ & $1.00$ & $0.085$ & $1.00$ & $0.005$ & $0.64$ \\
Mistral-Nemo-12B & $0.111$ & $1.00$ & $0.058$ & $1.00$ & $0.053$ & $1.00$ \\
Gemma3-12B       & $0.108$ & $1.00$ & $0.053$ & $0.99$ & $0.055$ & $1.00$ \\
OLMo2-13B        & $0.132$ & $1.00$ & $0.086$ & $1.00$ & $0.047$ & $1.00$ \\
Phi-4-14B        & $0.142$ & $1.00$ & $0.085$ & $1.00$ & $0.058$ & $1.00$ \\
\addlinespace
\hline
\multicolumn{7}{@{}l}{\textbf{Quoref}}\\
Falcon3-10B      & $0.064$ & $1.00$ & $0.067$ & $1.00$ & $-0.003$ & $0.43$ \\
Mistral-Nemo-12B & $0.122$ & $1.00$ & $0.082$ & $1.00$ & $0.040$ & $1.00$ \\
Gemma3-12B       & $0.130$ & $1.00$ & $0.096$ & $1.00$ & $0.033$ & $0.98$ \\
OLMo2-13B        & $0.134$ & $1.00$ & $0.093$ & $1.00$ & $0.042$ & $1.00$ \\
Phi-4-14B        & $0.090$ & $1.00$ & $0.071$ & $1.00$ & $0.020$ & $0.90$ \\
\addlinespace
\hline
\multicolumn{7}{@{}l}{\textbf{NewsQA}}\\
Falcon3-10B      & $0.072$ & $1.00$ & $0.077$ & $1.00$ & $-0.005$ & $0.39$ \\
Mistral-Nemo-12B & $0.081$ & $1.00$ & $0.078$ & $1.00$ & $0.004$ & $0.59$ \\
Gemma3-12B       & $0.094$ & $1.00$ & $0.052$ & $1.00$ & $0.043$ & $1.00$ \\
OLMo2-13B        & $0.097$ & $1.00$ & $0.114$ & $1.00$ & $-0.017$ & $0.12$ \\
Phi-4-14B        & $0.012$ & $0.81$ & $0.048$ & $1.00$ & $-0.036$ & $0.01$ \\
\addlinespace
\hline
\multicolumn{7}{@{}l}{\textbf{NaturalQuestions (MRQA)}}\\
Falcon3-10B      & $0.088$ & $1.00$ & $0.035$ & $0.96$ & $0.053$ & $1.00$ \\
Mistral-Nemo-12B & $0.110$ & $1.00$ & $0.043$ & $0.99$ & $0.068$ & $1.00$ \\
Gemma3-12B       & $0.095$ & $1.00$ & $0.052$ & $1.00$ & $0.043$ & $0.99$ \\
OLMo2-13B        & $0.112$ & $1.00$ & $0.031$ & $0.95$ & $0.081$ & $1.00$ \\
Phi-4-14B        & $0.054$ & $1.00$ & $0.038$ & $0.98$ & $0.016$ & $0.82$ \\
\addlinespace
\hline
\multicolumn{7}{@{}l}{\textbf{AdversarialQA}}\\
Falcon3-10B      & $0.079$ & $1.00$ & $0.050$ & $1.00$ & $0.029$ & $0.99$ \\
Mistral-Nemo-12B & $0.127$ & $1.00$ & $0.065$ & $1.00$ & $0.062$ & $1.00$ \\
Gemma3-12B       & $0.106$ & $1.00$ & $0.078$ & $1.00$ & $0.028$ & $0.98$ \\
OLMo2-13B        & $0.108$ & $1.00$ & $0.067$ & $1.00$ & $0.041$ & $1.00$ \\
Phi-4-14B        & $0.103$ & $1.00$ & $0.102$ & $1.00$ & $0.001$ & $0.53$ \\
\bottomrule
\end{tabular}
\end{table}

\begin{table}[!ht]
\centering
\small
\setlength{\tabcolsep}{10pt}
\renewcommand{\arraystretch}{1.1}
\caption{Per dataset--model paired $\Delta$AUPRC (semantic entropy, $K=3$) for the three
pairwise contrasts among the constructors FP, U-ME, and NE. $\Delta$ is the mean
$\Delta$AUPRC and $P=P(\Delta>0)$ over $2000$ repeats; a positive $\Delta$ favors the
first-named constructor, and a difference is reliably positive when $P\geq 0.975$. AUPRC is base-rate
dependent, so per-pair $\Delta$AUPRC is a noisier statistic than the rank-based $\Delta$AUROC
of Table~\ref{tab:per_pair_auroc}.}
\label{tab:per_pair_auprc}
\begin{tabular}{@{}lcccccc@{}}
\\
\toprule
& \multicolumn{2}{c}{\textbf{NE $-$ FP}} & \multicolumn{2}{c}{\textbf{NE $-$ U-ME}} & \multicolumn{2}{c}{\textbf{U-ME $-$ FP}} \\
\cmidrule(lr){2-3}\cmidrule(lr){4-5}\cmidrule(lr){6-7}
\textbf{QA model} & $\Delta$ & $P$ & $\Delta$ & $P$ & $\Delta$ & $P$ \\
\midrule
\multicolumn{7}{@{}l}{\textbf{SQuAD}}\\
Falcon3-10B      & $0.048$ & $0.98$ & $0.073$ & $1.00$ & $-0.025$ & $0.15$ \\
Mistral-Nemo-12B & $0.124$ & $1.00$ & $0.062$ & $0.99$ & $0.062$ & $1.00$ \\
Gemma3-12B       & $0.078$ & $1.00$ & $0.049$ & $0.98$ & $0.029$ & $0.94$ \\
OLMo2-13B        & $0.133$ & $1.00$ & $0.120$ & $1.00$ & $0.012$ & $0.71$ \\
Phi-4-14B        & $0.127$ & $1.00$ & $0.092$ & $1.00$ & $0.035$ & $0.92$ \\
\addlinespace
\hline
\multicolumn{7}{@{}l}{\textbf{Quoref}}\\
Falcon3-10B      & $0.035$ & $0.95$ & $0.062$ & $1.00$ & $-0.027$ & $0.10$ \\
Mistral-Nemo-12B & $0.121$ & $1.00$ & $0.067$ & $1.00$ & $0.054$ & $1.00$ \\
Gemma3-12B       & $0.131$ & $1.00$ & $0.076$ & $1.00$ & $0.055$ & $0.99$ \\
OLMo2-13B        & $0.136$ & $1.00$ & $0.083$ & $1.00$ & $0.052$ & $0.99$ \\
Phi-4-14B        & $0.055$ & $0.99$ & $0.062$ & $0.99$ & $-0.007$ & $0.38$ \\
\addlinespace
\hline
\multicolumn{7}{@{}l}{\textbf{NewsQA}}\\
Falcon3-10B      & $0.071$ & $1.00$ & $0.075$ & $1.00$ & $-0.004$ & $0.41$ \\
Mistral-Nemo-12B & $0.087$ & $1.00$ & $0.093$ & $1.00$ & $-0.005$ & $0.38$ \\
Gemma3-12B       & $0.111$ & $1.00$ & $0.050$ & $0.98$ & $0.061$ & $1.00$ \\
OLMo2-13B        & $0.100$ & $1.00$ & $0.103$ & $1.00$ & $-0.004$ & $0.42$ \\
Phi-4-14B        & $-0.019$ & $0.18$ & $0.045$ & $0.98$ & $-0.063$ & $0.00$ \\
\addlinespace
\hline
\multicolumn{7}{@{}l}{\textbf{NaturalQuestions (MRQA)}}\\
Falcon3-10B      & $0.085$ & $1.00$ & $0.028$ & $0.81$ & $0.056$ & $0.98$ \\
Mistral-Nemo-12B & $0.128$ & $1.00$ & $0.032$ & $0.85$ & $0.096$ & $1.00$ \\
Gemma3-12B       & $0.096$ & $1.00$ & $0.066$ & $0.97$ & $0.030$ & $0.82$ \\
OLMo2-13B        & $0.128$ & $1.00$ & $0.054$ & $0.95$ & $0.073$ & $0.99$ \\
Phi-4-14B        & $0.021$ & $0.77$ & $0.049$ & $0.95$ & $-0.028$ & $0.17$ \\
\addlinespace
\hline
\multicolumn{7}{@{}l}{\textbf{AdversarialQA}}\\
Falcon3-10B      & $0.018$ & $0.84$ & $0.025$ & $0.91$ & $-0.006$ & $0.37$ \\
Mistral-Nemo-12B & $0.101$ & $1.00$ & $0.050$ & $1.00$ & $0.051$ & $1.00$ \\
Gemma3-12B       & $0.073$ & $1.00$ & $0.071$ & $1.00$ & $0.002$ & $0.53$ \\
OLMo2-13B        & $0.075$ & $1.00$ & $0.051$ & $1.00$ & $0.024$ & $0.92$ \\
Phi-4-14B        & $0.068$ & $1.00$ & $0.089$ & $1.00$ & $-0.021$ & $0.12$ \\
\bottomrule
\end{tabular}
\end{table}

\begin{table}[ht]
\centering
\small
\setlength{\tabcolsep}{4pt}
\renewcommand{\arraystretch}{1.1}
\caption{Failure-detection \textbf{AUROC} across $K$ on semantic entropy, macro
over the $25$ dataset--model pairs ($95\%$ subselection percentile intervals). }
\label{tab:ksweep_auroc}
\begin{tabular}{@{}lccccc@{}}
\toprule
\textbf{Constructor} & $K{=}3$ & $K{=}4$ & $K{=}5$ & $K{=}6$ & $K{=}7$ \\
\midrule
\multicolumn{6}{@{}l}{\textit{Semantic entropy}}\\
FP            & $0.607$ {\tiny$[.603,.611]$} & $0.622$ {\tiny$[.619,.626]$} & $0.626$ {\tiny$[.623,.629]$} & $0.633$ {\tiny$[.629,.636]$} & $0.654$ {\tiny$[.648,.658]$} \\
U-ME          & $0.637$ {\tiny$[.633,.642]$} & $0.667$ {\tiny$[.662,.672]$} & $0.683$ {\tiny$[.679,.687]$} & $0.685$ {\tiny$[.681,.689]$} & $0.713$ {\tiny$[.707,.718]$} \\
\textbf{NE} & $\mathbf{0.705}$ {\tiny$[.701,.710]$} & $\mathbf{0.746}$ {\tiny$[.742,.750]$} & $\mathbf{0.765}$ {\tiny$[.761,.768]$} & $\mathbf{0.780}$ {\tiny$[.776,.784]$} & $\mathbf{0.805}$ {\tiny$[.801,.809]$} \\
\bottomrule
\end{tabular}
\end{table}
 
\begin{table}[ht]
\centering
\small
\setlength{\tabcolsep}{4pt}
\renewcommand{\arraystretch}{1.1}
\caption{Failure-detection \textbf{AUPRC} across $K$, on semantic entropy, macro
over the $25$ dataset--model pairs ($95\%$ subselection percentile intervals).}
\label{tab:ksweep_auprc}
\begin{tabular}{@{}lccccc@{}}
\toprule
\textbf{Constructor} & $K{=}3$ & $K{=}4$ & $K{=}5$ & $K{=}6$ & $K{=}7$ \\
\midrule
\multicolumn{6}{@{}l}{\textit{Semantic entropy}}\\
FP            & $0.430$ {\tiny$[.424,.435]$} & $0.434$ {\tiny$[.429,.440]$} & $0.427$ {\tiny$[.422,.432]$} & $0.431$ {\tiny$[.426,.436]$} & $0.502$ {\tiny$[.496,.507]$} \\
U-ME          & $0.450$ {\tiny$[.443,.457]$} & $0.468$ {\tiny$[.461,.476]$} & $0.482$ {\tiny$[.475,.490]$} & $0.488$ {\tiny$[.481,.495]$} & $0.552$ {\tiny$[.543,.561]$} \\
\textbf{NE} & $\mathbf{0.515}$ {\tiny$[.508,.522]$} & $\mathbf{0.546}$ {\tiny$[.538,.553]$} & $\mathbf{0.559}$ {\tiny$[.552,.565]$} & $\mathbf{0.577}$ {\tiny$[.571,.583]$} & $\mathbf{0.629}$ {\tiny$[.623,.636]$} \\
\bottomrule
\end{tabular}
\end{table}

\subsection{Failure Detection per Dataset–Model Pair}\label{app:per_dataset_model_pair}

\begin{table}[ht]
\centering
\small
\setlength{\tabcolsep}{2.5pt}
\caption{Per-pair differences in AUROC between NE and each comparator, for the $25$ dataset--model pairs. Positive values favor NE; the $19$ of $150$ values against NE are underlined. Each difference is computed on the instances the two methods share within that pair under the $\min K=2$ requirement, so a difference is a within-pair contrast and is not the difference of two separately pooled levels. The NE column gives its AUROC on its own pool for scale and should not be used to recover a comparator level by subtraction. Column means equal the paired macro differences reported in Table~\ref{tab:main}. Figure~\ref{fig:per_pair_consistency} plots these values.}
\label{tab:app_perpair_auroc}
\begin{tabular}{l l r r r r r r r}
\toprule
& & & \multicolumn{3}{c}{Neighborhood constructed} & \multicolumn{3}{c}{Original context}\\
\cmidrule(lr){4-6}\cmidrule(lr){7-9}
Dataset & Model & NE & U-ME & SBS & FP & PE & NLL & DSE\\
\midrule
SQuAD v1.1 & Falcon3-10B & $0.759$ & $+0.086$ & $+0.092$ & $+0.104$ & $+0.067$ & $+0.078$ & $+0.058$\\
 & Gemma3-12B & $0.720$ & $+0.068$ & $+0.072$ & $+0.135$ & $+0.082$ & $+0.077$ & $+0.201$\\
 & Mistral-NeMo-12B & $0.727$ & $+0.055$ & $+0.082$ & $+0.134$ & \underline{$-0.011$} & \underline{$-0.004$} & $+0.021$\\
 & OLMo2-13B & $0.768$ & $+0.077$ & $+0.085$ & $+0.158$ & $+0.072$ & $+0.089$ & $+0.046$\\
 & Phi-4-14B & $0.786$ & $+0.098$ & $+0.108$ & $+0.138$ & $+0.162$ & $+0.165$ & $+0.053$\\
\addlinespace[2pt]
Quoref & Falcon3-10B & $0.719$ & $+0.058$ & $+0.111$ & $+0.090$ & $+0.118$ & $+0.119$ & $+0.119$\\
 & Gemma3-12B & $0.781$ & $+0.109$ & $+0.155$ & $+0.163$ & $+0.148$ & $+0.147$ & $+0.262$\\
 & Mistral-NeMo-12B & $0.748$ & $+0.073$ & $+0.132$ & $+0.145$ & $+0.073$ & $+0.076$ & $+0.123$\\
 & OLMo2-13B & $0.776$ & $+0.095$ & $+0.174$ & $+0.163$ & $+0.061$ & $+0.060$ & $+0.132$\\
 & Phi-4-14B & $0.754$ & $+0.063$ & $+0.119$ & $+0.103$ & $+0.053$ & $+0.071$ & $+0.072$\\
\addlinespace[2pt]
NewsQA & Falcon3-10B & $0.683$ & $+0.037$ & $+0.031$ & $+0.089$ & \underline{$-0.007$} & $+0.008$ & $+0.023$\\
 & Gemma3-12B & $0.665$ & $+0.034$ & $+0.142$ & $+0.109$ & $+0.012$ & $+0.013$ & $+0.111$\\
 & Mistral-NeMo-12B & $0.674$ & $+0.046$ & $+0.113$ & $+0.104$ & \underline{$-0.023$} & \underline{$-0.028$} & \underline{$-0.003$}\\
 & OLMo2-13B & $0.716$ & $+0.056$ & $+0.079$ & $+0.101$ & $+0.015$ & $+0.002$ & \underline{$-0.010$}\\
 & Phi-4-14B & $0.695$ & $+0.021$ & $+0.084$ & $+0.039$ & $+0.003$ & $+0.004$ & \underline{$-0.028$}\\
\addlinespace[2pt]
Natural Questions & Falcon3-10B & $0.712$ & $+0.026$ & $+0.065$ & $+0.063$ & \underline{$-0.014$} & \underline{$-0.016$} & $+0.063$\\
 & Gemma3-12B & $0.687$ & $+0.050$ & $+0.082$ & $+0.094$ & \underline{$-0.002$} & \underline{$-0.003$} & $+0.148$\\
 & Mistral-NeMo-12B & $0.682$ & $+0.041$ & $+0.129$ & $+0.133$ & \underline{$-0.083$} & \underline{$-0.068$} & $+0.010$\\
 & OLMo2-13B & $0.647$ & \underline{$-0.004$} & $+0.047$ & $+0.105$ & \underline{$-0.088$} & \underline{$-0.089$} & \underline{$-0.005$}\\
 & Phi-4-14B & $0.708$ & $+0.035$ & $+0.078$ & $+0.055$ & $+0.014$ & \underline{$-0.001$} & $+0.008$\\
\addlinespace[2pt]
AdversarialQA & Falcon3-10B & $0.756$ & $+0.063$ & $+0.054$ & $+0.093$ & $+0.092$ & $+0.103$ & $+0.066$\\
 & Gemma3-12B & $0.753$ & $+0.091$ & $+0.094$ & $+0.140$ & $+0.133$ & $+0.141$ & $+0.195$\\
 & Mistral-NeMo-12B & $0.731$ & $+0.088$ & $+0.085$ & $+0.149$ & $+0.085$ & $+0.082$ & $+0.060$\\
 & OLMo2-13B & $0.743$ & $+0.079$ & $+0.090$ & $+0.133$ & $+0.112$ & $+0.123$ & $+0.063$\\
 & Phi-4-14B & $0.749$ & $+0.096$ & $+0.086$ & $+0.114$ & $+0.168$ & $+0.188$ & $+0.055$\\
\midrule
\multicolumn{2}{l}{\textit{Pairs favoring NE}} & & $24/25$ & $25/25$ & $25/25$ & $18/25$ & $18/25$ & $21/25$\\
\multicolumn{2}{l}{\textit{Mean difference}} & & $+0.062$ & $+0.096$ & $+0.114$ & $+0.050$ & $+0.053$ & $+0.074$\\
\bottomrule
\end{tabular}
\end{table}

\begin{table}[ht]
\centering
\small
\setlength{\tabcolsep}{2.5pt}
\caption{Per-pair differences in AUPRC, same protocol and layout as Table~\ref{tab:app_perpair_auroc}, with $14$ of $150$ values against NE. The NE column gives AUPRC lift over the base rate of that pair, as in Table~\ref{tab:main}, while the differences are differences in AUPRC. A difference in AUPRC and the corresponding difference in AUPRC lift are identical, because both methods face the same base rate on the instances they share. Figure \ref{fig:per_pair_consistency} plots these comparisons.}
\label{tab:app_perpair_auprc}
\begin{tabular}{l l r r r r r r r}
\toprule
& & & \multicolumn{3}{c}{Neighborhood constructed} & \multicolumn{3}{c}{Original context}\\
\cmidrule(lr){4-6}\cmidrule(lr){7-9}
Dataset & Model & NE & U-ME & SBS & FP & PE & NLL & DSE\\
\midrule
SQuAD v1.1 & Falcon3-10B & $0.239$ & $+0.077$ & $+0.085$ & $+0.074$ & $+0.143$ & $+0.153$ & $+0.024$\\
 & Gemma3-12B & $0.190$ & $+0.077$ & $+0.070$ & $+0.113$ & $+0.149$ & $+0.148$ & $+0.168$\\
 & Mistral-NeMo-12B & $0.242$ & $+0.054$ & $+0.095$ & $+0.165$ & $+0.047$ & $+0.058$ & $+0.033$\\
 & OLMo2-13B & $0.291$ & $+0.114$ & $+0.118$ & $+0.186$ & $+0.167$ & $+0.167$ & $+0.074$\\
 & Phi-4-14B & $0.345$ & $+0.150$ & $+0.154$ & $+0.196$ & $+0.274$ & $+0.285$ & $+0.069$\\
\addlinespace[2pt]
Quoref & Falcon3-10B & $0.185$ & $+0.057$ & $+0.112$ & $+0.064$ & $+0.101$ & $+0.099$ & $+0.111$\\
 & Gemma3-12B & $0.274$ & $+0.139$ & $+0.205$ & $+0.195$ & $+0.190$ & $+0.181$ & $+0.253$\\
 & Mistral-NeMo-12B & $0.230$ & $+0.049$ & $+0.139$ & $+0.152$ & $+0.045$ & $+0.040$ & $+0.127$\\
 & OLMo2-13B & $0.289$ & $+0.116$ & $+0.229$ & $+0.196$ & $+0.082$ & $+0.077$ & $+0.125$\\
 & Phi-4-14B & $0.239$ & $+0.066$ & $+0.125$ & $+0.064$ & $+0.080$ & $+0.100$ & $+0.044$\\
\addlinespace[2pt]
NewsQA & Falcon3-10B & $0.143$ & $+0.005$ & $+0.024$ & $+0.067$ & \underline{$-0.016$} & $+0.005$ & $+0.009$\\
 & Gemma3-12B & $0.153$ & $+0.016$ & $+0.139$ & $+0.097$ & $+0.049$ & $+0.052$ & $+0.109$\\
 & Mistral-NeMo-12B & $0.131$ & $+0.034$ & $+0.090$ & $+0.100$ & \underline{$-0.027$} & \underline{$-0.027$} & \underline{$-0.014$}\\
 & OLMo2-13B & $0.166$ & $+0.015$ & $+0.043$ & $+0.104$ & $+0.018$ & $+0.005$ & \underline{$-0.043$}\\
 & Phi-4-14B & $0.167$ & $+0.024$ & $+0.075$ & $+0.031$ & $+0.010$ & $+0.015$ & \underline{$-0.025$}\\
\addlinespace[2pt]
Natural Questions & Falcon3-10B & $0.188$ & $+0.008$ & $+0.046$ & $+0.069$ & $+0.018$ & $+0.015$ & $+0.049$\\
 & Gemma3-12B & $0.185$ & $+0.078$ & $+0.079$ & $+0.104$ & $+0.056$ & $+0.055$ & $+0.140$\\
 & Mistral-NeMo-12B & $0.185$ & $+0.039$ & $+0.157$ & $+0.131$ & \underline{$-0.074$} & \underline{$-0.059$} & \underline{$-0.001$}\\
 & OLMo2-13B & $0.144$ & \underline{$-0.016$} & $+0.047$ & $+0.115$ & \underline{$-0.026$} & \underline{$-0.024$} & \underline{$-0.006$}\\
 & Phi-4-14B & $0.234$ & $+0.069$ & $+0.104$ & $+0.052$ & $+0.090$ & $+0.084$ & \underline{$-0.016$}\\
\addlinespace[2pt]
AdversarialQA & Falcon3-10B & $0.200$ & $+0.026$ & $+0.040$ & $+0.034$ & $+0.103$ & $+0.110$ & $+0.018$\\
 & Gemma3-12B & $0.229$ & $+0.081$ & $+0.097$ & $+0.112$ & $+0.147$ & $+0.158$ & $+0.187$\\
 & Mistral-NeMo-12B & $0.195$ & $+0.073$ & $+0.085$ & $+0.131$ & $+0.085$ & $+0.083$ & $+0.037$\\
 & OLMo2-13B & $0.210$ & $+0.066$ & $+0.088$ & $+0.114$ & $+0.102$ & $+0.115$ & $+0.040$\\
 & Phi-4-14B & $0.223$ & $+0.075$ & $+0.090$ & $+0.088$ & $+0.198$ & $+0.205$ & $+0.043$\\
\midrule
\multicolumn{2}{l}{\textit{Pairs favoring NE}} & & $24/25$ & $25/25$ & $25/25$ & $21/25$ & $22/25$ & $19/25$\\
\multicolumn{2}{l}{\textit{Mean difference}} & & $+0.060$ & $+0.101$ & $+0.110$ & $+0.080$ & $+0.084$ & $+0.062$\\
\bottomrule
\end{tabular}
\end{table}

\subsection{Archive Diagnostics}\label{app:archive_diag}

Table~\ref{tab:diagnostics_details} reports what each constructor spends and what it
keeps. \emph{Editor calls} counts every proposal issued, including those later
rejected. \emph{Retained} is what enters the archive, and \emph{distinct} removes
exact repeats of a rewrite already held, so the two differ only for a constructor
without a duplicate check. \emph{Realized type} is the share of retained rewrites
whose realized edit matches the one requested, which is one by construction wherever
the realization gate is on. \emph{Cells occupied} counts distinct cells of
$\mathcal{D}$ under realized edit types, as in Table~\ref{tab:cells}. The rewrites-by-type rows sum to the retained count for every constructor except U-ME, whose remaining 5799 rewrites have no resolvable realized type.

\begin{table}[ht]
\centering
\small
\setlength{\tabcolsep}{4.5pt}
\caption{Archive construction diagnostics, per instance and pooled over the 5000
instances. Rejection reasons are fractions of editor calls and do not sum to the
total, since a proposal may fail for more than one reason and only the principal ones
are listed.}
\label{tab:diagnostics_details}
\begin{tabular}{@{}l cccc@{}}
\toprule
& FP & SBS & U-ME & \textbf{NE}\\
\midrule
\multicolumn{5}{@{}l}{\textit{Per instance, mean over 5000 instances}}\\
Editor calls                & 15.43 & 25.00 & 24.26 & 24.21\\
Proposals passing the gates &  3.69 & 10.35 & 21.67 & 12.12\\
Retained in the archive     &  3.69 &  2.01 &  5.90 & \textbf{4.43}\\
Distinct                    &  3.69 &  2.01 &  4.94 & \textbf{4.43}\\
Duplicate rate              &  0.00 &  0.00 &  0.17 & \textbf{0.00}\\
Realized type               &  1.00 &  1.00 &  0.27 & \textbf{1.00}\\
Cells occupied, of twelve   &  0.98 &  2.00 &  2.33 & \textbf{4.43}\\
\addlinespace
\multicolumn{5}{@{}l}{\textit{Totals over the 5000 instances}}\\
Editor calls                & 77145 & 125000 & 121300 & 121043\\
Retained                    & 18439 &  10024 &  29506 & \textbf{22163}\\
\addlinespace
\multicolumn{5}{@{}l}{\textit{Retained rewrites by realized edit type}}\\
Distractor                  &     0 &   2960 &     24 & \textbf{7408}\\
Reorder                     &     0 &   2454 &    430 & \textbf{5583}\\
Style rewrite               & 18439 &   4610 &  23253 & \textbf{9172}\\
\addlinespace
\multicolumn{5}{@{}l}{\textit{Rejection reasons, as a fraction of editor calls}}\\
Answer no longer recoverable & 0.43 & 0.06 & 0.10 & 0.04\\
Requested edit not realized  & 0.02 & 0.41 & ---  & 0.43\\
Duplicate of an archived rewrite & 0.02 & 0.11 & --- & 0.24\\
Unchanged from its parent    & ---  & ---  & ---  & 0.22\\
Unchanged from the original  & 0.01 & 0.07 & 0.01 & 0.06\\
\addlinespace
Any reason                   & 0.76 & 0.59 & 0.11 & 0.50\\
\bottomrule
\end{tabular}
\end{table}
 
\begin{table}[ht]
\centering
\small
\setlength{\tabcolsep}{4pt}
\renewcommand{\arraystretch}{1.12}
\caption{Distinct cells of $\mathcal{D}$ occupied per neighborhood, by dataset. Each
entry is a mean over that dataset's 1000 instances, with the standard deviation in
brackets.\\}
\label{tab:cells_per_dataset}
\begin{tabular}{@{}l cccc@{}}
\toprule
Dataset & FP & SBS & U-ME & \textbf{NE}\\
\midrule
SQuAD v1.1        & $1.19$ \tiny{$(0.68)$} & $2.87$ \tiny{$(0.95)$} & $2.44$ \tiny{$(1.05)$} & $\mathbf{5.40}$ \tiny{$(2.26)$}\\
Quoref            & $0.99$ \tiny{$(0.55)$} & $1.85$ \tiny{$(1.04)$} & $2.56$ \tiny{$(1.03)$} & $\mathbf{4.67}$ \tiny{$(2.06)$}\\
AdversarialQA     & $1.04$ \tiny{$(0.66)$} & $2.87$ \tiny{$(0.90)$} & $2.38$ \tiny{$(1.10)$} & $\mathbf{5.69}$ \tiny{$(2.06)$}\\
Natural Questions & $0.92$ \tiny{$(0.69)$} & $1.17$ \tiny{$(0.85)$} & $2.17$ \tiny{$(1.10)$} & $\mathbf{3.33}$ \tiny{$(2.25)$}\\
NewsQA            & $0.76$ \tiny{$(0.70)$} & $1.27$ \tiny{$(1.25)$} & $2.12$ \tiny{$(1.06)$} & $\mathbf{3.07}$ \tiny{$(2.08)$}\\
\midrule
Macro             & $0.98$ \tiny{$(0.66)$} & $2.00$ \tiny{$(1.01)$} & $2.33$ \tiny{$(1.07)$} & $\mathbf{4.43}$ \tiny{$(2.15)$}\\
\bottomrule
\end{tabular}
\end{table}

\begin{figure}[ht]
\centering
\includegraphics[width=\linewidth]{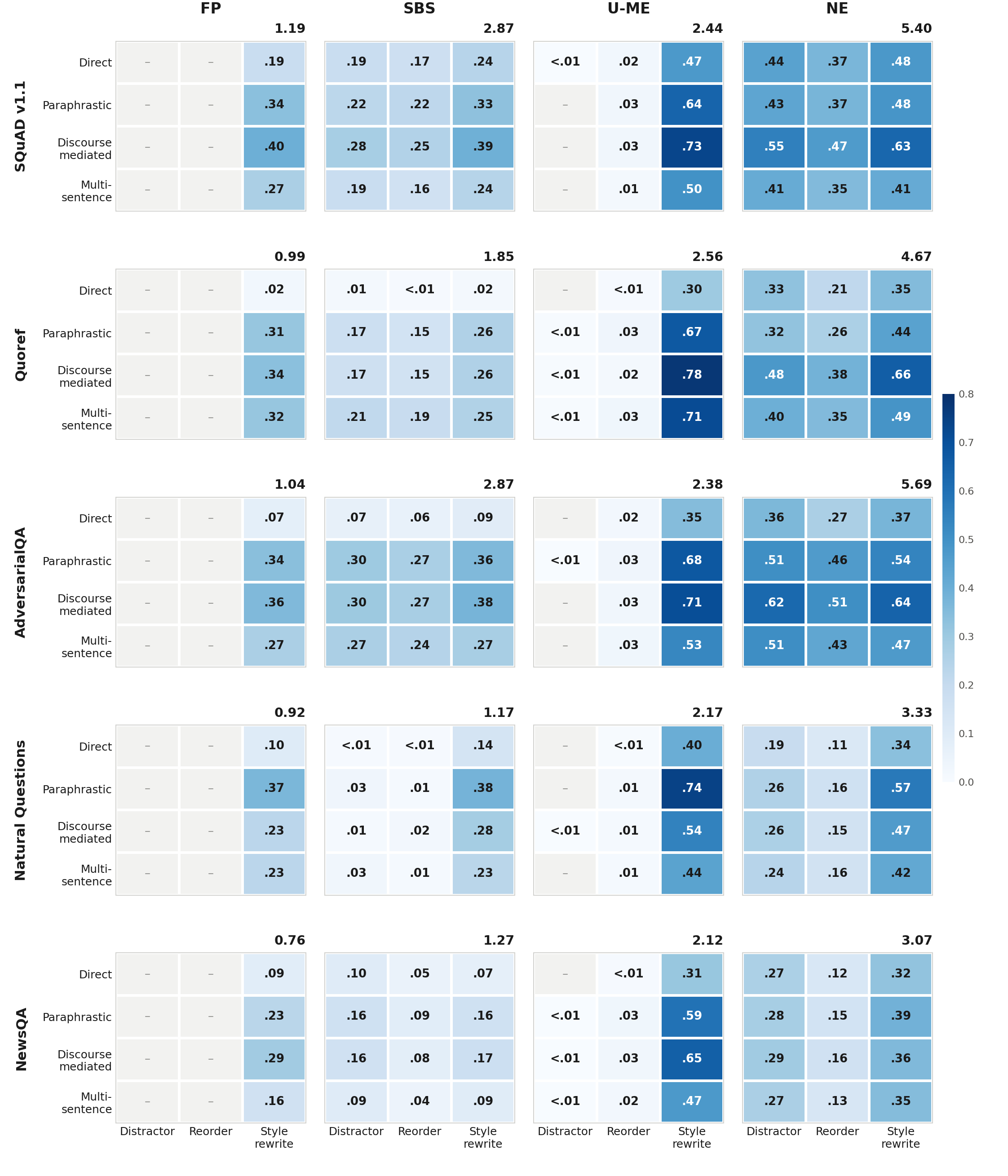}
\caption{Descriptor space across datasets. For each cell, the proportion of the $5000$ instances whose neighborhood places at least one rewrite in that cell. }
\label{fig:archive_per_dataset}
\end{figure}

A cell is left empty when the region holds fewer than one hundred rewrites, which is too few to estimate a rate from, so an empty cell here does not carry the meaning it does in Figure~\ref{fig:archives_main}, where it marks a region the constructor never reached. U-ME has no distractor column for this reason: across all datasets and QA models its realized distractors amount to $47$ scored rewrites against $55{,}187$ style rewrites. That the constructor produces so few is a separate finding, reported in Table~\ref{tab:diagnostics_details}, and not a statement about how hard those regions are.

\subsection{Detection With and Without a Minimum Neighborhood Size}\label{app:as_deployed}
 
This section reports the detection results behind Table~\ref{tab:main} in full, under two
instance populations and three cuts of the data. The first population uses every instance a
method scores, whatever the size of the neighborhood it has. The second restricts to instances
holding at least two rewrites or samples, which is the smallest count at which semantic entropy
is defined, since a single prediction admits no partition to measure. All values are
macro-averaged over the 25 dataset--model pairs.
 
Columns are as defined in Appendix~\ref{app:evaluation_protocol}.
 
\paragraph{Overall.}
Tables~\ref{tab:as_deployed_overall} and~\ref{tab:mink2_overall} give the comparison across all
pairs. Imposing the minimum separates the two kinds of score cleanly. All three constructors
gain, NE by 0.023 AUROC and 0.022 lift, free paraphrasing by 0.016 and the untargeted search by
0.003, while all three scores read from the original context are unchanged, at 0.676 for
predictive entropy, 0.674 for the length-normalised likelihood and 0.652 for decoder semantic
entropy. The restriction is definitional rather than tuned, and it moves exactly the methods
whose score needs more than one prediction to be defined.
 
\begin{table}[ht]
\centering
\caption{Failure detection on the \emph{full archives}, with every method scoring each instance
from all of the rewrites or samples it has and no minimum imposed.}
\label{tab:as_deployed_overall}
\small
\setlength{\tabcolsep}{4pt}
\begin{tabular}{@{}llcccccccc@{}}
\toprule
& Method & AUROC & AUPRC & Lift & Base & $n$ & Cov. & \textsc{sd} & \textsc{min}--\textsc{max} \\
\midrule
1 & \textbf{NE} & 0.702 & 0.546 & 0.189 & 0.357 & 23739 & 0.950 & 0.053 & 0.602--0.776 \\
2 & PE & 0.676 & 0.488 & 0.131 & 0.357 & 23739 & 0.950 & 0.047 & 0.582--0.772 \\
3 & NLL & 0.674 & 0.492 & 0.131 & 0.361 & 24984 & 0.999 & 0.050 & 0.563--0.761 \\
4 & U-ME & 0.658 & 0.496 & 0.146 & 0.351 & 24060 & 0.962 & 0.020 & 0.624--0.686 \\
5 & DSE & 0.652 & 0.498 & 0.149 & 0.349 & 21920 & 0.877 & 0.066 & 0.518--0.734 \\
6 & FP & 0.605 & 0.421 & 0.097 & 0.323 & 19569 & 0.783 & 0.031 & 0.546--0.651 \\
\bottomrule
\end{tabular}
\end{table}
 
\begin{table}[ht]
\centering
\caption{Failure detection restricted to instances holding at least two rewrites or samples.
Semantic entropy is identically zero on a single prediction, since there is no partition to
measure, so two is the smallest count at which the score is defined. NE, PE and DSE sit on an
identical population of 21920 instances.}
\label{tab:mink2_overall}
\small
\setlength{\tabcolsep}{4pt}
\begin{tabular}{@{}llcccccccc@{}}
\toprule
& Method & AUROC & AUPRC & Lift & Base & $n$ & Cov. & \textsc{sd} & \textsc{min}--\textsc{max} \\
\midrule
1 & \textbf{NE} & 0.726 & 0.561 & 0.211 & 0.349 & 21920 & 0.877 & 0.038 & 0.647--0.786 \\
2 & PE & 0.676 & 0.480 & 0.131 & 0.349 & 21920 & 0.877 & 0.046 & 0.580--0.765 \\
3 & NLL & 0.674 & 0.492 & 0.131 & 0.361 & 24984 & 0.999 & 0.050 & 0.563--0.761 \\
4 & U-ME & 0.661 & 0.497 & 0.148 & 0.349 & 23920 & 0.957 & 0.019 & 0.627--0.689 \\
5 & DSE & 0.652 & 0.498 & 0.149 & 0.349 & 21920 & 0.877 & 0.066 & 0.518--0.734 \\
6 & FP & 0.621 & 0.428 & 0.113 & 0.315 & 17864 & 0.715 & 0.032 & 0.563--0.679 \\
\bottomrule
\end{tabular}
\end{table}
 
\paragraph{By dataset.}
Tables~\ref{tab:as_deployed_per_dataset} and~\ref{tab:mink2_per_dataset} break the same
comparison down by dataset. NE leads on SQuAD v1.1, Quoref and AdversarialQA under both
populations. The two remaining datasets behave differently. On NewsQA the original-context
scores lead on the full archives, and the minimum closes most of that gap, taking NE from 0.647
to 0.687 against 0.691 for the length-normalised likelihood. Natural Questions is the one
dataset where the likelihood scores lead clearly under both populations.
Tables~\ref{tab:lowN_counts} and~\ref{tab:lowN_effect} trace both cases to the share of
instances carrying fewer than three rewrites.
 
\begin{table}[ht]
\centering
\caption{Failure detection per dataset on the full archives, macro-averaged over the five QA
models. Best value per row in bold.}
\label{tab:as_deployed_per_dataset}
\small
\begin{tabular}{@{}lcccccc@{}}
\toprule
Dataset & NE & U-ME & FP & DSE & PE & NLL \\
\midrule
\multicolumn{7}{@{}l}{\emph{AUROC}}\\
SQuAD v1.1 & \textbf{0.742} & 0.671 & 0.616 & 0.676 & 0.673 & 0.666 \\
Quoref & \textbf{0.740} & 0.674 & 0.619 & 0.614 & 0.663 & 0.662 \\
NewsQA & 0.647 & 0.636 & 0.581 & 0.668 & 0.686 & \textbf{0.691} \\
Natural Questions & 0.638 & 0.652 & 0.594 & 0.643 & 0.728 & \textbf{0.731} \\
AdversarialQA & \textbf{0.741} & 0.659 & 0.613 & 0.658 & 0.628 & 0.620 \\
\midrule
\multicolumn{7}{@{}l}{\emph{AUPRC}}\\
SQuAD v1.1 & \textbf{0.480} & 0.383 & 0.299 & 0.413 & 0.331 & 0.325 \\
Quoref & \textbf{0.510} & 0.430 & 0.389 & 0.392 & 0.420 & 0.422 \\
NewsQA & 0.656 & 0.659 & 0.567 & 0.669 & 0.680 & \textbf{0.694} \\
Natural Questions & 0.432 & 0.430 & 0.320 & 0.426 & 0.480 & \textbf{0.495} \\
AdversarialQA & \textbf{0.654} & 0.578 & 0.530 & 0.591 & 0.529 & 0.523 \\
\bottomrule
\end{tabular}
\end{table}
 
\begin{table}[ht]
\centering
\caption{Failure detection per dataset at two or more rewrites, macro-averaged over the five QA
models. Best value per row in bold.}
\label{tab:mink2_per_dataset}
\small
\begin{tabular}{@{}lcccccc@{}}
\toprule
Dataset & NE & U-ME & FP & DSE & PE & NLL \\
\midrule
\multicolumn{7}{@{}l}{\emph{AUROC}}\\
SQuAD v1.1 & \textbf{0.752} & 0.672 & 0.628 & 0.676 & 0.678 & 0.666 \\
Quoref & \textbf{0.756} & 0.674 & 0.626 & 0.614 & 0.665 & 0.662 \\
NewsQA & 0.687 & 0.641 & 0.611 & 0.668 & 0.687 & \textbf{0.691} \\
Natural Questions & 0.687 & 0.654 & 0.617 & 0.643 & 0.722 & \textbf{0.731} \\
AdversarialQA & \textbf{0.746} & 0.662 & 0.623 & 0.658 & 0.628 & 0.620 \\
\midrule
\multicolumn{7}{@{}l}{\emph{AUPRC}}\\
SQuAD v1.1 & \textbf{0.486} & 0.383 & 0.302 & 0.413 & 0.331 & 0.325 \\
Quoref & \textbf{0.524} & 0.430 & 0.397 & 0.392 & 0.424 & 0.422 \\
NewsQA & 0.677 & 0.660 & 0.571 & 0.669 & 0.670 & \textbf{0.694} \\
Natural Questions & 0.459 & 0.431 & 0.336 & 0.426 & 0.446 & \textbf{0.495} \\
AdversarialQA & \textbf{0.657} & 0.579 & 0.533 & 0.591 & 0.530 & 0.523 \\
\bottomrule
\end{tabular}
\end{table}
 
\paragraph{By QA model.}
Tables~\ref{tab:as_deployed_per_model} and~\ref{tab:mink2_per_model} give the same cut by QA
model. Detection varies far less across models than across datasets, which is what the
construction procedure predicts: a neighborhood is built from the context alone and never
queries the answering model, so the share of instances holding fewer than three rewrites is
18.5 to 18.6 percent for every model here, while across datasets it ranges from 5.6 to 37.7
percent. NE leads on four of the five models on both measures on the full archives, trailing
only on Mistral-Nemo-12B, and on all five under the minimum. Decoder semantic entropy falls
close to chance on Gemma3-12B at 0.538 AUROC, where resampling the decoder leaves the answer
unchanged while rewriting the context still separates failures.
 
\begin{table}[ht]
\centering
\caption{Failure detection per QA model on the full archives, macro-averaged over the five
datasets. Best value per row in bold.}
\label{tab:as_deployed_per_model}
\small
\begin{tabular}{@{}lcccccc@{}}
\toprule
QA model & NE & U-ME & FP & DSE & PE & NLL \\
\midrule
\multicolumn{7}{@{}l}{\emph{AUROC}}\\
Gemma3-12B & \textbf{0.698} & 0.650 & 0.588 & 0.538 & 0.648 & 0.646 \\
OLMo2-13B-Instruct & \textbf{0.708} & 0.661 & 0.594 & 0.685 & 0.693 & 0.693 \\
Phi4-14B & \textbf{0.713} & 0.668 & 0.641 & 0.707 & 0.656 & 0.654 \\
Mistral-Nemo-12B & 0.692 & 0.648 & 0.578 & 0.670 & \textbf{0.706} & 0.705 \\
Falcon3-10B-Instruct & \textbf{0.698} & 0.665 & 0.621 & 0.660 & 0.676 & 0.672 \\
\midrule
\multicolumn{7}{@{}l}{\emph{AUPRC}}\\
Gemma3-12B & \textbf{0.470} & 0.405 & 0.330 & 0.313 & 0.375 & 0.381 \\
OLMo2-13B-Instruct & \textbf{0.576} & 0.520 & 0.421 & 0.550 & 0.528 & 0.535 \\
Phi4-14B & \textbf{0.574} & 0.512 & 0.471 & 0.570 & 0.467 & 0.464 \\
Mistral-Nemo-12B & 0.572 & 0.527 & 0.427 & 0.547 & 0.575 & \textbf{0.581} \\
Falcon3-10B-Instruct & \textbf{0.539} & 0.516 & 0.455 & 0.511 & 0.495 & 0.498 \\
\bottomrule
\end{tabular}
\end{table}
 
\begin{table}[ht]
\centering
\caption{Failure detection per QA model at two or more rewrites, macro-averaged over the five
datasets. Best value per row in bold.}
\label{tab:mink2_per_model}
\small
\begin{tabular}{@{}lcccccc@{}}
\toprule
QA model & NE & U-ME & FP & DSE & PE & NLL \\
\midrule
\multicolumn{7}{@{}l}{\emph{AUROC}}\\
Gemma3-12B & \textbf{0.721} & 0.652 & 0.601 & 0.538 & 0.647 & 0.646 \\
OLMo2-13B-Instruct & \textbf{0.730} & 0.664 & 0.609 & 0.685 & 0.696 & 0.693 \\
Phi4-14B & \textbf{0.738} & 0.670 & 0.662 & 0.707 & 0.659 & 0.654 \\
Mistral-Nemo-12B & \textbf{0.712} & 0.650 & 0.589 & 0.670 & 0.704 & 0.705 \\
Falcon3-10B-Instruct & \textbf{0.726} & 0.668 & 0.644 & 0.660 & 0.674 & 0.672 \\
\midrule
\multicolumn{7}{@{}l}{\emph{AUPRC}}\\
Gemma3-12B & \textbf{0.485} & 0.406 & 0.334 & 0.313 & 0.367 & 0.381 \\
OLMo2-13B-Instruct & \textbf{0.588} & 0.521 & 0.425 & 0.550 & 0.520 & 0.535 \\
Phi4-14B & \textbf{0.592} & 0.512 & 0.487 & 0.570 & 0.462 & 0.464 \\
Mistral-Nemo-12B & \textbf{0.584} & 0.527 & 0.430 & 0.547 & 0.569 & 0.581 \\
Falcon3-10B-Instruct & \textbf{0.553} & 0.517 & 0.463 & 0.511 & 0.484 & 0.498 \\
\bottomrule
\end{tabular}
\end{table}
 
\paragraph{Pair by pair.}
Table~\ref{tab:mink2_headtohead} compares NE against each other method one pair at a time, on
the instances the two share. Against the other two constructors the advantage is uniform, in
all 25 pairs on AUROC. Against the three scores read from the original context the mean margin
is comparable but the advantage is less uniform, holding in 18 to 21 pairs of 25.
 
\begin{table}[ht]
\centering
\caption{Per-pair comparison of NE against each other method at two or more rewrites, over the
25 dataset--model pairs. Counts are pairs won by NE against pairs lost, on the instances the
two methods share, with the mean difference alongside.}
\label{tab:mink2_headtohead}
\small
\begin{tabular}{@{}lcccc@{}}
\toprule
& \multicolumn{2}{c}{AUROC} & \multicolumn{2}{c}{AUPRC} \\
\cmidrule(lr){2-3}\cmidrule(lr){4-5}
NE versus & Pairs won & Mean $\Delta$ & Pairs won & Mean $\Delta$ \\
\midrule
U-ME & 25/25 & $+0.065$ & 24/25 & $+0.064$ \\
FP & 25/25 & $+0.104$ & 25/25 & $+0.133$ \\
DSE & 21/25 & $+0.074$ & 19/25 & $+0.062$ \\
PE & 18/25 & $+0.050$ & 21/25 & $+0.080$ \\
NLL & 18/25 & $+0.052$ & 19/25 & $+0.069$ \\
\bottomrule
\end{tabular}
\end{table}
 
\paragraph{Where the difference between the two populations comes from.}
The last two tables account for the gap. Table~\ref{tab:lowN_counts} reports how many rewrites
each method holds per instance and how often those instances fail. NE's low-rewrite instances
are markedly harder than the rest, failing at 0.462 at one rewrite and 0.460 at two against
0.330 at three or more, and there are 4401 of them in its 23739 scored instances. Semantic
entropy over a single rewrite is identically zero, so all 1819 one-rewrite instances receive
the same minimum score, which places the hardest instances at the most confident end of the
ranking. Table~\ref{tab:lowN_effect} shows the consequence for NE by dataset. The cost of
including those instances tracks their share: it is largest on NewsQA and Natural Questions,
where close to a third of instances carry fewer than three rewrites, and negligible on
AdversarialQA and SQuAD v1.1, where few do.
 
\begin{table}[ht]
\centering
\caption{Rewrites available per instance over each method's full archive, with the failure rate
at each count. Unscored counts instances the method produces no score for at all, against 1000
per dataset--model pair. Predictive entropy is defined at a single sample and is listed for its
counts only, which match NE's because it is computed on the same instances.}
\label{tab:lowN_counts}
\small
\setlength{\tabcolsep}{4pt}
\begin{tabular}{@{}lrrrrrrccc@{}}
\toprule
& \multicolumn{4}{c}{Instances} & \multicolumn{2}{c}{Share} & \multicolumn{3}{c}{Failure rate} \\
\cmidrule(lr){2-5}\cmidrule(lr){6-7}\cmidrule(lr){8-10}
Method & Scored & $N{=}1$ & $N{=}2$ & $N{\geq}3$ & $N{<}3$ & Unsc. & $N{=}1$ & $N{=}2$ & $N{\geq}3$ \\
\midrule
\textbf{NE} & 23739 & 1819 & 2582 & 19338 & 18.5\% & 1261 & 0.462 & 0.460 & 0.330 \\
U-ME & 24060 & 140 & 505 & 23415 & 2.7\% & 940 & 0.679 & 0.677 & 0.341 \\
FP & 19569 & 1705 & 1915 & 15949 & 18.5\% & 5431 & 0.407 & 0.405 & 0.293 \\
DSE & 21920 & 0 & 2582 & 19338 & 11.8\% & 3080 & -- & 0.460 & 0.330 \\
PE & 23739 & 1819 & 2582 & 19338 & 18.5\% & 1261 & 0.462 & 0.460 & 0.330 \\
NLL & 24984 & -- & -- & -- & -- & 16 & -- & -- & -- \\
\bottomrule
\end{tabular}
\end{table}
 
\begin{table}[ht]
\centering
\caption{\textbf{Neighborhood elites (NE) only.} Detection per dataset on the full archive
against the same computation restricted to instances holding at least three rewrites,
macro-averaged over the five QA models. $\Delta$ is the full-archive value minus the restricted
one.}
\label{tab:lowN_effect}
\small
\setlength{\tabcolsep}{5pt}
\begin{tabular}{@{}lrr ccr ccr@{}}
\toprule
& & & \multicolumn{3}{c}{AUROC} & \multicolumn{3}{c}{AUPRC} \\
\cmidrule(lr){4-6}\cmidrule(lr){7-9}
Dataset & Scored & $N{<}3$ & Full & $N{\geq}3$ & $\Delta$ & Full & $N{\geq}3$ & $\Delta$ \\
\midrule
SQuAD v1.1 & 4980 & 8.2\% & 0.742 & 0.759 & $-0.017$ & 0.480 & 0.485 & $-0.006$ \\
Quoref & 4943 & 14.1\% & 0.740 & 0.767 & $-0.027$ & 0.510 & 0.527 & $-0.017$ \\
NewsQA & 4438 & 37.7\% & 0.647 & 0.712 & $-0.065$ & 0.656 & 0.682 & $-0.025$ \\
Natural Questions & 4385 & 30.7\% & 0.638 & 0.709 & $-0.071$ & 0.432 & 0.468 & $-0.036$ \\
AdversarialQA & 4993 & 5.6\% & 0.741 & 0.748 & $-0.007$ & 0.654 & 0.656 & $-0.002$ \\
\bottomrule
\end{tabular}
\end{table}

\subsection{Sensitivity to Editor Size}\label{app:editor_size}

This section reports NE with three editors, Qwen3-8B, 14B and 32B. Columns are as defined in
Appendix~\ref{app:evaluation_protocol}. Three populations appear and are not interchangeable. The
tables use each editor's own full archive, or that archive restricted to instances holding at
least two rewrites or samples. Figure~\ref{fig:editor} uses a third, the 21031 instances for
which NE retains at least two rewrites at all three editor sizes, so the differences it reports
are not the differences between the table values.

\subsubsection{Detection Across the Three Editor Sizes}\label{app:editor_size_overview}

\begin{figure}[ht]
\centering
\includegraphics[width=\linewidth]{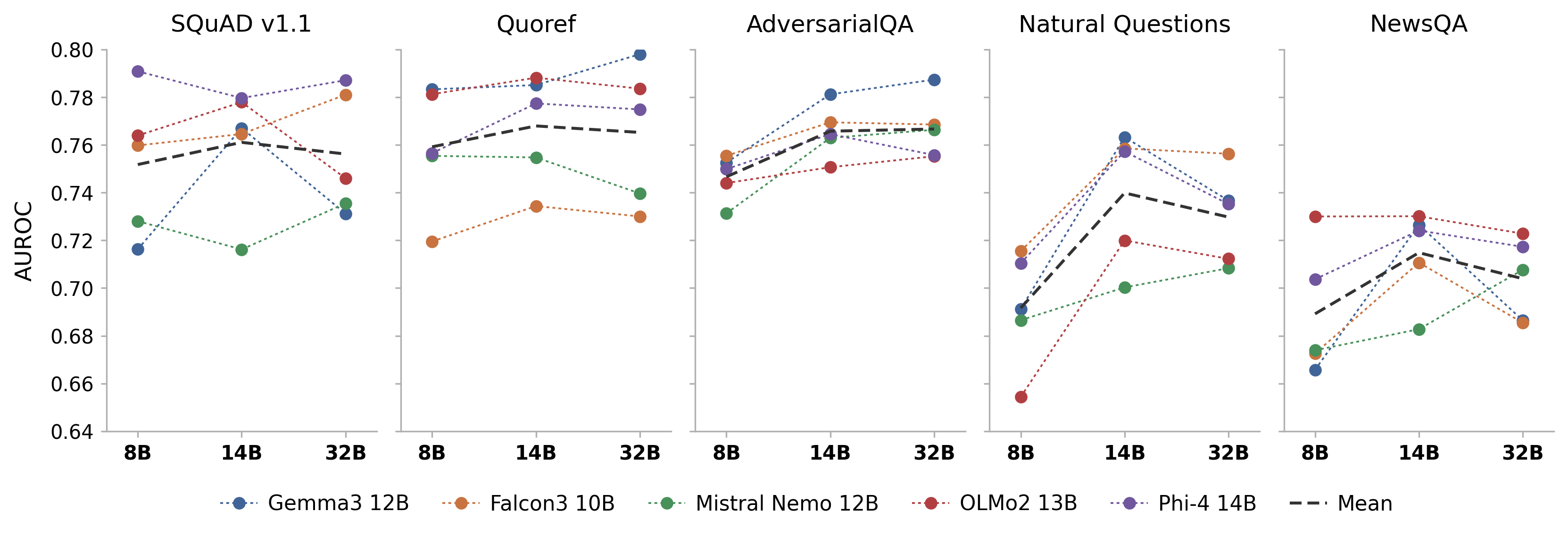}
\caption{NE AUROC at three editor sizes, on the 21031 instances for which NE retains at least
two rewrites at all three.}
\label{fig:editor}
\end{figure}

\paragraph{Detection improves from the 8B editor to the 14B and does not improve further at
32B.}
Figure~\ref{fig:editor} reports NE at three editor sizes on the 21031 instances for which it
retains at least two rewrites at all three. From 8B to 14B, mean AUROC rises by 0.0222 [0.0135,
0.0316] and AUPRC lift by 0.0294 [0.0178, 0.0409], the latter in 22 of 25 pairs. From 14B to
32B, the changes are $-0.0055$ [$-0.0121$, 0.0010] AUROC in 9 of 25 pairs and $-0.0095$
[$-0.0186$, $-0.0008$] AUPRC lift in 11 of 25. Neighborhoods grow throughout, from 4.97
rewrites per instance at 8B to 5.91 at 14B and 6.46 at 32B. The first step is uneven across
datasets: Natural Questions gains 0.0483 AUROC and NewsQA 0.0256, against 0.0191 for
AdversarialQA, 0.0093 for SQuAD v1.1 and 0.0088 for Quoref. In the second step four of five
datasets lose, by 0.0101 on Natural Questions, 0.0109 on NewsQA, 0.0049 on SQuAD v1.1 and
0.0027 on Quoref, while AdversarialQA gains 0.0009.

\subsubsection{The 14B Editor}\label{app:editor_size_14b}

U-ME has no 14B run and is absent from every table in this part.

\paragraph{Overall.}
Tables~\ref{tab:e14_overall} and~\ref{tab:e14k2_overall} give the comparison across all pairs,
on the full archives and restricted to instances holding at least two rewrites or samples.

\begin{table}[ht]
\centering
\caption{Failure detection with the Qwen3-14B editor on the full archives, macro-averaged over
the 25 dataset--model pairs.}
\label{tab:e14_overall}
\small
\setlength{\tabcolsep}{4pt}
\begin{tabular}{@{}llcccccccc@{}}
\toprule
& Method & AUROC & AUPRC & Lift & Base & $n$ & Cov. & \textsc{sd} & \textsc{min}--\textsc{max} \\
\midrule
1 & \textbf{NE} & 0.731 & 0.579 & 0.222 & 0.357 & 24133 & 0.965 & 0.038 & 0.653--0.784 \\
2 & NLL & 0.674 & 0.491 & 0.130 & 0.361 & 24993 & 1.000 & 0.050 & 0.563--0.760 \\
3 & PE & 0.673 & 0.486 & 0.129 & 0.357 & 24133 & 0.965 & 0.047 & 0.584--0.770 \\
4 & DSE & 0.658 & 0.511 & 0.158 & 0.353 & 23224 & 0.929 & 0.066 & 0.527--0.753 \\
5 & FP & 0.638 & 0.455 & 0.127 & 0.328 & 21584 & 0.863 & 0.031 & 0.564--0.685 \\
\bottomrule
\end{tabular}
\end{table}

\begin{table}[ht]
\centering
\caption{Failure detection with the Qwen3-14B editor, restricted to instances holding at least
two rewrites or samples. NE, PE and DSE sit on an identical population of 23224 instances.}
\label{tab:e14k2_overall}
\small
\setlength{\tabcolsep}{4pt}
\begin{tabular}{@{}llcccccccc@{}}
\toprule
& Method & AUROC & AUPRC & Lift & Base & $n$ & Cov. & \textsc{sd} & \textsc{min}--\textsc{max} \\
\midrule
1 & \textbf{NE} & 0.744 & 0.588 & 0.235 & 0.353 & 23224 & 0.929 & 0.030 & 0.676--0.786 \\
2 & NLL & 0.674 & 0.491 & 0.130 & 0.361 & 24993 & 1.000 & 0.050 & 0.563--0.760 \\
3 & PE & 0.673 & 0.483 & 0.129 & 0.353 & 23224 & 0.929 & 0.047 & 0.584--0.768 \\
4 & DSE & 0.658 & 0.511 & 0.158 & 0.353 & 23224 & 0.929 & 0.066 & 0.527--0.753 \\
5 & FP & 0.649 & 0.460 & 0.137 & 0.323 & 20600 & 0.824 & 0.031 & 0.568--0.693 \\
\bottomrule
\end{tabular}
\end{table}

\paragraph{By dataset.}
Tables~\ref{tab:e14_per_dataset} and~\ref{tab:e14k2_per_dataset} break the same comparison
down by dataset, macro-averaged over the five QA models.

\begin{table}[ht]
\centering
\caption{Failure detection per dataset with the Qwen3-14B editor on the full archives,
macro-averaged over the five QA models. Best per row in bold.}
\label{tab:e14_per_dataset}
\small
\begin{tabular}{@{}lccccc@{}}
\toprule
Dataset & NE & FP & DSE & PE & NLL \\
\midrule
\multicolumn{6}{@{}l}{\emph{AUROC}}\\
SQuAD v1.1 & \textbf{0.750} & 0.649 & 0.673 & 0.670 & 0.666 \\
Quoref & \textbf{0.749} & 0.651 & 0.616 & 0.660 & 0.662 \\
NewsQA & 0.677 & 0.619 & 0.673 & 0.683 & \textbf{0.691} \\
Natural Questions & 0.713 & 0.623 & 0.666 & 0.728 & \textbf{0.730} \\
AdversarialQA & \textbf{0.766} & 0.647 & 0.660 & 0.625 & 0.620 \\
\midrule
\multicolumn{6}{@{}l}{\emph{AUPRC}}\\
SQuAD v1.1 & \textbf{0.487} & 0.337 & 0.414 & 0.329 & 0.324 \\
Quoref & \textbf{0.512} & 0.401 & 0.396 & 0.420 & 0.422 \\
NewsQA & 0.687 & 0.611 & 0.675 & 0.671 & \textbf{0.695} \\
Natural Questions & \textbf{0.526} & 0.365 & 0.477 & 0.482 & 0.493 \\
AdversarialQA & \textbf{0.684} & 0.560 & 0.595 & 0.528 & 0.523 \\
\bottomrule
\end{tabular}
\end{table}

\begin{table}[ht]
\centering
\caption{Failure detection per dataset with the Qwen3-14B editor at two or more rewrites,
macro-averaged over the five QA models. Best per row in bold.}
\label{tab:e14k2_per_dataset}
\small
\begin{tabular}{@{}lccccc@{}}
\toprule
Dataset & NE & FP & DSE & PE & NLL \\
\midrule
\multicolumn{6}{@{}l}{\emph{AUROC}}\\
SQuAD v1.1 & \textbf{0.758} & 0.659 & 0.673 & 0.672 & 0.666 \\
Quoref & \textbf{0.760} & 0.658 & 0.616 & 0.659 & 0.662 \\
NewsQA & \textbf{0.709} & 0.643 & 0.673 & 0.684 & 0.691 \\
Natural Questions & 0.729 & 0.631 & 0.666 & 0.724 & \textbf{0.730} \\
AdversarialQA & \textbf{0.766} & 0.657 & 0.660 & 0.626 & 0.620 \\
\midrule
\multicolumn{6}{@{}l}{\emph{AUPRC}}\\
SQuAD v1.1 & \textbf{0.493} & 0.340 & 0.414 & 0.327 & 0.324 \\
Quoref & \textbf{0.519} & 0.406 & 0.396 & 0.417 & 0.422 \\
NewsQA & \textbf{0.703} & 0.621 & 0.675 & 0.666 & 0.695 \\
Natural Questions & \textbf{0.540} & 0.372 & 0.477 & 0.475 & 0.493 \\
AdversarialQA & \textbf{0.685} & 0.563 & 0.595 & 0.529 & 0.523 \\
\bottomrule
\end{tabular}
\end{table}

\paragraph{By QA model.}
Tables~\ref{tab:e14_per_model} and~\ref{tab:e14k2_per_model} give the same cut by QA model,
macro-averaged over the five datasets.

\begin{table}[ht]
\centering
\caption{Failure detection per QA model with the Qwen3-14B editor on the full archives,
macro-averaged over the five datasets. Best per row in bold.}
\label{tab:e14_per_model}
\small
\begin{tabular}{@{}lccccc@{}}
\toprule
QA model & NE & FP & DSE & PE & NLL \\
\midrule
\multicolumn{6}{@{}l}{\emph{AUROC}}\\
Gemma3-12B & \textbf{0.745} & 0.634 & 0.543 & 0.648 & 0.646 \\
OLMo2-13B-Instruct & \textbf{0.738} & 0.639 & 0.691 & 0.689 & 0.693 \\
Phi4-14B & \textbf{0.736} & 0.663 & 0.711 & 0.658 & 0.653 \\
Mistral-Nemo-12B & \textbf{0.708} & 0.607 & 0.678 & 0.704 & 0.705 \\
Falcon3-10B-Instruct & \textbf{0.727} & 0.645 & 0.664 & 0.668 & 0.671 \\
\midrule
\multicolumn{6}{@{}l}{\emph{AUPRC}}\\
Gemma3-12B & \textbf{0.517} & 0.377 & 0.330 & 0.378 & 0.381 \\
OLMo2-13B-Instruct & \textbf{0.614} & 0.482 & 0.552 & 0.522 & 0.536 \\
Phi4-14B & \textbf{0.592} & 0.485 & 0.583 & 0.472 & 0.463 \\
Mistral-Nemo-12B & \textbf{0.594} & 0.458 & 0.556 & 0.571 & 0.580 \\
Falcon3-10B-Instruct & \textbf{0.578} & 0.473 & 0.536 & 0.486 & 0.496 \\
\bottomrule
\end{tabular}
\end{table}

\begin{table}[ht]
\centering
\caption{Failure detection per QA model with the Qwen3-14B editor at two or more rewrites,
macro-averaged over the five datasets. Best per row in bold.}
\label{tab:e14k2_per_model}
\small
\begin{tabular}{@{}lccccc@{}}
\toprule
QA model & NE & FP & DSE & PE & NLL \\
\midrule
\multicolumn{6}{@{}l}{\emph{AUROC}}\\
Gemma3-12B & \textbf{0.757} & 0.643 & 0.543 & 0.647 & 0.646 \\
OLMo2-13B-Instruct & \textbf{0.753} & 0.653 & 0.691 & 0.690 & 0.693 \\
Phi4-14B & \textbf{0.750} & 0.674 & 0.711 & 0.657 & 0.653 \\
Mistral-Nemo-12B & \textbf{0.719} & 0.617 & 0.678 & 0.705 & 0.705 \\
Falcon3-10B-Instruct & \textbf{0.743} & 0.660 & 0.664 & 0.666 & 0.671 \\
\midrule
\multicolumn{6}{@{}l}{\emph{AUPRC}}\\
Gemma3-12B & \textbf{0.527} & 0.382 & 0.330 & 0.377 & 0.381 \\
OLMo2-13B-Instruct & \textbf{0.624} & 0.488 & 0.552 & 0.521 & 0.536 \\
Phi4-14B & \textbf{0.601} & 0.493 & 0.583 & 0.467 & 0.463 \\
Mistral-Nemo-12B & \textbf{0.601} & 0.460 & 0.556 & 0.569 & 0.580 \\
Falcon3-10B-Instruct & \textbf{0.586} & 0.479 & 0.536 & 0.479 & 0.496 \\
\bottomrule
\end{tabular}
\end{table}

\paragraph{Pair by pair.}
Tables~\ref{tab:e14_headtohead} and~\ref{tab:e14k2_headtohead} compare NE against each other
method one pair at a time, on the instances the two share.

\begin{table}[ht]
\centering
\caption{Per-pair comparison of NE against each other method with the Qwen3-14B editor on the
full archives, over the 25 dataset--model pairs. Counts are pairs won by NE.}
\label{tab:e14_headtohead}
\small
\begin{tabular}{@{}lcccc@{}}
\toprule
& \multicolumn{2}{c}{AUROC} & \multicolumn{2}{c}{AUPRC} \\
\cmidrule(lr){2-3}\cmidrule(lr){4-5}
NE versus & Pairs won & Mean $\Delta$ & Pairs won & Mean $\Delta$ \\
\midrule
FP & 25/25 & $+0.093$ & 25/25 & $+0.124$ \\
DSE & 19/25 & $+0.073$ & 22/25 & $+0.068$ \\
PE & 18/25 & $+0.057$ & 21/25 & $+0.093$ \\
NLL & 17/25 & $+0.057$ & 20/25 & $+0.088$ \\
\bottomrule
\end{tabular}
\end{table}

\begin{table}[ht]
\centering
\caption{Per-pair comparison of NE against each other method with the Qwen3-14B editor at two
or more rewrites, over the 25 dataset--model pairs. Counts are pairs won by NE.}
\label{tab:e14k2_headtohead}
\small
\begin{tabular}{@{}lcccc@{}}
\toprule
& \multicolumn{2}{c}{AUROC} & \multicolumn{2}{c}{AUPRC} \\
\cmidrule(lr){2-3}\cmidrule(lr){4-5}
NE versus & Pairs won & Mean $\Delta$ & Pairs won & Mean $\Delta$ \\
\midrule
FP & 25/25 & $+0.095$ & 25/25 & $+0.128$ \\
DSE & 23/25 & $+0.087$ & 22/25 & $+0.077$ \\
PE & 21/25 & $+0.072$ & 23/25 & $+0.105$ \\
NLL & 21/25 & $+0.071$ & 22/25 & $+0.097$ \\
\bottomrule
\end{tabular}
\end{table}

\subsubsection{The 32B Editor}\label{app:editor_size_32b}

FP and U-ME have no 32B run, so this part compares NE against the three scores read from the
original context and carries no constructor comparison. No minimum-rewrite tables are
available for this editor.

\paragraph{Overall.}
Table~\ref{tab:e32_overall} gives the comparison across all pairs.

\begin{table}[ht]
\centering
\caption{Failure detection with the Qwen3-32B editor on the full archives, macro-averaged over
the 25 dataset--model pairs. FP and U-ME have no 32B run.}
\label{tab:e32_overall}
\small
\setlength{\tabcolsep}{4pt}
\begin{tabular}{@{}llcccccccc@{}}
\toprule
& Method & AUROC & AUPRC & Lift & Base & $n$ & Cov. & \textsc{sd} & \textsc{min}--\textsc{max} \\
\midrule
1 & \textbf{NE} & 0.724 & 0.568 & 0.211 & 0.356 & 24475 & 0.979 & 0.045 & 0.634--0.790 \\
2 & PE & 0.674 & 0.486 & 0.129 & 0.356 & 24475 & 0.979 & 0.046 & 0.582--0.762 \\
3 & NLL & 0.674 & 0.491 & 0.130 & 0.361 & 25000 & 1.000 & 0.049 & 0.563--0.762 \\
4 & DSE & 0.662 & 0.512 & 0.162 & 0.350 & 23865 & 0.955 & 0.068 & 0.523--0.743 \\
\bottomrule
\end{tabular}
\end{table}

\paragraph{By dataset and by QA model.}
Tables~\ref{tab:e32_per_dataset} and~\ref{tab:e32_per_model} give the two cuts.
Table~\ref{tab:e14_per_dataset} holds the corresponding per-dataset values for the 14B editor.

\begin{table}[ht]
\centering
\caption{Failure detection per dataset with the Qwen3-32B editor on the full archives,
macro-averaged over the five QA models. Best per row in bold.}
\label{tab:e32_per_dataset}
\small
\begin{tabular}{@{}lcccc@{}}
\toprule
Dataset & NE & DSE & PE & NLL \\
\midrule
\multicolumn{5}{@{}l}{\emph{AUROC}}\\
SQuAD v1.1 & \textbf{0.743} & 0.681 & 0.669 & 0.667 \\
Quoref & \textbf{0.754} & 0.616 & 0.660 & 0.661 \\
NewsQA & 0.660 & 0.677 & 0.683 & \textbf{0.691} \\
Natural Questions & 0.701 & 0.667 & 0.727 & \textbf{0.730} \\
AdversarialQA & \textbf{0.765} & 0.668 & 0.630 & 0.620 \\
\midrule
\multicolumn{5}{@{}l}{\emph{AUPRC}}\\
SQuAD v1.1 & \textbf{0.470} & 0.420 & 0.325 & 0.325 \\
Quoref & \textbf{0.508} & 0.399 & 0.419 & 0.421 \\
NewsQA & 0.667 & 0.671 & 0.672 & \textbf{0.695} \\
Natural Questions & \textbf{0.512} & 0.470 & 0.482 & 0.493 \\
AdversarialQA & \textbf{0.682} & 0.601 & 0.530 & 0.524 \\
\bottomrule
\end{tabular}
\end{table}

\begin{table}[ht]
\centering
\caption{Failure detection per QA model with the Qwen3-32B editor on the full archives,
macro-averaged over the five datasets. Best per row in bold.}
\label{tab:e32_per_model}
\small
\begin{tabular}{@{}lcccc@{}}
\toprule
QA model & NE & DSE & PE & NLL \\
\midrule
\multicolumn{5}{@{}l}{\emph{AUROC}}\\
Gemma3-12B & \textbf{0.727} & 0.544 & 0.644 & 0.646 \\
OLMo2-13B-Instruct & \textbf{0.723} & 0.698 & 0.689 & 0.693 \\
Phi4-14B & \textbf{0.733} & 0.718 & 0.654 & 0.654 \\
Mistral-Nemo-12B & \textbf{0.716} & 0.682 & 0.710 & 0.705 \\
Falcon3-10B-Instruct & \textbf{0.723} & 0.666 & 0.673 & 0.670 \\
\midrule
\multicolumn{5}{@{}l}{\emph{AUPRC}}\\
Gemma3-12B & \textbf{0.488} & 0.317 & 0.372 & 0.382 \\
OLMo2-13B-Instruct & \textbf{0.592} & 0.560 & 0.517 & 0.535 \\
Phi4-14B & 0.578 & \textbf{0.579} & 0.469 & 0.465 \\
Mistral-Nemo-12B & \textbf{0.593} & 0.569 & 0.578 & 0.581 \\
Falcon3-10B-Instruct & \textbf{0.587} & 0.535 & 0.493 & 0.494 \\
\bottomrule
\end{tabular}
\end{table}

\paragraph{Pair by pair.}
Table~\ref{tab:e32_headtohead} compares NE against each other method one pair at a time.

\begin{table}[ht]
\centering
\caption{Per-pair comparison of NE against each other method with the Qwen3-32B editor on the
full archives, over the 25 dataset--model pairs. Counts are pairs won by NE.}
\label{tab:e32_headtohead}
\small
\begin{tabular}{@{}lcccc@{}}
\toprule
& \multicolumn{2}{c}{AUROC} & \multicolumn{2}{c}{AUPRC} \\
\cmidrule(lr){2-3}\cmidrule(lr){4-5}
NE versus & Pairs won & Mean $\Delta$ & Pairs won & Mean $\Delta$ \\
\midrule
DSE & 19/25 & $+0.063$ & 20/25 & $+0.056$ \\
PE & 16/25 & $+0.051$ & 21/25 & $+0.082$ \\
NLL & 15/25 & $+0.051$ & 18/25 & $+0.076$ \\
\bottomrule
\end{tabular}
\end{table}

\subsubsection{The Three Editors Side by Side}\label{app:editor_size_scaling}

Table~\ref{tab:e32_scaling} collects the overall values. The three scores read from the
original context do not consult the editor, but the instances they are scored on do: each is
computed on the population that run makes scorable, so NLL's $n$ moves from 24984 to 24993 to
25000 across the three and coverage rises for every method.

\begin{table}[ht]
\centering
\caption{Detection and coverage as the editor is scaled, on the full archives, macro-averaged
over the 25 dataset--model pairs. Values are each editor's own full archive, not the matched
population of Figure~\ref{fig:editor}.}
\label{tab:e32_scaling}
\small
\begin{tabular}{@{}lcccccc@{}}
\toprule
& \multicolumn{3}{c}{AUROC} & \multicolumn{3}{c}{Coverage} \\
\cmidrule(lr){2-4}\cmidrule(lr){5-7}
Method & 8B & 14B & 32B & 8B & 14B & 32B \\
\midrule
\textbf{NE} & 0.702 & 0.731 & 0.724 & 0.950 & 0.965 & 0.979 \\
PE & 0.676 & 0.673 & 0.674 & 0.950 & 0.965 & 0.979 \\
NLL & 0.674 & 0.674 & 0.674 & 0.999 & 1.000 & 1.000 \\
DSE & 0.652 & 0.658 & 0.662 & 0.877 & 0.929 & 0.955 \\
\midrule
\multicolumn{7}{@{}l}{\emph{No 32B run}}\\
FP & 0.605 & 0.638 & -- & 0.783 & 0.863 & -- \\
U-ME & 0.658 & -- & -- & 0.962 & -- & -- \\
\bottomrule
\end{tabular}
\end{table}

\subsection{Within-neighborhood and Between-neighborhood Variation}
\label{app:within_neighborhood}

\begin{figure}[ht]
\centering
\includegraphics[width=\linewidth]{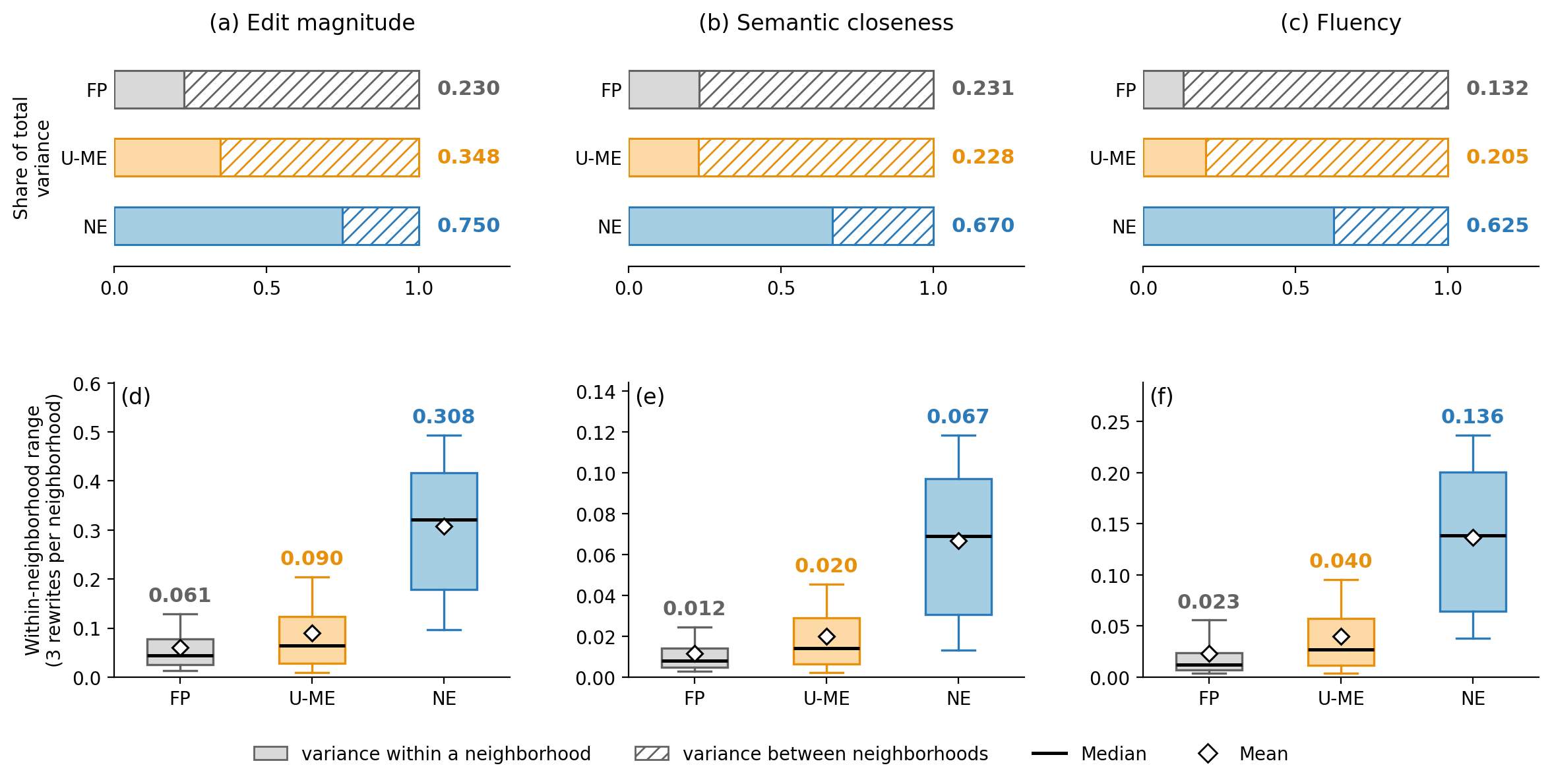}
\caption{Where rewrite variation lives, over the full archives and macro-averaged over the $25$ dataset--model pairs. (a)--(c) the share of each quality component's variance that falls within a neighborhood. (d)--(f) the within-neighborhood range of the same component at a fixed count of three rewrites ($m=3$ per neighborhood, averaged over $200$ random subsets, with boxes spanning $p_{25}$--$p_{75}$ and whiskers $p_{10}$--$p_{90}$ across neighborhoods. Each panel in (d)--(f) has its own scale.}
\label{fig:within_instance}
\end{figure}

The spread reported in Figure~\ref{fig:within_instance} is computed over all retained rewrites pooled together, so it confounds two sources of variation. The first is variation \emph{within} a neighborhood, among the rewrites of a single context. The second is variation \emph{between} neighborhoods: the level at which each context's rewrites sit, set by the passage as much as by the constructor. Only the first is available to an uncertainty estimate, since that estimate is computed from the rewrites of one context. We therefore decompose the pooled variance into a within- neighborhood and a between-neighborhood term.

\paragraph{Decomposition.} Fix a constructor, a dataset--model pair, and a quality-score component such as edit magnitude. Each instance has one neighborhood, holding the rewrites the constructor retained for that context. For the between-neighborhood term $\mathrm{Var}_{\text{between}}$, we average the component's score over each neighborhood's own rewrites, which reduces every neighborhood to a single scalar, and take the variance of those numbers across neighborhoods. For the within-neighborhood term $\mathrm{Var}_{\text{within}}$ we take the variance of the component' score inside each neighborhood separately, then average those variances over neighborhoods, so that every neighborhood counts once regardless of how many rewrites it holds. We report the share $\mathrm{Var}_{\text{within}} / (\mathrm{Var}_{\text{within}} + \mathrm{Var}_{\text{between}})$, computed per pair and averaged over the $25$ pairs.

\paragraph{Range at a fixed neighborhood size.} The share in Figure~\ref{fig:within_instance} panels (a)--(c) is a ratio of variances and is not affected by how many rewrites a neighborhood holds, but the range in panels (d)--(f) is, since the largest minus the smallest value grows with the number of values compared. Because the constructors retain different numbers of rewrites per neighborhood (Table~\ref{tab:diagnostics_details}), we hold that count fixed: from every neighborhood holding at least $m$ rewrites we
draw $200$ random subsets of size $m$, take the largest minus the smallest value in each, and average over the draws. Panels (d)--(f) report $m{=}3$. Here $m$ is a subsample size for this statistic alone.

\paragraph{Results.} Figure~\ref{fig:within_instance} reports both sources of variation for the three quality components over the full archives. The edit-magnitude shares are $0.75$ for NE, $0.35$ for U-ME and $0.23$ for FP, and the within-neighborhood ranges at $m{=}3$ are $0.308$, $0.09$ and $0.061$. Repeating the computation on the $K{=}3$ matched pool specifically, where all three constructors cover the identical $15730$ instances, gives shares of $0.77$, $0.39$ and $0.23$ and ranges of $0.32$, $0.10$ and $0.06$.

\subsection{Model Accuracies}\label{app:model_acc_results}

Table \ref{tab:qa_performance} indicates QA performance of the models evaluated in this work.
\begin{table}[ht]
\centering
\caption{QA performance of each model on each dataset, over the full evaluation population of
$1{,}000$ instances per dataset--model pair and $25{,}000$ in total. Accuracy is the fraction of
instances answered correctly under greedy decoding, using the stored correctness label that
defines the failure target throughout the paper. Token F1 is the token overlap between the
model's answer and the reference, which is sensitive to answer length in a way the correctness
label is not: Phi4-14B answers at $11.5$ words on average on AdversarialQA and $20.7$ on
NewsQA, against $2.3$--$4.7$ for Gemma3-12B, and its token F1 falls correspondingly further
below its accuracy. Row means are unweighted over the five datasets.}
\label{tab:qa_performance}
\small
\setlength{\tabcolsep}{2.5pt}
\begin{tabular}{lcccccc}

\toprule
Model & SQuAD v1.1 & Quoref & NewsQA & Natural Questions & AdversarialQA & Mean \\
\midrule
\multicolumn{6}{@{}l}{\emph{Accuracy}}\\
Gemma3-12B & $0.843$ & $0.781$ & $0.540$ & $0.755$ & $0.636$ & $0.711$ \\
OLMo2-13B-Instruct & $0.750$ & $0.698$ & $0.420$ & $0.692$ & $0.531$ & $0.618$ \\
Phi4-14B & $0.778$ & $0.764$ & $0.405$ & $0.689$ & $0.554$ & $0.638$ \\
Mistral-Nemo-12B & $0.729$ & $0.685$ & $0.415$ & $0.657$ & $0.528$ & $0.603$ \\
Falcon3-10B-Instruct & $0.759$ & $0.676$ & $0.467$ & $0.699$ & $0.520$ & $0.624$ \\
\cmidrule(l{2pt}){1-6}
Mean & $0.772$ & $0.721$ & $0.449$ & $0.698$ & $0.554$ & $0.639$ \\
\midrule
\multicolumn{6}{@{}l}{\emph{Token F1}}\\
Gemma3-12B & $0.797$ & $0.726$ & $0.596$ & $0.768$ & $0.603$ & $0.698$ \\
OLMo2-13B-Instruct & $0.737$ & $0.660$ & $0.483$ & $0.710$ & $0.516$ & $0.621$ \\
Phi4-14B & $0.661$ & $0.628$ & $0.363$ & $0.638$ & $0.436$ & $0.545$ \\
Mistral-Nemo-12B & $0.708$ & $0.631$ & $0.477$ & $0.680$ & $0.483$ & $0.596$ \\
Falcon3-10B-Instruct & $0.666$ & $0.540$ & $0.452$ & $0.671$ & $0.416$ & $0.549$ \\
\cmidrule(l{2pt}){1-6}
Mean & $0.714$ & $0.637$ & $0.474$ & $0.693$ & $0.491$ & $0.602$ \\
\bottomrule
\end{tabular}
\end{table}

\section{Neighborhood Construction}
\label{app:neighborhood_construction}

\subsection{Constructor Details}\label{app:constructor_details}

Algorithm ~\ref{alg:hard_inheritance_mapelites} and ~\ref{alg:soft_inheritance_mapelites} operate per QA example and share four operators.
\begin{itemize}[left=0pt]
    \item $\textsc{EmptyArchive}(\mathcal{D})$ initializes an empty archive $\mathcal{A}$ over the descriptor space $\mathcal{D}=\mathcal{T}\times\mathcal{S}$ (rewrite type $\times$ support explicitness).
    \item $\textsc{Edit}(\mathcal{E},P,\sigma)$ (Appendix~\ref{app:editor_edit_programs}) has the editor LLM $\mathcal{E}$ apply strategy $\sigma$ to a context $P$ and return an offspring rewrite $O$. For the seeds, $P=x$.
    \item $\textsc{Fitness}(O,x)$ (Appendix~\ref{app:quality_score}) scores $O$ against the original context $x$ as a normalized weighted mean of edit magnitude, fluency, and semantic closeness.
    \item $\textsc{Place}(\mathcal{A},O,\textsc{Fitness}(O,x))$ (Appendix~\ref{app:cell_assignment}) files $O$ in the cell given by its \emph{realized} edit type and its \emph{measured} support explicitness, replacing the current occupant only if $O$'s search objective value is higher.
\end{itemize}

The two algorithms differ in \emph{what they choose to edit} and in \emph{how strictly they validate the result}. Throughout, the budget $B$ is the \emph{per-example attempt budget}: the maximum number of editor calls (candidate rewrites attempted) for a single example, used identically by all three constructors.

\begin{table}[t]
\centering
\small
\setlength{\tabcolsep}{5pt}
\renewcommand{\arraystretch}{1.15}
\caption{The four neighborhood constructors. All four use the same editor, the same
answer-preservation gate and the same search objective, and they differ in whether
they search, in whether they verify the edit, and in whether the rewrite-type axis
carries a verified label or a requested one. Static bucket sampling verifies without
evolving and the untargeted search evolves without verifying, so between them they
bracket the two mechanisms NE combines. $\mathcal{D}$ is the twelve-cell descriptor
space spanned by edit type and answer support. A dash marks a property the
constructor has no machinery for.}
\label{tab:constructors}
\begin{tabular}{@{}l cccc@{}}
\toprule
& FP & SBS & U-ME & \textbf{NE}\\
\midrule
\multicolumn{5}{@{}l}{\textit{How candidates are generated}}\\
Archive over $\mathcal{D}$        & ---        & yes       & yes       & \textbf{yes}\\
Mutates an archived parent        & ---        & ---       & yes       & \textbf{yes}\\
Parent binds the offspring        & ---        & ---       & no        & \textbf{yes}\\
Targets empty cells               & ---        & no        & no        & \textbf{yes}\\
Edit types requested              & style only & all three & all three & \textbf{all three}\\
Prompt strategies                 & 1          & 1         & 2         & \textbf{10}\\
\addlinespace
\multicolumn{5}{@{}l}{\textit{What must hold before a rewrite is kept}}\\
Answer recoverable                & gate & gate & gate & \textbf{gate}\\
Requested edit realized           & gate & gate & ---  & \textbf{gate}\\
Distinct from those already kept  & gate & gate & ---  & \textbf{gate}\\
Differs from its parent           & ---  & ---  & ---  & \textbf{gate}\\
\addlinespace
\multicolumn{5}{@{}l}{\textit{What the archive records}}\\
Competes within a cell            & ---      & yes      & yes       & \textbf{yes}\\
Edit-type axis                    & verified & verified & requested & \textbf{verified}\\
Editor-call budget                & $\le 50$ & 25       & 25        & \textbf{25}\\
\bottomrule
\end{tabular}
\end{table}

\subsubsection{Neighborhood Elites (NE)}\label{app:algo_details_hard_map_elites}
Algorithm~\ref{alg:hard_inheritance_mapelites} adds three selection operators and a strict acceptance gate as follows:
\begin{itemize}[left=0pt]
    \item $\textsc{SelectTarget}(\mathcal{A})$ returns a cell to fill, preferring empty cells (to widen coverage) before weak occupied cells (to improve them).
    \item $\textsc{SelectParent}(\mathcal{A},t)$ returns the highest-scoring rewrite eligible for target type $t$, and $\textsc{SelectStrategy}(\Sigma,t)$ a strategy of that type.
    \item $\textsc{StrictValid}(O,x,\mathcal{Y})$ accepts $O$ only if it differs from $x$ and from its parent $P$, \emph{structurally realizes the requested edit type}, is not a duplicate of a rewrite already admitted to the archive, and keeps a complete answer group in the answer spec $\mathcal{Y}$ recoverable.
\end{itemize}
 
Intuitively, the search is goal-directed. It bootstraps by editing the original context $x$ under each applicable strategy, and from that point on, it repeats a \emph{target-mutate-validate} loop. It picks a descriptor cell to fill, mutates the best matching rewrite toward that cell under the hard $x\!\rightarrow\!P\!\rightarrow\!O$ contract (the original $x$ stays the validity reference), and admits the result only if it truly realizes the requested edit type. Because acceptance verifies the type, requested and realized types coincide, so every occupied cell holds a rewrite that genuinely \emph{is} what its coordinates indicate, thus the archive's rewrite-type axis is a faithful behavior label.

\begin{algorithm}[t]
\caption{NE (per example)}
\label{alg:hard_inheritance_mapelites}
\begin{algorithmic}[1]
\Require context $x$, question $q$, answer spec $\mathcal{Y}$, editor LLM $\mathcal{E}$, prompt strategy library $\Sigma$, descriptor space $\mathcal{D}=\mathcal{T}\times\mathcal{S}$, budget $B$
\State $\mathcal{A}\gets\textsc{EmptyArchive}(\mathcal{D})$
\For{$\sigma\in\Sigma$ applicable to $x$}\Comment{bootstrap}
  \State $O\gets\textsc{Edit}(\mathcal{E},\,x,\,\sigma)$
  \If{$\textsc{StrictValid}(O,\,x,\,\mathcal{Y})$}
    \State $\textsc{Place}\big(\mathcal{A},\,O,\,\textsc{Fitness}(O,x)\big)$
  \EndIf
\EndFor
\While{budget $B$ remains}
  \State $(t,s)\gets\textsc{SelectTarget}(\mathcal{A})$
  \State $P\gets\textsc{SelectParent}(\mathcal{A},\,t)$
  \State $\sigma\gets\textsc{SelectStrategy}(\Sigma,\,t)$
  \State $O\gets\textsc{Edit}(\mathcal{E},\,P,\,\sigma)$\Comment{hard inheritance: mutate $P$; $x$ stays the reference}
  \If{$\textsc{StrictValid}(O,\,x,\,\mathcal{Y})$}
    \State $\textsc{Place}\big(\mathcal{A},\,O,\,\textsc{Fitness}(O,x)\big)$
  \EndIf
\EndWhile
\State \Return $\textsc{Elites}(\mathcal{A})$
\end{algorithmic}
\end{algorithm}

\subsubsection{Untargeted Map Elites (U-ME)}\label{app:algo_details_soft_map_elites}
 
U-ME Algorithm~\ref{alg:soft_inheritance_mapelites} keeps the four shared operators but relaxes the hard variant in three ways as follows:

 \begin{itemize}[left=0pt]
     \item \textbf{No cell targeting.} Instead of the three selection operators ($\textsc{Select}{\ast}$), it seeds the archive from a fixed strategy set $\Sigma_0$ applied to $x$, and each round mutates the strategy $\sigma=(\tau,v)$ of each of the $k$ best rewrites $\textsc{TopElites} $$(\mathcal{A},k)$ --- a new edit type $\tau$ and a toggled prompt variant $v$. A candidate's cell is known only \emph{after} it is generated.     
     \item \textbf{Soft inheritance.} The parent rewrite is shown to the editor LLM only as \emph{guidance} in the edit prompt. The offspring is not required to differ from its parent.
     \item \textbf{A weaker gate.} It replaces $\textsc{StrictValid}$ with $\textsc{Valid}$, which still requires that $O\neq x$, and that a complete answer group in $\mathcal{Y}$ remain recoverable. However, it \emph{drops the edit-type realization check} and does not enforce edit type uniqueness. Consequently, $\textsc{Place}$ operator files each rewrite under the type it was \emph{requested}, whether or not the editor produced it---a cell labeled \textsc{distractor} may hold a paraphrase. Here the rewrite-type axis is a \emph{request} label, not the verified behavior label of the hard variant; only the support axis is measured from the output.
 \end{itemize}
 
\begin{algorithm}[t]
\caption{U-ME (per instance)}
\label{alg:soft_inheritance_mapelites}
\begin{algorithmic}[1]
\Require context $x$, question $q$, answer spec $\mathcal{Y}$, editor LLM $\mathcal{E}$, seed prompt strategies $\Sigma_0$, descriptor space $\mathcal{D}=\mathcal{T}\times\mathcal{S}$, attempt budget $B$
\State $\mathcal{A}\gets\textsc{EmptyArchive}(\mathcal{D})$
\For{$\sigma\in\Sigma_0$}\Comment{seed from $x$; no parent}
  \State $O\gets\textsc{Edit}(\mathcal{E},\,x,\,\sigma)$
  \If{$\textsc{Valid}(O,\,x,\,\mathcal{Y})$}
    \State $\textsc{Place}\big(\mathcal{A},\,O,\,\textsc{Fitness}(O,x)\big)$
  \EndIf
\EndFor
\While{budget $B$ remains}
  \For{$P\in\textsc{TopElites}(\mathcal{A},\,5)$}\Comment{global search objective value; no target cell}
    \State $\sigma\gets\textsc{Mutate}(P)$\Comment{new edit type $\neq P$'s; toggled prompt variant}
    \State $O\gets\textsc{Edit}(\mathcal{E},\,P,\,\sigma)$\Comment{soft inheritance: $P$ is only guidance}
    \If{$\textsc{Valid}(O,\,x,\,\mathcal{Y})$}
      \State $\textsc{Place}\big(\mathcal{A},O,\textsc{Fitness}(O,x)\big)$
    \EndIf
  \EndFor
\EndWhile
\State \Return $\textsc{Elites}(\mathcal{A})$
\end{algorithmic}
\end{algorithm}

\subsubsection{Free form Paraphrasing (FP) }\label{app:algo_details_llm_paraphrase}
The paraphrase constructor (Algorithm~\ref{alg:paraphrase}) reuses the same
per-example operators as the two MAP-Elites constructors but removes their evolutionary search structure entirely. It does not have archive, cell targeting, and it never selects or mutates a parent. A single fixed strategy $\sigma_{\mathrm{par}}$ (a \textsc{style\_rewrite} paraphrase prompt) is applied to the \emph{original} context $x$ within budget $B$. Each call is an independent draw, so the candidates are
i.i.d.\ paraphrases of $x$ rather than descendants of one another. This makes the constructor an \emph{unstructured, matched-size} control. It produces a neighborhood of the same construction cost and the same archive size as the hard archive it is paired with, but with no coverage objective and no behavioral-type structure.
 
Concretely, the three shared operators keep their exact meaning.
$\textsc{Edit}(\mathcal{E},x,\sigma_{\mathrm{par}})$ has the editor LLM $\mathcal{E}$ produce one paraphrase offspring $O$ of $x$. Because the parent is always $x$, there is no inheritance in this setting. $\textsc{StrictValid}(O,x,\mathcal{Y})$ is the \emph{same} strict gate used by the hard variant. $O$ must differ from $x$, structurally realize \textsc{style\_rewrite}, and contain a complete answer group in the answer set $\mathcal{Y}$ recoverable such that every admitted
paraphrase clears the identical answer-preservation condition. $\textsc{Fitness}(O,x)$ scores each valid paraphrase exactly as the other methods but here search objective value is only \emph{recorded}, never used to admit rewrites, or rank them as there are no descriptor cells to compete over. Admission is therefore first-come rather than competitive. Each valid, non-duplicate paraphrase is \emph{appended} to a flat list $\mathcal{P}$ (in contrast to $\textsc{Place}$, which would file it in a descriptor cell and keep only the local rewrite), and the loop stops once $\mathcal{P}$ reaches the matched size $K$ or the attempt budget is hit. A running attempt counter $a$ tracks how many paraphrases have been drawn. It caps the loop at $B$ attempts so that a hard example whose paraphrases repeatedly fail the $\textsc{StrictValid}$ gate or duplicate an existing member cannot spin indefinitely, and the neighborhood is returned with a reported shortfall $K-|\mathcal{P}|$ if the cap is reached first. The size $K$ is set to the final elite count of the paired hard- inheritance archive, so the two neighborhoods are the same size by construction and any downstream difference in the uncertainty signal is attributable to \emph{how} the neighborhood was built, not to how many members it has.

\begin{algorithm}[t]
\caption{FP (per example)}
\label{alg:paraphrase}
\begin{algorithmic}[1]
\Require context $x$, question $q$, answer set $\mathcal{Y}$, editor LLM $\mathcal{E}$, fixed paraphrase strategy $\sigma_{\mathrm{par}}$, matched size $K$, attempt budget $B$
\State $\mathcal{P}\gets\varnothing$;\quad $a\gets 0$
\While{$|\mathcal{P}|<K$ \textbf{and} $a<B$}
  \State $a\gets a+1$
  \State $O\gets\textsc{Edit}(\mathcal{E},\,x,\,\sigma_{\mathrm{par}})$\Comment{independent paraphrase of $x$; no parent, no mutation}
  \If{$\textsc{StrictValid}(O,\,x,\,\mathcal{Y})$ \textbf{and} $O\notin\mathcal{P}$}
    \State record $\textsc{Fitness}(O,x)$;\quad append $O$ to $\mathcal{P}$\Comment{fitness recorded, does not gate admission}
  \EndIf
\EndWhile
\State \Return $\mathcal{P}$
\end{algorithmic}
\end{algorithm}

\subsubsection{Static Bucket Sampling (SBS)}\label{app:algo_details_static_bucket_sampling}
Static bucket sampling (Algorithm~\ref{alg:static_bucket_sampling}) keeps the archive
and its gates but removes the search. It spends the same per-example budget $B$ as the
two MAP-Elites constructors, and every call is an independent proposal from the
original context $x$ under one fixed prompt strategy $\sigma_{\mathrm{sbs}}$, with the
requested rewrite type cycling through $\mathcal{T}$ in turn. The generation pattern is
therefore $x\rightarrow\tilde{x}_1$, $x\rightarrow\tilde{x}_2,\ \ldots$ with no parent
and no lineage, so a candidate's coordinates are known only \emph{after} it has been
produced. Placement is decided after generation rather than aimed at, which is the one
property that separates this constructor from NE.

Concretely, the shared operators keep their exact meaning.
$\textsc{Edit}(\mathcal{E},x,(\tau,\sigma_{\mathrm{sbs}}))$ has the editor LLM
$\mathcal{E}$ produce one rewrite $O$ of $x$ under requested type $\tau$. Because the
base is always $x$, there is no inheritance and the $O\neq P$ condition has nothing to
apply to. $\textsc{StrictValid}(O,x,\mathcal{Y})$ is the \emph{same} strict gate the
hard variant uses, read as it is for the paraphrase constructor: $O$ must differ from $x$,
must structurally realize the requested type $\tau$, must not duplicate a rewrite
already admitted to the archive, and must keep a complete answer group in the answer
spec $\mathcal{Y}$ recoverable. $\textsc{Place}$ then files the survivor in the cell
given by its realized type and its measured support explicitness, where it competes for
that cell on the search objective exactly as in the two MAP-Elites constructors.
Because the realization check is in force, requested and realized types coincide, so
the rewrite-type axis carries a verified behavior label here as it does for NE.

This constructor isolates the evolutionary component of the proposed constructor. It asks
whether the archive, the gates, and an explicit request for each edit type are by
themselves enough to recover the coverage that lineage and cell targeting produce.

\begin{algorithm}[t]
\caption{Static bucket sampling neighborhood (per example)}
\label{alg:static_bucket_sampling}
\begin{algorithmic}[1]
\Require context $x$, question $q$, answer spec $\mathcal{Y}$, editor LLM $\mathcal{E}$, fixed prompt strategy $\sigma_{\mathrm{sbs}}$, rewrite types $\mathcal{T}$, descriptor space $\mathcal{D}=\mathcal{T}\times\mathcal{S}$, attempt budget $B$
\State $\mathcal{A}\gets\textsc{EmptyArchive}(\mathcal{D})$;\quad $a\gets 0$
\While{$a<B$}
  \State $a\gets a+1$
  \State $\tau\gets\textsc{NextType}(\mathcal{T},\,a)$\Comment{cycle the three types in turn; no target cell}
  \State $O\gets\textsc{Edit}(\mathcal{E},\,x,\,(\tau,\sigma_{\mathrm{sbs}}))$\Comment{independent proposal from $x$; no parent}
  \If{$\textsc{StrictValid}(O,\,x,\,\mathcal{Y})$}
    \State $\textsc{Place}\big(\mathcal{A},\,O,\,\textsc{Fitness}(O,x)\big)$
  \EndIf
\EndWhile
\State \Return $\textsc{Elites}(\mathcal{A})$
\end{algorithmic}
\end{algorithm}
\subsection{Descriptor Space $\mathcal{D}$ and Cell Assignment}\label{app:descriptor_space}

First, we must distinguish between two objects that are easily confused.
The \emph{descriptor space} $\mathcal{D}$ is the fixed behavioral coordinate system, shared by every example and every constructor. The \emph{archive} $\mathcal{A}_x$ is the concrete, per-example container that stores at most one rewrite (an \emph{elite}) in each occupied coordinate. In short, $\mathcal{D}$ is the map and $\mathcal{A}_x$ is what a given instance places on it: each source QA instance $x$ receives its own archive.

We build an archive $\mathcal{A}_x$ per example per method. Each source QA instance receives its own MAP-Elites archive. The descriptor space $\mathcal{D}$ is the Cartesian product of a \textbf{rewrite-type} and a \textbf{support-explicitness} axes ($\mathcal{D}=\mathcal{T}\times\mathcal{S}$) where
\begin{itemize}[left=0pt]
    \item $\mathcal{T}=\{\texttt{distractor},\ \texttt{reorder},\ \texttt{style rewrite}\}$
    \item$\mathcal{S}=\{\texttt{direct support},\ \texttt{paraphrastic support},\\
\texttt{discourse mediated support},\ \texttt{multi sentence support}\},$
\end{itemize}
providing $3\times4=12$ possible cells in the archive. 

 
 


A candidate's cell in $\mathcal{D}$ is fixed by two coordinates obtained \emph{differently}:
\begin{itemize}[left=0pt]
  \item the $\mathcal{T}$ coordinate is the \emph{requested} edit operation. In the hard-inheritance constructor (and the paraphrase comparator) a candidate is admitted only after the realization gate (Appendix~\ref{app:realization}) confirms the edit is successfully carried out, so requested and realized types coincide and this coordinate is a \emph{verified} behavior label; the soft-inheritance constructor omits this gate and files by requested type alone (Table~\ref{tab:constructors}).
  \item the $\mathcal{S}$ coordinate is \emph{measured} from the produced context by the deterministic rule of Appendix~\ref{app:support_explicitness}, and may differ from any support level the editor is asked to target.
\end{itemize}

Only these two coordinates define a cell in $\mathcal{D}$: the prompt strategy (Appendix~\ref{app:editor_prompt_templates}) is \emph{not} a descriptor, so two candidates from different strategies compete in the same cell whenever their edit type and measured support bucket agree.

\subsubsection{Rewrite-type Axis $\mathcal{T}$}
\label{app:rewrite_type}
The rewrite-type axis records \emph{what kind of edit} produced the rewrite,
\[
\mathcal{T}=\{\texttt{distractor},\ \texttt{reorder},\ \texttt{style\_rewrite}\}.
\]
This axis is the \emph{requested} edit operation carried in the edit program
(\texttt{operation\_type}), not a type inferred from the output. In the
hard-inheritance and the paraphrase constructors, a candidate is admitted
only after the realization gate (Appendix~\ref{app:realization}) confirms the
requested edit is actually applied. Requested and realized types therefore
coincide for every stored rewrite, and this coordinate is a \emph{verified}
behavior label. The soft-inheritance constructor omits that gate and files by the
requested type alone, so there the axis is a request label rather than a verified one (Table~\ref{tab:constructors}).
 
\subsubsection{Support-Explicitness Axis $\mathcal{S}$} \label{app:support_explicitness}
The support-explicitness axis registers \emph{how directly} the edited context
states the evidence for the answer,
\[
\mathcal{S}=\{\texttt{direct},\ \texttt{paraphrastic},\ \texttt{discourse\_mediated},\ \texttt{multi\_sentence}\}.
\]
Unlike the rewrite type, this coordinate is not requested but \emph{measured} from
the produced context by a deterministic lexical/structural rule.
Intuitively: \texttt{direct} means the answer and its supporting cue sit together
in a single clause; \texttt{discourse\_mediated} means recovering the answer
requires following a coreference link; \texttt{multi\_sentence} means the support
is spread across two or more sentences; and \texttt{paraphrastic} is the residual
bucket for support that is present but meets none of the above.

Given the question, the edited context, and the gold answer, the cell is
assigned by a deterministic pipeline in four steps:

\begin{enumerate}[left=0pt]
  \item \textbf{Extract question anchors.} We tokenize the question 
  , lowercase it, and remove \emph{wh}-words and configured function words. We keep each remaining token $a$ and tag them with a rule-based tagger (suffix heuristic, not a trained model) into adjective, noun, number, proper noun, or verb. After dropping duplicates we obtain the anchor set $Anchor_Q$. For a span of text $T$, anchor coverage is
  \[
  R_{\mathrm{anchor}}(T)=\frac{|\{a\in Anchor_Q:\ a\text{ occurs in }T\}|}{|Anchor_Q|}
  \qquad(0\text{ when }Anchor_Q=\varnothing).
  \]
  Matching is exact at the token level (case insensitive; no stemming, lemmatization).
 
  \item \textbf{Locate the primary support sentence.} We split the context is split sentences by a \texttt{[.!?]+} regular expression. The rule fixes on \emph{primary support sentence} where the answer is grounded. The selection prefers in order: the sentence the edit flagged as its support location (if one is recorded), otherwise the first sentence containing the answer.
 
  \item \textbf{Select the support window.} This step decides how much context
  counts as the support. Among the windows that exist around the primary sentence
  $S_i$, we consider $\{S_i\}$, $\{S_{i-1},S_i\}$, $\{S_i,S_{i+1}\}$,
  $\{S_{i-1},S_i,S_{i+1}\}$. The $S_i$ is the default best and window replaces the best only on a \emph{strict} increase in $R_{anchor}$. The ties are resolved in favor of the earlier window in the list above. The selected window finally follows three cases, applied in order: 
  \begin{enumerate}[label=(\alph*),ref=(\alph*)]
    \item \textbf{Pronoun-driven expansion.} When the answer sentence leans on a
    pronoun (e.g., \texttt{it}, \texttt{this}, \texttt{these}) to point
    back to entity named earlier, it often does not itself carry the question's
    key terms. In that case, if a neighboring sentence supplies a term the answer
    sentence is missing, the window expands to include it, so the support is read
    across both sentences.
    \item \textbf{Keep the primary alone} when the support is already informative on its own: $R_{\mathrm{anchor}}\ge 0.5$ \emph{or} it contains at least two distinct
    anchors (this second condition can retain the primary even when its coverage
    is below $0.5$).
    \item \textbf{Otherwise expand} to the best window. This can reach a neighbor even with no pronoun present.
  \end{enumerate}
  The number of selected sentences is \texttt{support\_sentence\_count}: the
  extent of the chosen window and it is not a separately verified count of logically
  necessary evidence sentences.
 
  \item \textbf{Assign the bucket.} On the selected support, we compute three features. The \emph{support-sentence count} is the number of sentences
  selected above. The \emph{coreference proxy} is true when the primary sentence
  uses a configured coreference marker (Table~\ref{app:config}) yet does not itself state all of the question's anchors so the answer is reached through a referring expression rather than the question's own words (a multi-sentence support sets this proxy automatically, though rule~(a) below fires first). The \emph{same-clause test} is true when, after the sentence is split into clauses at the delimiters (Table~\ref{app:config}), some single clause contains both the answer and at least one question anchor. The bucket is then the first match of:
  \begin{enumerate}[label=(\alph*),ref=(\alph*)]
    \item \texttt{multi\_sentence} if \texttt{support\_sentence\_count}$\ge 2$;
    \item else \texttt{discourse\_mediated} if the coreference proxy is true;
    \item else \texttt{direct} if the same-clause test holds \emph{and} the
    selected sentence's coverage is $R_{\mathrm{anchor}}\ge 0.5$;
    \item else \texttt{paraphrastic} (residual; also the outcome for no anchors,
    no detected support, or an empty context).
  \end{enumerate}
Clauses for the same-clause test are cut at a fixed set of delimiters (Table~\ref{app:config}). 

Note that relative \emph{who}
  is a clause boundary and is \emph{not} in the coreference-marker set, so a
  support of the form ``\dots, who is represented by \dots'' is neither
  same-clause nor coreference-mediated and falls to \texttt{paraphrastic}.
\end{enumerate}
 
The bucket selected in the final step is the candidate's $\mathcal{S}$ coordinate;
together with the requested rewrite type (Appendix~\ref{app:rewrite_type}) it
fixes the archive cell (Appendix~\ref{app:cell_assignment}).

\subsubsection{Cell Assignment and Competition}
\label{app:cell_assignment}
A candidate is filed in the cell given by its requested rewrite type and its
measured support bucket. The editor is also parameterized by a \emph{prompt
strategy}---a finer instruction template within a rewrite type
(Appendix~\ref{app:editor_edit_programs})---but the strategy is \emph{not} a descriptor: two
candidates produced by different strategies land in the same cell whenever their
rewrite type and measured support bucket agree, and they then compete there. (A
per-candidate ``strategy realized'' flag is logged for analysis but plays no role
in placement.)
 
Each occupied cell holds exactly one elite. A valid candidate fills an empty
cell, or replaces the incumbent only if its scalar search objective value
(Appendix~\ref{app:quality_score}) is \emph{strictly} greater; ties keep the incumbent.
Candidates that fail validation or carry no search score cannot enter
$\mathcal{A}_x$.


\subsection{Editor and Edit Programs}\label{app:editor_edit_programs}

The editor in our study serves as a \emph{variation operator} that turns one context into a new one. Keeping this separate from the archive machinery is important. The editor provides linguistic variation, while the descriptor space and search (Appendices~\ref{app:descriptor_space} and  and~\ref{app:search_scheduler}) decide where a candidate lands and whether it is kept, resulting in an archive. The editor is a fixed local LLM, queried once per candidate. We provide its identity and generation settings (temperature, token limit, decoding) in the configuration table and we keep these parameters fixed across all datasets and QA models.

We describe the editor and its role in this study in three parts. The \emph{interface} describes the input query components that editor receives, the question, its gold answer, and the context to rewrite, and the outputs it generates. The \emph{edit program} drives the query with a requested rewrite type paired with a prompt strategy, an edit program the search evolves. The \emph{deterministic gates} (Appendix~\ref{app:deterministic_gates}) then screen every output, checking whether that answer is preserved and the requested edit is realized before it may enter the archive. All in all, the editor mutates a context under a typed instruction, and only rewrites that keep the answer and realize the edit survive. Below, we describe these steps in depth.
 
\paragraph{Editor interface.} We standardize each editor input and output query schema. Regardless of the edit program, we provide the same set of input fields and we instruct the editor to return the same set of output fields for each query. 
 
We provide three input components to the editor:
\begin{itemize}[left=0pt]
  \item the question $q$ and its gold answer $y$ such that editor receives what the rewrite must keep answerable;
  \item the \emph{base context} to rewrite. This is the parent $P$ when we mutate an
  existing rewrite, or the original context $x$ when we start fresh from a seed;
  \item the original context $x$ once more, hold aside as a fixed \emph{reference}
  for the later checks.
\end{itemize}

We split the base context from the reference to drive the inheritance mechanism act at the moment of generation, and it is precisely where the three constructors differ. Under the main \emph{hard-inheritance} constructor, the base is a selected parent rewrite $P$ and the reference is the original $x$: the editor mutates $P$ while $x$ stays pinned as the fixed validity reference, and the child must differ from both ($O\neq P$ and $O\neq x$). The \emph{soft-inheritance, U-ME} constructor still supplies a parent, but only as guidance. The child is not required to move away from it ($O\neq P$ is not enforced). The \emph{LLM-paraphrase} uses no parent at all: it always edits the original $x$, so its base and reference coincide, exactly as for a seed. The three are compared in full in Appendix~\ref{app:constructor_details} (Table~\ref{tab:constructors}).
 
The editor returns four fields, but only one is load-bearing:
\begin{itemize}[left=0pt]
  \item \texttt{EDITED\_CONTEXT} is the rewritten context. We use only this field downstream: we measure its descriptor (Appendix~\ref{app:descriptor_space}) and send it to the gates (Appendix~\ref{app:deterministic_gates}).
  \item \texttt{EXPECTED\_ANSWER} is the editor's own guess at the answer in its
  rewrite. We record this field and compare it to $y$, but it never decides admission.
  \item \texttt{STATUS} and \texttt{EDIT\_RATIONALE} is for the diagnostic purposes only.
\end{itemize}

Please note we employ only deterministic gates for validation decisions rather than an editor so the validations are editor-independent (Appendix~\ref{app:ans_preserving_val}). We provide the exact prompts we used in Appendix~\ref{app:editor_prompt_templates}.

\paragraph{Edit program.} We query the editor with an
\emph{edit program}, which is a requested rewrite type (\texttt{operation\_type}
$\in\mathcal{T}$) paired with a \emph{prompt strategy}. The two levels play different roles. The rewrite type names the \emph{behavior we want} such as a distractor, a reorder, or a style rewrite. Once realized, it becomes the candidate's $\mathcal{T}$ coordinate in descriptor space (Appendix~\ref{app:rewrite_type}). 

However, a bare rewrite type the lacks the refinement necessary to guide a general-purpose editor. An instruction as broad as ''rewrite this in a different style'' admits countless outputs and gives the search no way to aim for a \emph{particular} kind of rewrite. Therefore, we provide concrete guidance to fill in the missing specificity for realizing the rewrite type with prompt strategies. 

For a \texttt{style\_rewrite}, for instance, \texttt{support\_direct\_consolidation} prompt strategy asks the editor to pull the answer's support into a single clause, whereas \texttt{multi\_sentence\_factored} asks it to spread that support across several sentences. Both are style rewrites, yet they push the output toward \emph{different} \emph{support-explicitness cells}. An edit program is thus a (type, strategy) pair: (\texttt{style\_rewrite}, \texttt{multi\_sentence\_factored}) or (\texttt{distractor}, \texttt{distractor\_question\_overlap}). This structure indicates both the target behavior and the tactic used to realize it. We provide the full set of strategies and their rewrite types in Table~\ref{tab:rewrite_types_prompt_strategies_table}.

This two-level design is what lets an LLM act as a \emph{controllable} variation operator. The type fixes the axis of behavior; the strategy aims within it. The style strategies bias the output toward different support-explicitness cells: two candidates from different strategies that land in the cell compete there. The strategy is therefore a means, not a coordinate: it is how a candidate is \emph{steered}, not how it is \emph{filed}. Many strategies map to one rewrite type, and two candidates from different strategies that land in the same cell simply compete there (Appendix~\ref{app:cell_assignment}).

\begin{table}[t]
\caption{The ten prompt strategies grouped by rewrite type. Candidates from different strategies can occupy the same cell in descriptor space $\mathcal{D}$, where they compete on quality. We detail the prompt templates in Appendix~\ref{app:editor_prompt_templates}.} 
\vspace{5pt}
\centering
\small
\begin{tabular}{@{}l l@{}}
\toprule
\textbf{Rewrite type} & \textbf{Prompt strategies} \\
\midrule
\multirow{2}{*}{\texttt{distractor}}
  & \texttt{distractor\_support\_proximal} \\
  & \texttt{distractor\_question\_overlap} \\
\midrule
\multirow{3}{*}{\texttt{reorder}}
  & \texttt{reorder\_adjacent\_swap} \\
  & \texttt{reorder\_support\_back} \\
  & \texttt{reorder\_global} \\
\midrule
\multirow{5}{*}{\texttt{style\_rewrite}}
  & \texttt{support\_direct\_consolidation} \\
  & \texttt{lexical\_paraphrase\_biased} \\
  & \texttt{paraphrase\_biased} \\
  & \texttt{controlled\_coreference\_rewrite} \\
  & \texttt{multi\_sentence\_factored} \\
\bottomrule
\end{tabular}
\label{tab:rewrite_types_prompt_strategies_table}
\end{table}

\subsection{Deterministic Gates}\label{app:deterministic_gates}
 
We admit a candidate rewrite into the archive only after it passes deterministic checks that consult neither the QA model $f_\theta$ nor the editor. The rewrite must be non-empty, must differ from the original  $x$ (and, for offspring, from the parent $P$), and must not duplicate a rewrite already in the archive. Two substantive gates then decide admissibility: \emph{answer preservation} and \emph{edit-type realization}. Together they form the $\textsc{StrictValid}$ operator used by the NE and the FP  (Appendix~\ref{app:constructor_details}). The U-ME uses the lighter $\textsc{Valid}$ operator, which applies answer preservation but omits the realization guard.
 
\subsubsection{Answer-Preserving Validation}\label{app:ans_preserving_val}
We require the gold answer to remain \emph{recoverable} from the edited context. We test this with a deterministic lexical check against an \emph{answer specification} (\texttt{AnswerSpec}).
\begin{itemize}[left=0pt]
  \item An \texttt{AnswerSpec} may list several alternative answer groups (for
  example, acceptable variants of the gold answer). The edited context passes if
  every required value of \emph{at least one} group is lexically recoverable in it.
  \item Passing this gate certifies only that the answer remains recoverable, not that the edit preserves every source fact. The editor may drop or alter unrelated details and the candidate still passes. We therefore call the edits \emph{answer-preserving}, not meaning- or information-preserving.
  \item We log the editor's \texttt{EXPECTED\_ANSWER}
  (Appendix~\ref{app:editor_edit_programs}) and compare it to the gold answer, but a mismatch does not block admission; only the \texttt{AnswerSpec} test gates.
\end{itemize}
 
\subsubsection{Edit-Type Realization Guard} \label{app:realization}
We file a candidate under the edit type it \emph{requested}
(Appendix~\ref{app:rewrite_type}), so we verify that the editor actually performed that type; otherwise the axis would record a request rather than a behavior. We reject a candidate that fails as \texttt{EDIT\_TYPE\_NOT\_REALIZED}. Table~\ref{tab:realization} gives the per-type conditions; two requirements apply to all three types:

\begin{itemize}[left=0pt]
  \item \textbf{Reference.} We compare the child against the realization reference (the original $x$ for seeds, the parent $P$ for offspring) and require every accepted child to \emph{also} differ from the original $x$.
  \item \textbf{Normalization.} Every comparison runs on a \emph{canonical form} of the text, one standardized rendering that all cosmetically-different variants map so that formatting alone is never mistaken for an edit. For the sentence-level checks, we split the context into sentences at \texttt{.!?} (discarding those end marks) and standardize each sentence's Unicode form (NFKC) and spacing. Two sentences that differ only in punctuation or spacing then considered as identical. For the \texttt{style\_rewrite} check, we reduce each context to its ordered list of words (case-folded \texttt{\textbackslash w+} tokens), punctuation dropped and case ignored, so \texttt{The} and \texttt{the} are the same word. Then we compare those lists we obtain. Only a genuine change in wording or in sentence order therefore registers as an edit.
\end{itemize}

\begin{table}[t]
\centering
\small
\caption{The three edit types: the prompt intention (what the editor is asked to do) and the deterministic condition verified between the child and its realization reference (the parent $P$ for offspring, the original context $x$ for seeds).}
\label{tab:realization}
\begin{tabular}{@{}p{2.4cm}p{4.5cm}p{5.3cm}@{}}
\toprule
\textbf{Rewrite type} & \textbf{Prompt intention} & \textbf{Deterministic realization condition} \\
\midrule
\texttt{distractor} &
Add plausible but answer-irrelevant material while preserving the evidence. &
The child differs from the reference, adds at least one sentence, and retains every
reference sentence as an exact \emph{ordered subsequence}. \\
\addlinespace
\texttt{reorder} &
Change sentence order without adding, deleting, or paraphrasing sentences. &
The reference has at least two sentences; the child differs from it and has the
\emph{same normalized sentence multiset} in a different order. \\
\addlinespace
\texttt{style\_rewrite} &
Rewrite substantive wording or syntax. &
The child's token sequence changes, and the child is \emph{not} merely a sentence
permutation and \emph{not} merely a sentence insertion. \\
\bottomrule
\end{tabular}
\end{table}

Because admission verifies the type, requested and realized types coincide for every rewrite stored under the hard-inheritance constructor and the FP (which realizes \texttt{style\_rewrite}), which makes the $\mathcal{T}$ axis a \emph{verified} behavior label. The U-ME omits this guard and files by the requested type alone (Table~\ref{tab:constructors}). The guard certifies edit \emph{structure}, not semantics. Therefore, it does not establish distractor irrelevance or paraphrase faithfulness.

 
\subsection{Search Objective}\label{app:quality_score}
The search objective value is the scalar fitness MAP-Elites uses to select a cell's best rewrite. Because MAP-Elites keeps only one rewrite per occupied cell, it needs a scalar to decide, among several valid rewrites that fall in the same descriptor cell of $\mathcal{D}$, which one to keep (Appendix~\ref{app:cell_assignment}). The value of the search objective is that scalar ranks only candidates that have already cleared the gates (Appendix~\ref{app:deterministic_gates}). It never decides admissibility, and it plays no part in the uncertainty assessment. Because validity is already guaranteed by the deterministic gate, the value is free to express a preference among the valid candidates competing for a cell. In our study, we favor rewrites that vary the surface form while remaining fluent and close to the original meaning.

We keep this heuristic intentionally free of any language-model judge, for two reasons. First, computational cost: MAP-Elites scores every candidate it generates, so a judge would add a model inference call to each of the many search iterations per example, limiting the feasibility at the scale at which we build archives. Second, determinism and reproducibility: a deterministic score makes each archive exactly reproducible, whereas a judge would inject stochasticity, version drift, and prompt sensitivity into the experimental flow, for no stronger validity guarantee than the deterministic gate already provides. Our choice is also standard QD practice: classic MAP-Elites uses a cheap, deterministic scalar fitness, so we follow the mainstream QD design rather than deviating from it.

We combine three component scores as a normalized weighted mean,
\[
Fitness(c) \;=\; \frac{w_{\mathrm{edit}}\,q_{\mathrm{edit}}(c) + w_{\mathrm{flu}}\,q_{\mathrm{flu}}(c) + w_{\mathrm{sem}}\,q_{\mathrm{sem}}(c)}{w_{\mathrm{edit}} + w_{\mathrm{flu}} + w_{\mathrm{sem}}},
\]
with fixed weights given in the configuration table and held constant across all
datasets and QA models.
 
\begin{itemize}[left=0pt]
  \item \textbf{Edit-Magnitude Score $q_{\mathrm{edit}}$} measures how much the candidate's text changed relative to the original context $x$. We compare two unit sequences, the tokens (or sentences) $a$ of original $x$ and $b$ of the candidate $c$, with \texttt{difflib.SequenceMatcher} \citep{ratcliff1988gestalt}, which identifies their matching and differing regions (a diff), and summarize the difference by the change ratio
  \[
  R(a,b)=\frac{\sum_{\text{non-matching blocks}}\max(\#\text{deleted},\,\#\text{inserted})}{\max(|a|,|b|,1)},
  \]
  the size of the non-matching regions as a fraction of the longer sequence. $q_{\mathrm{edit}}$ is the mean of this ratio over tokens and over sentences, plus an answer-span change-concentration term when the answer's span can be located.\footnote{The fraction of all changed tokens that fall within the character span of the answer's first occurrence in $x$. It is included when that span exists and near zero whenever the answer text is preserved.} It is a surface measure, not a parse-based notion of syntactic difference.
  \item \textbf{Fluency $q_{\mathrm{flu}}$} is a deterministic proxy: the mean of a length-balance term, a flag for the absence of runs of three or more repeated punctuation marks, and a printability check. It is not a learned or human-validated fluency model.
  \item \textbf{Semantic closeness $q_{\mathrm{sem}}$} is the BERTScore F1 \citep{zhang2020bertscore} between the original context $x$ and the child, computed after validation, using a fixed encoder recorded in the configuration table.
\end{itemize}
 
$q_{\mathrm{edit}}$ and $q_{\mathrm{sem}}$ are computed on the same pair $(x\to c)$ and pull in opposite directions. $q_{\mathrm{edit}}$ rewards moving away from the original surface form, and $q_{\mathrm{sem}}$ rewards staying close to its meaning. Therefore, their balance expresses exactly the neighbor we aim to form: surface-different but meaning preserving. Admissibility thus remains fully deterministic and model-free (Appendix~\ref{app:deterministic_gates}); a neural model enters only through the non-gating closeness term $q_{\mathrm{sem}}$.

\subsection{Search and Scheduler}\label{app:search_scheduler}
 
The search is the per-example procedure that spends a fixed evaluation budget $B$ to populate and then improve the archive $A_x$. Each \emph{evaluation} is one editor call: we form a single edit program, generate one context, and pass it through the gates, descriptor assignment, and competition described above. The budget counts attempts rather than successes. Rejected, discarded, and accepted candidates each consume one unit so a harder instance yields a sparser archive within the same budget rather than more calls. NE, U-ME and SBS each spend $25$ evaluations per example whereas FP consumes up to $50$ attempts. All four reuse the same editor (Appendix~\ref{app:editor_edit_programs}) and the same gates (Appendix~\ref{app:deterministic_gates}). They differ only in how the next edit program is chosen and how an accepted context is stored. Table~\ref{tab:constructors} compares the four construction methods.
 
\subsubsection{One Evaluation}\label{app:search_step}
Given an edit program, every evaluation proceeds the same way. The editor produces one context (Appendix~\ref{app:editor_edit_programs}); the deterministic gates accept or reject it (Appendix~\ref{app:deterministic_gates}); and an accepted context is stored. The three archive-based constructors store it by \emph{placement}: the context is assigned to its cell from its measured descriptor (Appendix~\ref{app:descriptor_space}) and competes for that cell by search objective score (Appendix~\ref{app:quality_score}), filling the cell if empty or replacing the occupant only if its score is strictly higher. FP has no grid and instead \emph{appends} the accepted context to a flat list. Insertion is sequential in every case. The archive (or list) is updated after each evaluation.
 
\subsubsection{Targeted Staged Scheduler (NE)}\label{app:search_staged}
NE spends its budget in three stages, \emph{bootstrap}, \emph{coverage}, then \emph{fitness}. Each stage has a distinct objective. Bootstrap and coverage together may use up to $B-r$ evaluations, where $B$ is the per-example budget and $r$ is a reserve budget held for the fitness-improvement stage (here $B=25$ and $r=5$, so at most $20$ evaluations precede the reserve). The reserve is protected capacity, not a fixed count. If coverage stops early the fitness stage inherits the unused evaluations, and if no action is possible even the reserve may go unspent.
 
\paragraph{Bootstrap.} We seed the archive with one attempt per strategy. For each (rewrite type, strategy) pair, taken in a fixed order, we generate a single candidate directly from the original context $x$ with no parent. This populates initial cells cheaply and gives the later stages candidate rewrites to mutate. With the ten strategies of Appendix~\ref{app:editor_edit_programs}, bootstrap issues up to ten attempts.
 
\paragraph{Coverage.} The coverage stage spends the pre-reserve budget filling empty
cells. Each iteration chooses a target cell, a parent, and a strategy:
\begin{itemize}[left=0pt]
  \item \textbf{Target.} We consider only empty cells still worth trying, which we call \emph{actionable}. A cell qualifies if it has been attempted fewer than twice, at least one strategy can target it, and the archive holds at least one rewrite eligible as its parent. Among the actionable cells we pick the one attempted the fewest times so far, breaking ties by a seeded random draw.
  \item \textbf{Parent.} We rank rewrites by decreasing search objective score and take the first eligible one. For \texttt{distractor} and \texttt{reorder} the parent must already occupy the target support cell, since these edits inherit their parent's support coordinate (Appendix~\ref{app:support_explicitness}). For \texttt{style\_rewrite} any elite may serve and the strategy itself aims at the cell.
  \item \textbf{Strategy.} Among strategies compatible with the target cell, we take the globally least-attempted, skipping any that has no eligible parent.
\end{itemize}
Coverage ends when no empty cell is actionable, when a patience of five consecutive attempts fails to occupy a new cell, or when the budget reaches the fitness-reserve boundary.

\paragraph{Fitness improvement.} The fitness-improvement stage spends the reserve improving occupied cells, weakest first. We order the current archived rewrites by increasing search objective score and cycle through them. For each, we propose a new candidate for that rewrite's cell using the least-attempted compatible strategy, and it replaces the occupant only if its search objective value is strictly greater (Appendix~\ref{app:quality_score}). Because a candidate is placed by its own measured descriptor, an improvement attempt may land in a different cell than intended. Targeting is therefore best-effort, and final placement always follows
the realized descriptor. The stage ends when the budget is exhausted or no cell admits a further attempt.
 
\paragraph{Illustration.} The split among the three stages is emergent, not configured. On one SQuAD instance the $25$ evaluations divided into $10$ bootstrap, $4$ coverage,
and $11$ fitness improvement, filling $6$ of the $12$ cells: all ten strategies were eligible at bootstrap, coverage halted after four attempts once no empty cell remained actionable, and the remaining eleven evaluations fell to the fitness stage.

\subsubsection{Generational Search (U-ME)}\label{app:search_soft}
U-ME spends the same $25$-evaluation budget as an untargeted, generational loop. It never aims at a particular cell, and a candidate's cell is known only after it is generated and measured. It begins by generating five seeds directly from $x$ with no parent, cycling through the three rewrite types (distractor, reorder, style rewrite, then distractor and reorder again) under the fixed prompt variant
\texttt{support\_rewrite\_aggressive}. It then repeats a generational step until the budget is spent. As in NE, we rank the current rewrites by decreasing search objective score, take up to the top five, and generate exactly one offspring from each. Each offspring's edit program is a deterministic mutation of its parent's. The requested rewrite type is changed to a different type, and the prompt variant is toggled between the constructors's two variants (\texttt{support\_rewrite\_aggressive} and \texttt{local\_conservative}). In this case, the mutation is a function of the attempt index rather than a random draw. Offspring are validated, placed, and compete exactly as in Appendix~\ref{app:search_step}, except that this constructor applies the lighter validity gate: it requires answer preservation without the realization guard (Appendix~\ref{app:deterministic_gates}) and files each rewrite by its \emph{requested} type. The loop stops when the budget is exhausted or when no elite is available to serve as a parent.
 
\subsubsection{Append Loop (FP)} \label{app:search_paraphrase}
FP has no archive, no parent and no cell targeting. It draws independent paraphrases of the original context $x$ and appends the valid ones until it matches the size of the main method's archive. For each test instance, we set the target count $K$ to the number of accepted elites in that example's NE archive, so the constructor is compared at the same neighborhood size. We then attempt up to $50$ proposals, each a single fixed edit program, a \texttt{style\_rewrite} under the \texttt{llm\_paraphrase} prompt generated from $x$ with no parent. Each proposal passes through the same gates as the paraphrase-realizing main-method candidates (Appendix~\ref{app:deterministic_gates}) and is additionally rejected if it duplicates an
already-accepted paraphrase (exact match after Unicode NFKC and whitespace normalization). Accepted paraphrases are appended in acceptance order; unlike the archive-based constructors there is no cell, no competition and no replacement. The search objective is recorded but never used to admit or rank (Appendix~\ref{app:quality_score}). The loop stops at $K$ accepted paraphrases or $50$ attempts, whichever comes first. We allow more attempts here than the $25$ of the MAP-Elites constructors because FP reaches $K$ less reliably; the larger attempt budget strengthens the FP, making the matched-$K$ comparison conservative.


\section{Prediction and Uncertainty Scoring}\label{app:prediction_scoring}

We turn each archive into one uncertainty score by asking the frozen QA model to answer every rewrite in the neighborhood and measuring how much its answers disagree. The steps below fix the parts of that computation that the main text states only in outline: the query, the meaning-based grouping, and the properties of the resulting scores.

We query $f_\theta$ once on each archived rewrite and never on the original context $x$. For an instance $(q,x,y)$ we hold the question $q$ fixed, present each rewrite kept in $\mathcal{A}_x$ to $f_\theta$ under a single fixed prompt. The prompt provides the same instruction for every
dataset and QA model, and the question before the context and asking for the answer alone (shown below). We keep the first cleaned line of the answer, with any \texttt{answer:} preamble removed, as the prediction: one per rewrite. Decoding is greedy (we set the temperature $0$, at most $64$ new tokens), so the predictions are deterministic given the archive. We exclude $x$ itself, because whether $f_\theta$ answers $x$ correctly is the failure we later try to detect, and it must not enter the signal used to detect it. How many of these predictions, $K$, enter each score is fixed by the evaluation protocol (Appendix~\ref{app:evaluation_protocol}).

\paragraph{QA-model prompt.}
Every query uses the same system and user messages:
\begin{promptbox}[title=System prompt (QA model)]
\begin{Verbatim}[breaklines=true, breakanywhere=true, breaksymbolleft={}, fontsize=\small]
You are a question answering system with highest accuracy.
\end{Verbatim}
\end{promptbox}
\begin{promptbox}[title=User template (QA model)]
\begin{Verbatim}[breaklines=true, breakanywhere=true, breaksymbolleft={}, fontsize=\small]
Answer the question using only the provided context.
 
Question:
{question}
 
Context:
{context}
 
Return only the answer, make sure that answer is correct and do not generate a sentence explaining or describing the answer.
\end{Verbatim}
\end{promptbox}
We deliver these through each model's chat template (with any thinking mode disabled) or, for a model that exposes no chat template, as a single plain-text prompt with \texttt{System:}, \texttt{User:}, and \texttt{Assistant:} labels; the wording is identical across datasets and QA models, and only the chat-template control tokens differ.

We group the predictions by meaning in two stages, where the second stage only merges the classes the first stage forms and never splits them. We first collapse identical predictions into exact-answer classes under a fixed normalization, then merge classes that carry the same meaning with a frozen NLI model (\texttt{DeBERTa-v3-base-mnli-fever-anli}), following the semantic entropy construction of \citet{kuhn2023semantic}. For two exact classes, we render one representative of each as the question-conditioned statement ''\texttt{The answer to "$q$" is $a$.}'' and classify the pair in both directions. The two classes are equivalent when the model's top label is \emph{entailment} each way. In addition, the semantic classes are the connected components of this relation, so a chain of equivalent pairs joins a single cluster. For example, if a neighborhood yields the predictions \emph{Uruguay}, \emph{uruguay.}, \emph{the Uruguay national team}, and \emph{Argentina}, normalization already places \emph{Uruguay} and \emph{uruguay.} in one exact class; the model then judges \emph{the Uruguay national team} equivalent and merges it in, while \emph{Argentina} stays apart, so the four raw strings collapse to two answer meanings. The model sees only the question and the predictions; it never sees the gold answer $y$ or the failure label, and no context is included. The normalization behind the exact classes strips any \texttt{answer:}-style prefix and keeps the first line, applies Unicode \textsc{nfkc} and maps common Unicode punctuation to ASCII, lower-cases, deletes punctuation, drops the articles \emph{a}, \emph{an}, and \emph{the}, collapses whitespace, and rewrites a pure number-word answer (for example \emph{twenty-one}) as digits; it applies no stemming and no accent folding.

\section{Evaluation Protocol} \label{app:evaluation_protocol}

\paragraph{Correctness and failure labels.}
We score each constructor by how well its neighborhood-derived uncertainty separates
correct from incorrect original-context predictions. The prediction we flag is
$f_\theta$'s answer on the original context $x$, and we judge its correctness with a
permissive match against the instance's answer specification: the answer counts as correct
if it matches any accepted gold group---alternative gold answers are singleton groups, so
matching any one suffices---through a cascade of increasingly permissive checks: an exact
match; a normalized match (the normalization of Appendix~\ref{app:prediction_scoring},
extended with month-name date canonicalization); a conservative containment test, in which
a short gold string that appears verbatim inside the normalized prediction counts as a
match; and, last, a bidirectional-entailment test using the same NLI model as the semantic
clustering (Appendix~\ref{app:prediction_scoring}). The failure label is the negation of
this correctness, and the cascade is permissive in the direction that matters: a paraphrase
of the gold answer is scored correct rather than counted as a failure, so surface variation
does not inflate the failure set.

\paragraph{Failure detection: AUROC and AUPRC.}
We compute AUROC and AUPRC within each of the 25 dataset--model pairs and report the
macro-average over pairs, so every pair carries equal weight whatever its size. AUPRC is
reported as lift over the pair's base rate, because base rates move with the pool and range
from 0.315 to 0.361 across methods. A difference between two methods is computed on the
instances both of them score, and its 95\% interval comes from 2000 bootstrap replicates
that resample instances within each pair and recompute the macro-average. This bootstrap is
distinct from the sub-selection repeats of the matched design
(Appendix~\ref{app:matched_eval}), which also number 2000 but resample which $K$ rewrites
enter a neighborhood rather than which instances enter a pair.

\paragraph{Scoring pool.}
We score every method on the instances it generates. Each dataset--model pair contributes
1000 instances, and a method scores an instance when it has at least two predictions for it.
Two is the smallest count at which semantic entropy is defined, since the measure is
identically zero on a single prediction. The restriction is imposed by the measure rather
than chosen: the three scores read from the original context are defined on a single
prediction and none of them improves under it, so the restriction cannot be what produces
the comparison we report. Unless a table states otherwise, no subselection is applied and
every rewrite a constructor retained for an instance enters that instance's score.

\paragraph{Reported columns.}
The result tables throughout the appendix share a common set of columns, defined once here.
The number of instances scored differs by method, so we report it as $n$, alongside
\emph{coverage}, the share of the 1000 instances per pair that a method scores at all. Base
rates therefore differ by method as well, and AUPRC is reported beside the base rate it sits
above, because AUPRC is not comparable across populations of different prevalence;
\emph{lift} is AUPRC minus that base rate. Where a table reports \textsc{sd}, \textsc{min}
and \textsc{max}, these summarise the 25 per-pair AUROC values and describe how evenly a
method performs across pairs, not how precisely its mean is estimated.
\section{Configuration}\label{app:configuration}
We fix a single configuration for all experiments and report it here for reproducibility. Every value is held constant across datasets and QA models. Table~\ref{app:config} lists the settings, grouped by component (editor, support-explicitness rule, fitness (search objective) score, gates, and search); the mechanisms that use them are described in the referenced appendices.

\begin{table}[!ht]
\caption{Configuration used throughout, held fixed across all datasets and QA models.\\}
\centering
\small
\setlength{\tabcolsep}{6pt}
\renewcommand{\arraystretch}{1.2}
\begin{tabularx}{\linewidth}{@{}l >{\raggedright\arraybackslash}X@{}}
\toprule
\textbf{Setting} & \textbf{Value} \\
\midrule
Random seed & $7$ \\
\addlinespace
\multicolumn{2}{@{}l}{\textit{Editor} (Appendix~\ref{app:editor_edit_programs})}\\
Model & Qwen3-8B (local MLX, 8-bit) \\
Decoding parameters & temperature $0.7$; sampling on; max new tokens $1200$ \\ \hline
\addlinespace
\multicolumn{2}{@{}l}{\textit{Support-explicitness rule} (Appendix~\ref{app:support_explicitness})}\\
Sentence-split / token patterns & \texttt{[.!?]+} \ /\ \texttt{[A-Za-z0-9]+('[A-Za-z0-9]+)?} \\
Anchor overlap threshold; multi-sentence min & $0.5$; \ $2$ \\
POS kept by anchor tagger & adjective, noun, number, proper noun, verb \\
Coreference markers & \{he, she, they, them, it, its, his, her, their, this, that, these, those, former, latter, such\} \\
Clause-split delimiters & comma, colon, semicolon, parenthesis; \emph{and, but, because, while, although, though}; \emph{which, who, that, where, when} \\
\addlinespace
\multicolumn{2}{@{}l}{\textit{Search objective score} (Appendix~\ref{app:quality_score})}\\
Weights $(w_{\mathrm{edit}},w_{\mathrm{flu}},w_{\mathrm{sem}})$ & $(0.3,\,0.2,\,0.1)$; normalizer $\sum w=0.6$ \\
Semantic-closeness encoder & BERTScore F1, RoBERTa-large (English default), no baseline rescaling \\ \hline
\addlinespace
\multicolumn{2}{@{}l}{\textit{Gates} (Appendix~\ref{app:deterministic_gates})}\\
Realization guard & FP: on;\ \ SBS: on;\ \ U-ME: off;\ \ NE: on \\ \hline
\addlinespace
\multicolumn{2}{@{}l}{\textit{Search} (Appendix~\ref{app:search_scheduler})}\\
Budget & FP: up to $50$ attempts;\ \ SBS, U-ME, NE: $25$ evaluations each \\
Hard staged scheduler & \texttt{bootstrap\_attempts\_per\_strategy}$=1$; \texttt{coverage\_patience}$=5$; \texttt{target\_attempt\_limit}$=2$; \texttt{quality\_evaluation\_reserve}$=5$; \texttt{target\_selection}$=$\texttt{least\_attempted\_seeded} \\
Soft-inheritance & \texttt{init\_seed\_candidates}$=5$; \texttt{parent\_selection}$=$\texttt{best}; \texttt{offspring\_per\_iteration}$=5$; prompt variants \{\texttt{support\_rewrite\_aggressive}, \texttt{local\_conservative}\} \\
LLM paraphrase & $K=$ matched hard-archive elite count; \texttt{max\_attempts}$=50$ \\ \hline
Static bucket sampling & requested type cycles $\mathcal{T}$ in turn under one fixed prompt strategy \texttt{local\_conservative}\; no parent, no cell targeting \\
\addlinespace
\multicolumn{2}{@{}l}{\textit{Prediction and uncertainty scoring} (Appendix~\ref{app:prediction_scoring})}\\
QA decoding & greedy: temperature $0$; sampling off; max new tokens $64$ \\
QA prompt & \texttt{grounded\_qa} (question before context; answer only; fixed across datasets and QA models) \\
Semantic-clustering model & \texttt{DeBERTa-v3-base-mnli-fever-anli} (HF NLI backend) \\
Equivalence rule & bidirectional entailment, argmax label both directions; max pairwise comparisons $128$ \\ \hline
\addlinespace
\multicolumn{2}{@{}l}{\textit{Evaluation} (Appendix~\ref{app:evaluation_protocol})}\\
Neighborhood size $K$ & $\{3,4,5,6,7\}$; separate matched pool per $K$ \\
Sub-selection repeats & $R=2000$; report mean and empirical $[2.5,97.5]$ percentile interval \\
Per-draw seed & first $64$ bits of SHA-256$($seed, dataset, QA model, constructor, instance, $K$, repeat$)$ \\
\bottomrule
\end{tabularx}
\label{app:config}
\end{table}

\section{Editor Prompt Templates}\label{app:editor_prompt_templates}

The two MAP-Elites constructors uses one fixed prompt template on each query with two parts. The \emph{system prompt} states the editor's role and the rules it must follow and is identical on every query. The \emph{user prompt} contains one query's task: the question $q$, the gold answer $y$, and three contexts in an $A\!\rightarrow\!B\!\rightarrow\!C$ structure where $A$ is the original context $x$ (the fixed validity reference), $B$ is the base the editor rewrites (the parent $P$, or $x$ for a seed), and $C$ is the child it must produce. A part that changes from query to query is the \emph{edit program}: a requested rewrite type paired with a prompt strategy (Appendix~\ref{app:editor_edit_programs}; the ten strategies appear in Table~\ref{tab:rewrite_types_prompt_strategies_table}). The editor's identity and generation settings are in the configuration table (Appendix~\ref{app:configuration}).

We build each query's user prompt by filling the template's blanks with the example's fields and the edit program. The \emph{inheritance block} supplies $B$: the seed version when there is no parent (so $B=x$), or the offspring version when there is a parent to build on ($B=P$). The edit program fills two further fields. The rewrite type supplies the text at \texttt{\{edit\_type\_guidance\}} and the strategy the text at \texttt{\{prompt\_variant\_guidance\}}. We then send the system prompt and the filled user prompt to the editor together.

\begin{promptbox}[title=System prompt (MAP-Elites)]
\begin{Verbatim}[breaklines=true, breakanywhere=true, breaksymbolleft={}, fontsize=\small]
You are a controlled evidence editor for question-answering datasets. 

Produce exactly the requested invariant edit while preserving the correct answer and every fact needed to support it. Treat the original context only as the immutable validity reference A. When a parent B is provided, mutate B; do not independently rewrite A.

Return structured output only. Respond with exactly four lines starting with:
'STATUS:', 'EDITED_CONTEXT:', 'EXPECTED_ANSWER:', and 'EDIT_RATIONALE:'.

EXPECTED_ANSWER is bookkeeping only and must copy the provided original answer exactly without paraphrasing or expanding it. Use STATUS: proposed when you can produce a valid answer-preserving edit. Use STATUS: failed when you cannot produce one safely.
\end{Verbatim}
\end{promptbox}
 
\begin{promptbox}[title=User prompt]
\begin{Verbatim}[breaklines=true, breakanywhere=true, breaksymbolleft={}, fontsize=\small]
ORIGINAL VALIDITY REFERENCE (A)
Question: {question}
Fixed original answer: {answer}
Original context: {original_context}
 
{inheritance_block}
REQUESTED CHILD CONTRACT (C)
Child rewrite type: {requested_edit_type}
Child prompt variant: {prompt_variant}
Required textual operation: {edit_type_guidance}
Mutation-magnitude guidance: {prompt_variant_guidance}
The textual-operation contract takes precedence over the prompt variant. Do not substitute a different edit type. The dataset answer specification is fixed: copy the answer exactly, preserve its evidence, and do not introduce a new supported answer. 

If a parent B is provided, C must differ from B after Unicode and whitespace normalization.
\end{Verbatim}
\end{promptbox}
 
\paragraph{Inheritance block.}
We use the following block for a seed candidate:
\begin{promptbox}[title=Seed inheritance block]
\begin{Verbatim}[breaklines=true, breakanywhere=true, breaksymbolleft={}, fontsize=\small]
SEED MUTATION BASE (B)
No archived parent exists. For this seed only, use original context A as B.
\end{Verbatim}
\end{promptbox}
We indicate the parent and the explicit delta for an offspring:
\begin{promptbox}[title=Offspring inheritance block]
\begin{Verbatim}[breaklines=true, breakanywhere=true, breaksymbolleft={}, fontsize=\small]
ARCHIVED PARENT MUTATION BASE (B)
Parent candidate id: {parent_candidate_id}
Parent edited context: {parent_edited_context}
 
EXPLICIT MUTATION DELTA (B -> C)
Rewrite type: {parent_edit_type} -> {child_edit_type}
Prompt variant: {parent_prompt_variant} -> {child_prompt_variant}
Apply this delta to B. B is the text to modify; A is only the validity reference.
\end{Verbatim}
\end{promptbox}
 
\paragraph{Broad operation strings.}
The requested rewrite type fills \texttt{Required textual operation} field in \emph{user prompt} with one of:
\begin{promptbox}[title=Broad operation strings]
\begin{Verbatim}[breaklines=true, breakanywhere=true, breaksymbolleft={}, fontsize=\small]
distractor: retain every sentence of mutation base B in the same order and add at least one new plausible sentence that is irrelevant to answering the question

reorder: change the order of at least two complete sentences from mutation base B without adding, deleting, or paraphrasing those sentences

style_rewrite: rewrite the wording or syntax of substantive text from mutation base B; a mere sentence insertion or sentence reordering does not satisfy this operation
\end{Verbatim}
\end{promptbox}
If the strategy and the required edit type ever conflict, the editor must satisfy the edit type as the strategy shapes only how the edit is carried out.
 
\paragraph{Per-strategy guidance.}
Each of the ten strategies (Table~\ref{tab:rewrite_types_prompt_strategies_table}) supplies the \texttt{Mutation-magnitude guidance} field in \emph{user prompt}. A per-candidate ``strategy realized'' flag does \emph{not} gate cell placement; only the broad-type realization guard can reject (Appendix~\ref{app:realization}). These guidance strings are:
\begin{promptbox}[title=Per-strategy guidance strings]
\begin{Verbatim}[breaklines=true, breakanywhere=true, breaksymbolleft={}, fontsize=\small]
distractor_support_proximal:
  Retain mutation base B verbatim and insert exactly one plausible but answer-irrelevant sentence immediately before or after the sentence containing the fixed answer. The inserted sentence must not state, contradict, or help derive the answer.
  
distractor_question_overlap:
  Retain mutation base B verbatim and insert exactly one plausible but answer-irrelevant sentence. Reuse meaningful non-answer terms from the question so the distractor is lexically tempting, but do not state, contradict, or help derive the fixed answer.
  
reorder_adjacent_swap:
  Swap exactly one adjacent pair of complete sentences in mutation base B. Keep every sentence byte-for-byte otherwise and add or delete nothing.
  
reorder_support_back:
  Move the complete sentence containing the fixed answer to the final sentence position. Preserve the relative order and exact wording of every other sentence.
  
reorder_global:
  Globally permute at least three complete sentences from mutation base B. This must not be only one adjacent swap; preserve every sentence exactly and add or delete nothing.
  
support_direct_consolidation:
  Rewrite the support into one self-contained sentence in which the fixed answer and the question-defining relation are explicit in the same clause. Avoid a pronoun that requires another sentence to resolve the answer.
  
lexical_paraphrase_biased:
  Prefer a fluent paraphrase that preserves the exact answer and supporting meaning, but changes wording substantially. Keep the support in a single sentence when possible. Reduce direct lexical copying from the original wording.
  
paraphrase_biased:
  Prefer a fluent paraphrase or style rewrite that keeps the answer span stable while
  changing wording and sentence form.
  
controlled_coreference_rewrite:
  Rewrite the support using one explicit, unambiguous antecedent followed by a controlled pronoun or referring expression. The supporting relation should require resolving that reference, while the fixed answer remains verbatim and fully supported.
  
multi_sentence_factored:
  Prefer a rewrite that splits or distributes the supporting information across multiple sentences. The original answer must remain recoverable from the edited context, but the evidence may be spread across multiple sentences.
\end{Verbatim}
\end{promptbox}

Three further variant strings appear in the editor code but not in the hard staged
registry: \texttt{support\_rewrite\_aggressive} and \texttt{local\_conservative} are
the prompt variants used by the U-ME (Appendix~\ref{app:search_soft}). The FP uses a single fixed \texttt{llm\_paraphrase} prompt (Appendix~\ref{app:search_paraphrase}).

\paragraph{FP prompt.} The FP does \emph{not} use the templates above. It requests a direct paraphrase of the original context and never conditions on a parent, so it uses its own fixed system and user prompt without $A\!\rightarrow\!B\!\rightarrow\!C$ structure, inheritance block, and per-strategy guidance. It shares only the four-line output contract and the chat-format delivery. Its two prompts are:
\begin{promptbox}[title=System prompt (LLM-paraphrase)]
\begin{Verbatim}[breaklines=true, breakanywhere=true, breaksymbolleft={}, fontsize=\small]
You are a controlled evidence paraphraser for question-answering datasets. Rewrite the provided original context directly, without using or imagining a parent edit. Preserve every fact, omission, evidential relation, and the correct answer. Do not add distractors, new claims, interpretations, or unsupported details. 

Return structured output only. Respond with exactly four lines starting with: 'STATUS:', 'EDITED_CONTEXT:', 'EXPECTED_ANSWER:', and 'EDIT_RATIONALE:'. 
EXPECTED_ANSWER is bookkeeping only and must copy the fixed answer exactly. Use STATUS: proposed only when the result is a faithful answer-preserving paraphrase; otherwise use STATUS: failed.
\end{Verbatim}
\end{promptbox}
\begin{promptbox}[title=User prompt (LLM-paraphrase)]
\begin{Verbatim}[breaklines=true, breakanywhere=true, breaksymbolleft={}, fontsize=\small]
Question: {question}
Fixed answer: {answer}
Original context to paraphrase: {context}
 
Task: Produce one fluent paraphrase of the complete original context. Change substantive wording and/or syntax so the text is observably different, while preserving the fixed answer string and all evidence needed to answer the question. Do not merely reorder sentences, insert a sentence, summarize, omit information, or introduce any new information.
\end{Verbatim}
\end{promptbox}

\paragraph{Output format and parsing.}
A valid response is four indicated field lines:
\begin{promptbox}[title=Output format]
\begin{Verbatim}[breaklines=true, breakanywhere=true, fontsize=\small]
STATUS: proposed
EDITED_CONTEXT: {complete edited context}
EXPECTED_ANSWER: {exact copy of the fixed original answer}
EDIT_RATIONALE: {explanation of the edit and answer preservation}
\end{Verbatim}
\end{promptbox}
If the editor cannot produce the edit successfully, we expect it to return \texttt{STATUS: failed} with an empty edited context. The parser reads the lines beginning with the four exact fields; a \texttt{proposed} response with no edited context is marked malformed. Only \texttt{EDITED\_CONTEXT} is consumed downstream; \texttt{EXPECTED\_ANSWER} is compared to the gold answer but does not gate admission (Appendix~\ref{app:deterministic_gates}).

\section{Datasets}\label{app:datasets}
We evaluate on five extractive QA datasets chosen to span domains and difficulty. SQuAD~v1.1 \citep{rajpurkar-etal-2016-squad} provides standard Wikipedia-based span questions; Quoref \citep{dasigi-etal-2019-quoref} targets questions whose answers require resolving coreference; NewsQA \citep{trischler-etal-2017-newsqa} draws on news articles with longer contexts; the Natural Questions partition of MRQA \citep{fisch-etal-2019-mrqa,kwiatkowski-etal-2019-natural} supplies naturally occurring information-seeking questions; and AdversarialQA \citep{bartolo-etal-2020-beat} contributes model-adversarial questions that are hard by construction. All five are extractive, so the gold answer is a span recoverable from the context. We provide dataset statistics in Table \ref{tab:datasets}
 
We draw a fixed sample of $1000$ examples per dataset and share it across all constructors. Sampling is random without replacement under the global seed ($7$), and the same sampled example identifiers are used by every constructor within a dataset, so constructors are never compared on different examples. We filter the questions such that only answerable questions remain as samples in our study. Therefore, we discard the examples with empty context, question, or answer. All of the samples included in the experiments have at least one gold answer.

We map every dataset's gold answers into one uniform format, an \emph{answer specification}, before any validation or scoring, such that the downstream checks do not depend on how a dataset store its answers. A question's answer specification is the set of answer strings that count as correct, with the rule that matching any of them suffices. For instance, a question whose listed gold answers are \{''Franklin D.\ Roosevelt'', ''FDR'', ''Roosevelt Franklin D.\'', ''Roosevelt''\} yields an answer specification accepting all four, and a prediction matching any one of them (under the normalization of Appendix~\ref{app:prediction_scoring}) answers the question. This uniform representation is what lets answer preservation (Appendix~\ref{app:deterministic_gates}) and the failure label (Appendix~\ref{app:evaluation_protocol}) apply identically across datasets.

\begin{table}[!ht]
\centering
\small
\setlength{\tabcolsep}{6pt}
\caption{The five evaluation datasets, with statistics over the sampled $1000$ examples
each. All lengths are whitespace-word counts; answer length is that of the primary (first)
gold answer. ``Multi-gold'' is the fraction of questions with more than one gold answer.\\}
\label{tab:datasets}
\begin{tabular}{@{}lcccc@{}}
\toprule
\textbf{Dataset} & \textbf{Context (mean/max)} & \textbf{Question (mean)} & \textbf{Answer (mean)} & \textbf{Multi-gold} \\
\midrule
SQuAD v1.1              & $120.9 / 486$   & $10.0$ & $3.1$ & $11.2$ \\
Quoref                  & $334.9 / 781$   & $15.4$ & $1.7$ & $9.0$ \\
NewsQA                  & $704.0 / 1980$  & $7.8$  & $5.5$ & $0.0$ \\
NaturalQuestions (MRQA) & $161.9 / 800$   & $9.1$  & $4.4$ & $9.8$ \\
AdversarialQA           & $116.0 / 327$   & $9.8$  & $3.1$ & $0.0$ \\
\bottomrule
\end{tabular}
\end{table}
 
Per-dataset original-context accuracy, which characterizes how hard each dataset is for the
QA models, is reported with the results in Appendix~\ref{app:model_acc_results}, providing the
failure-detection task a base rate on every dataset.

\section{Guarantees and Limitations}\label{app:guarantees_limitations}
Two properties hold for every rewrite in the archive by construction. First, admission is deterministic and independent of the QA model $f_\theta$: a rewrite is stored only once it clears the gates of Appendix~\ref{app:deterministic_gates}, none of which query $f_\theta$.  A pretrained model (BERTScore) enters only at the later ranking stage, when the search objective score ranks the valid rewrites within a cell. It can change \emph{which} valid rewrite is kept, never \emph{whether} an invalid one is admitted (Appendix~\ref{app:quality_score}). Second, the rewrite-type axis $\mathcal{T}$ is a \emph{verified} label, not merely a requested one, for NE, SBS and FP: their realization guard rejects any rewrite that does not exhibit the requested edit. U-ME omits that guard and labels by request alone (Table~\ref{tab:constructors}).
 
Three properties are deliberately not certified, and we do not claim them. First, meaning is not verified: the support-explicitness axis records \emph{where} the answer's cues sit, not that meaning is preserved, and its \texttt{paraphrastic} bucket is a residual with no
paraphrase test (Appendix~\ref{app:support_explicitness}). Likewise an inserted distractor is checked only to be a genuinely new sentence, not to be answer-irrelevant (Appendix~\ref{app:realization}). Second, the search objecive terms are lightweight surface proxies, they are not learned by a model or human judgments of syntax, fluency, or semantics. They shape ranking only, never admission (Appendix~\ref{app:quality_score}). Third, generations are not bit-for-bit reproducible: the editor samples without a fixed per-call seed, so we treat each run's archive as a fixed artifact and compute all results on it (Appendix~\ref{app:search_scheduler}). Wherever a property is imposed by the prompt rather than checked by a gate, we report it as requested, not verified.

\end{document}